%% file: main.tex
\documentclass[sigconf,edbt]{acmart-edbt2025}

\def\BibTeX{{\rm B\kern-.05em{\sc i\kern-.025em b}\kern-.08em
    T\kern-.1667em\lower.7ex\hbox{E}\kern-.125emX}}

\usepackage{balance}
\usepackage{booktabs} 

\setcopyright{rightsretained}

\acmDOI{}

\acmISBN{978-3-89318-098-1}

\acmConference[EDBT 2025]{28th International Conference on Extending Database Technology (EDBT)}{25th March-28th March, 2025}{Barcelona, Spain}
\acmYear{2025}

\usepackage{etoolbox}
\makeatletter
\patchcmd{\@thm}{\trivlist}{\list{}{\leftmargin=0pt}}{}{}
\makeatother
\usepackage{aligned-overset}

\usepackage{enumitem}
\usepackage{svg}

\usepackage{algorithm}
\usepackage{algorithmicx}
\usepackage{algpseudocode}
\usepackage{adjustbox}
\usepackage{graphicx}
\usepackage{subcaption}
\usepackage{amsthm}

\newcommand{\eat}[1]{}
\newcommand{\Indent}{\hspace{\algorithmicindent}}

\newcommand{\Dt}[1]{\Vec{u}_{t}^{(#1)}}
\newcommand{\Dtm}[0]{\overline{\Vec{u}}_t}
\renewcommand{\Vec}[1]{ \mathbf{#1} }
\newcommand{\OVec}[1]{\overline{\Vec{#1}}}

\newcommand{\IVec}[2]{\Vec{#1}^{( #2 )}}

\newcommand{\normsq}[1]{\left\Vert #1 \right\Vert_2^2}
\newcommand{\abs}[1]{\left| #1 \right|}
\newcommand{\norm}[1]{\left\Vert #1 \right\Vert_2}
\newcommand{\parenthesis}[1]{\left( #1 \right)}
\newcommand{\mybrackets}[1]{\left[ #1 \right]}
\newcommand{\dotp}[2]{\left\langle #1 \, , \, #2 \right\rangle}
\newcommand{\dotpN}[2]{\langle #1 \, , \, #2 \rangle}
\DeclareMathOperator{\sk}{sk}
\newcommand{\Var}[1]{\operatorname{\mathcal{V}\!\text{\textit{ar}}}\left(#1\right)}

\begin{document}
\title{Communication-Efficient Distributed Deep Learning via Federated Dynamic Averaging}

\author{Michail Theologitis}
\affiliation{%
  \institution{Technical University of Crete}
  \city{Chania}
  \country{Greece}
}
\email{mtheologitis@tuc.gr}

\author{Georgios Frangias}
\affiliation{%
  \institution{Technical University of Crete}
  \city{Chania}
  \country{Greece}
}
\email{gfrangias@tuc.gr}

\author{Georgios Anestis}
\affiliation{%
  \institution{Technical University of Crete}
  \city{Chania}
  \country{Greece}
}
\email{ganestis@tuc.gr}

\author{Vasilis Samoladas}
\affiliation{%
  \institution{Technical University of Crete}
  \city{Chania}
  \country{Greece}
}
\email{vsamoladas@tuc.gr}

\author{Antonios Deligiannakis}
\affiliation{%
  \institution{Technical University of Crete}
  \city{Chania}
  \country{Greece}
}
\email{adeli@softnet.tuc.gr}

\renewcommand{\shortauthors}{}

\begin{abstract}
  The ever-growing volume and decentralized nature of data, coupled with the need to harness it and extract knowledge, have led to the extensive use of distributed deep learning (DDL) techniques for training. These techniques rely on local training performed at distributed nodes using locally collected data, followed by a periodic synchronization process that combines these models to create a unified global model. However, the frequent synchronization of deep learning models, encompassing millions to many billions of parameters, creates a communication bottleneck, severely hindering scalability. Worse yet, DDL algorithms typically waste valuable bandwidth and render themselves less practical in bandwidth-constrained federated settings by relying on overly simplistic, periodic, and rigid synchronization schedules. These inefficiencies make the training process increasingly impractical as they demand excessive time for data communication. To address these shortcomings, we propose Federated Dynamic Averaging (\textsc{FDA}), a communication-efficient DDL strategy that dynamically triggers synchronization based on the value of the model variance. In essence, the costly synchronization step is triggered only if the local models---initialized from a common global model after each synchronization---have significantly diverged. This decision is facilitated by the transmission of a small local state from each distributed node. Through extensive experiments across a wide range of learning tasks we demonstrate that \textsc{FDA} reduces communication cost by orders of magnitude, compared to both traditional and cutting-edge communication-efficient algorithms. Additionally, we show that \textsc{FDA} maintains robust performance across diverse data heterogeneity settings.
\end{abstract}
%
%



\maketitle

\input{fda-body}

\bibliographystyle{ACM-Reference-Format}
\bibliography{references}

%
% ...

\end{document}

%% file: fda-body.tex
\section{Introduction}
The big data era has been marked by an unprecedented scale of training datasets~\cite{zhang2021data, RaufFP24}. 
These datasets are not only growing in size, but are often physically distributed and cannot be easily centralized due to business considerations, privacy concerns, bandwidth limitations---particularly in federated settings, such as drones collecting and collaboratively building a global model or view of an area---and data sovereignty laws~\cite{karouz2019advancementsFL, wu2020privacy_preserv_fl, CormodeMS24private}.
Such constraints pose significant challenges to the application of Deep Learning (DL) techniques.

Distributed Deep Learning (DDL) has emerged as an alternative paradigm to the traditional centralized approach~\cite{zinkevich2010parallelSGD, chilimbi2014distributed}, offering efficient learning over large-scale data across multiple worker-nodes, enhancing the speed of training DL models and paving the way for more scalable and resilient DL applications~\cite{um2023fastflow, ma2023fec_comm, lai2023queryDBdeeplearning, zhou2022vision, DavitkovaGM24}.
Most DDL methods are iterative, where each iteration involves a phase of local training followed by the \emph{synchronization} of the local models with the global one. 

The predominant method, based on the bulk synchronous parallel (BSP) approach~\cite{valiant1990bsp}, averages the local model updates and applies the averaged update to each local model~\cite{zinkevich2010parallelSGD}. 
Less synchronized variants have also been proposed to ameliorate the effect of \textit{straggler workers}, but they often compromise convergence speed and model quality~\cite{miao2021partial_async, fu2022stale}.

A significant challenge inherent in the traditional techniques, especially in federated DL settings where models are huge and worker interconnections are slow, is the communication bottleneck, which restricts system scalability~\cite{tang2023gossipfl, wang2016databases_and_ddl}. Specifically, the communication bottleneck arises from the frequent exchange (synchronization) of model parameters---often in the range of billions---across distributed workers. The synchronization process entails substantial data volume transfer and generally dominates the overall training time, leading to a low computation-to-communication ratio~\cite{shi2021surveyUsingBert338mil, fu2022stale}. Addressing this challenge to expedite DDL algorithms has been a focal point of research for many years; speeding-up SGD is arguably among the most impactful and transformative problems in machine learning~\cite{wang2021cooperativeSGD}.

The most direct method to alleviate the communication burden is to reduce the frequency of communication rounds. Local-SGD is the prime example of this approach. It allows workers to perform $\tau$ local update steps on their models before aggregating them, as opposed to averaging the updates in every step~\cite{yu2018parallel, haddadpour2019localSGD}. Although Local-SGD is effective in reducing communication while maintaining comparable model quality~\cite{wang2021cooperativeSGD},  determining the optimal value of $\tau$ presents a critical challenge, with only a handful of studies offering theoretical insights into its influence on convergence~\cite{wang2021cooperativeSGD, stich2019local, yu2018parallel}. 

To further reduce communication costs of Local-SGD, more sophisticated communication strategies introduce varying sequences of local update steps $\{\tau_0, ..., \tau_R\}$, instead of a 
fixed $\tau$. In~\cite{wang2018adaptive}, in order to minimize convergence error with respect to wall-time, the authors proposed a decreasing sequence of local update steps. Conversely, the focus in~\cite{haddadpour2019localSGD} was on reducing the number of communication rounds for a fixed number of model updates and an increasing sequence emerged. These contrasting approaches underscore the multifaceted nature of communication strategies in distributed deep learning, highlighting not only the absence of a one-size-fits-all solution but also the growing need for dynamic, context-aware strategies that can continuously adapt to the specific intricacies of the learning task.

\vspace{1mm}
\noindent \textbf{Main Idea and Contributions.} Our work addresses critical efficiency challenges in DDL, particularly in communication-constrained environments, such as the ones encountered in Federated Learning (FL) applications~\cite{karouz2019advancementsFL}. We introduce Federated Dynamic Averaging (\textsc{FDA}), a novel, adaptive distributed deep learning strategy that massively improves communication efficiency over previous work. 

\textsc{FDA} utilizes a novel 2-action, conditional synchronization protocol, designed to 
avoid the need to decide or guess the proper values of local update steps, or to synchronize after each training step, but rather only performs the costly synchronization process \emph{when needed}. Our \textsc{FDA} algorithm dynamically triggers synchronization based on the value of \emph{model variance} across worker-nodes. In a nutshell, the costly synchronization step is only triggered if the local models have diverged significantly, which implies that the global model may no longer be accurate. 

\begin{figure}[t]
  \centering
  \includegraphics[width=0.49\textwidth]{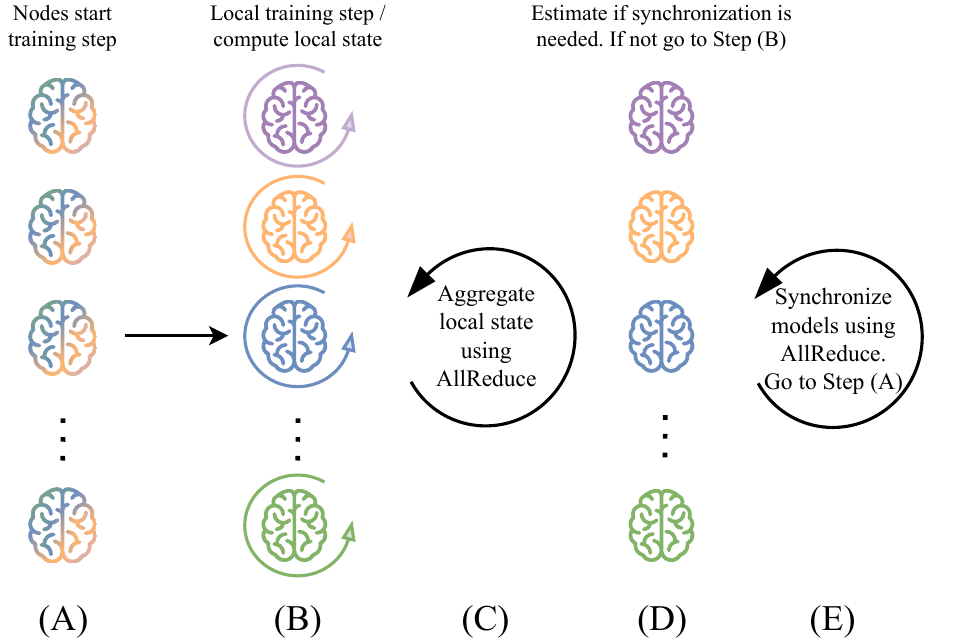}
  \caption{\textsc{FDA}. The local training step is followed by the computation of a local state by all worker-nodes. Then, the (small in size) local states are aggregated. Based on the aggregated result, all workers estimate if synchronization is required. In most cases, the expensive synchronization step of the models is avoided and local training continues}
  \label{fig:fda}
\end{figure}

As Figure~\ref{fig:fda} demonstrates, at the start, workers enter the local training step with the same global model (Figure~\ref{fig:fda}.A). Then, local training commences and each distributed worker-node computes its local state, which encapsulates helpful information for estimating the model variance (Figure~\ref{fig:fda}.B). This is followed by the transmission (Figure~\ref{fig:fda}.C) of these small-size local states, an operation that is bandwidth- and time-efficient because of their small size. During transmission, the local states are aggregated and their average is made available to all workers---an operation known as \textsc{AllReduce}. This operation does not require (or prohibit) the use of a central node. Based on the aggregated state, the workers can estimate (Figure~\ref{fig:fda}.D) whether the variance of the local models may have exceeded a threshold. If this is not the case, the costly synchronization step (Figure~\ref{fig:fda}.E) is avoided and local training continues. What is important is how to properly pick these local states computed at, and then transmitted by, the local workers. To address this problem, we propose two variants of our \textsc{FDA}  algorithm. 
Our contributions can be summarized as follows:
\begin{itemize}[leftmargin=*]
    \item We propose \textsc{FDA}, an algorithm that dynamically decides to synchronize local workers when \emph{model variance across workers} exceeds a threshold. This strategy drastically reduces communication, while preserving cohesive progress towards the shared training objective.

    \item We propose two variants of \textsc{FDA}, which differ in the amount of information preserved in the local states that are transmitted by each worker and aggregated for subsequent estimation of model variance. These two variants, termed \textsc{SketchFDA} and \textsc{LinearFDA}, offer a different balance between communication efficiency and approximation accuracy.
    
    \item We evaluate and compare \textsc{FDA} with other DDL algorithms through a comprehensive  suite of experiments with diverse datasets, models, and tasks. Our experiments demonstrate that \textsc{FDA} outperforms traditional and contemporary FL algorithms by 1-2 orders of magnitude in communication savings, while maintaining equivalent model performance.
    Furthermore, it effectively balances the competing demands of communication and computation, providing greatly improved trade-offs.

    \item We demonstrate \textsc{FDA}'s robustness in various challenging Non-IID settings, common in real-world Federated Learning applications. While state-of-the-art methods typically require substantially more resources to converge under \hbox{Non-IID} conditions, \textsc{FDA} maintains consistent and comparable performance across both IID and Non-IID settings.

\end{itemize}

\vspace{1mm}
\noindent \textbf{Outline.} The remainder of this paper is organized as follows: Section \ref{section:rel_work} reviews related work. Section \ref{section:FDA} introduces our DDL technique, Federated Dynamic Averaging (\textsc{FDA}), and its two variants. Section \ref{section:exps} details the experimental setup, and discusses the insights and conclusions drawn from our empirical investigation. Lastly, Section \ref{section:conclusion} contains concluding remarks.

\section{Related Work}\label{section:rel_work}

\vspace{1mm}
\textbf{Problem formulation}. Consider distributed training of deep neural networks over multiple workers~\cite{Dean2012distributed, li2021communicationefficient}. In this setting, each worker represents a data owner (equivalently, a local model owner) and has access to its own set of training data $\mathcal{D}_k$. Workers can utilize any available hardware they possess (e.g., GPUs, CPUs) to perform learning steps. The collective goal is to find a common model $\Vec{w}\in \mathbb{R}^d$ by minimizing the overall training loss. This scenario can be effectively modeled as a distributed optimization problem, formulated as follows:
\begin{equation}\label{eq:optimization_prob}
    \underset{\Vec{w} \in \mathbb{R}^d}{\mathrm{minimize}} \; \; F(\Vec{w}) \triangleq \frac{1}{K} \sum_{k=1}^{K} F_k(\Vec{w})
\end{equation}
where $K$ is the number of workers and $F_k(\Vec{w}) \triangleq \mathbb{E}_{\zeta_k \sim \mathcal{D}_k} \left[ \ell(\Vec{w}; \zeta_k) \right]$ is the local objective function for worker $k$. Function $\ell(\Vec{w}; \zeta_k)$ represents the \textit{loss} for data sample $\zeta_k$ given model $\Vec{w}$.

\vspace{1mm}
\noindent \textbf{Solution direction.} As noted in the seminal work \cite{karouz2019advancementsFL}, research in FL should focus primarily on synchronous solutions. This allows different lines of research (e.g., compression, privacy, etc.) to be developed independently and then combined seamlessly. Our work, along with most communication-efficient FL strategies, adheres to this synchronous paradigm. However, such approaches may be less effective in environments where each communication operation incurs significant overhead regardless of the size of the data being transmitted (e.g., high-latency). In these scenarios, asynchronous mechanisms become necessary, though they typically fall outside the primary focus of contemporary FL research. That said, FDA can be modified to work asynchronously (as explained in Section~\ref{section:discussion}).

\vspace{1mm}
\noindent \textbf{Communication efficient Local-SGD.} The work in~\cite{li2021communicationefficient} decomposes each round into two phases. In the first phase, each worker runs Local-SGD with $\tau=I_1$, while the second phase runs $I_2$ steps with $\tau=1$;~\cite{li2021communicationefficient} proposes to exponentially decay $I_1$ every $M$ rounds. In the heterogeneous setting, the work in~\cite{qin2022roleofSteps}, by analysing the convergence rate, proposes an increasing sequence of local update steps for strongly-convex local objectives and fixed local update steps for other types of local objectives. The study in~\cite{yu2019computation} dynamically increases batch sizes to reduce communication rounds, maintaining the same convergence rate as SSP-SGD. However, the large-batch approach leads to poor generalization~\cite{hoffer2018genGap}, a challenge addressed by the post-local SGD method~\cite{lin2018postLocal}, which divides training into two phases: BSP-SGD followed by Local-SGD with a fixed number of steps. In the Lazily Aggregated Algorithm (LAG)~\cite{chen2018lag}, a different approach was taken, using only new gradients from some selected workers and reusing the outdated gradients from the rest, which essentially skips communication rounds. 

Federated Averaging (FedAvg)~\cite{mcmahan2017FedAvg} is another representative of communication efficient Local-SGD algorithms, which is a pivotal method in Federated Learning (FL)~\cite{karouz2019advancementsFL}. In the FL setting with edge computing systems, the work in~\cite{wang2019adaptiveFL} tries to find the optimal synchronization period $\tau$ subject to local computation and aggregation constraints. Recently~\cite{mills2023faster}, in the FL setting with the assumption of strongly-convex objectives, by analysing the balance between fast convergence and higher-round completion rate, a decaying local update step scheme emerged.

Unlike previous approaches that rely on predetermined synchronization schedules (fixed, decaying, or otherwise), our work introduces a dynamic synchronization strategy. \textsc{FDA} adapts continuously during the training process, basing synchronization decisions on a real-time metric: the model variance across workers.

\vspace{1mm}
\noindent \textbf{Accelerating convergence.} An indirect, yet highly effective way to mitigate the communication burden in DDL, is to speed up convergence. Consequently, recent works have built upon communication efficient Local-SGD methods by deploying accelerated versions of SGD to the distributed setting. Specifically, FedAdam~\cite{reddi2021fedadam} extends Adam~\cite{kingma2017adam} and FedAvgM~\cite{hsu2019fedAvgM} extends SGD with momentum (SGD-M)~\cite{sutskever2013momentumSGD}. Recently, Mime~\cite{karimireddy2021mime} provides a framework to adapt arbitrary centralized optimization algorithms to the FL setting. However, these methods still suffer from the model divergence problem, particularly in heterogeneous settings. When solving \eqref{eq:optimization_prob}, the disparity between each worker's optimal solution $\Vec{w}_k^*$ for their objective $F_k$, and the global optimum $\Vec{w}^*$ for $F$, can potentially cause worker models to diverge (drift) towards their disparate minima~\cite{kale2019scaffold, reddi2021fedadam, WenigP22}. The result is slow and unstable convergence with significant communication overhead. To address this problem, the SCAFFOLD algorithm~\cite{kale2019scaffold} used control-variates (in the same spirit to SVRG), with significant speed-up. FedProx~\cite{kumar2018fedProx} re-parameterized FedAvg~\cite{mcmahan2017FedAvg} by adding $L^2$ regularization in the workers' objectives to be near the global model. Lastly, FedDyn~\cite{acar2021fedDyn} improved upon these ideas with a dynamic regularizer making sure that if local models converge to a consensus, this consensus point aligns with the stationary point of the global objective function.

While these approaches primarily focus on enhancing the optimization process and typically employ fixed synchronization intervals (e.g., every local epoch), our work addresses a complementary aspect: determining the optimal timing for synchronization. \textsc{FDA}'s dynamic synchronization strategy is orthogonal to these optimization techniques and can be integrated with them by simply adjusting the synchronization decision.

\vspace{1mm}

\noindent \textbf{Compression.} To reduce communication overhead in DDL, significant efforts have been directed towards minimizing message sizes. Key strategies include sparsification, where only crucial components of information are transmitted, as explored in~\cite{aji2017sparsification}, and quantization techniques, which involve transmitting only quantized gradients, as detailed in~\cite{shlezinger2020quantizationFL}. These techniques can be combined with Local-SGD methods to enhance communication-efficiency further. An example is Qsparse-local-SGD~\cite{basu2019qsparselocalsgd}, which integrates aggressive sparsification and quantization with Local-SGD, achieving substantial communication savings. Crucially, \textsc{FDA} is fully compatible with any technique that reduces the cost of synchronization (e.g. model compression). Our approach simply adjusts the timing of the synchronization decision without altering the data being synchronized. This ensures that any compression technique effective in traditional methods (BSP, Local-SGD, etc.) will be equally effective when deployed with \textsc{FDA}. Therefore, the communication savings demonstrated in the relevant literature~\cite{wang2023compressionInDDL} can be safely expected to carry over to our approach as well.

Additionally, sketching emerges as another fundamental tool in large-scale machine learning. It effectively compresses high-dimensional problems into lower dimensions to save runtime and memory, typically utilizing hash-based probabilistic data structures. For instance, \cite{Spring2019grad} use Count Sketches to compress auxiliary variables in optimization algorithms, significantly freeing up memory. Similarly, FetchSGD~\cite{rothchild2020fetchsgd} employs Count Sketches to compress model updates and leverages their linearity for efficient merging. In contrast to these applications, our approach utilizes sketches not for compression but to estimate local state information, and based on this to decide whether a synchronization is required---an orthogonal application to traditional use cases. A comprehensive survey of compression techniques in DDL can be found in~\cite{wang2023compressionInDDL}.

\section{Federated Dynamic Averaging}\label{section:FDA}
We now present our algorithms, based on our notion of Federated Dynamic Averaging (FDA). Our algorithms deviate from prior work in these two key ways:
\begin{enumerate}[leftmargin=20pt]
    \item The decision on when to synchronize.
    \item The actual synchronization process.
\end{enumerate}
To the best of our knowledge, this is the first Distributed Deep Learning algorithm that dynamically decides when to synchronize based on the current collective state of the training progress---whether it is advancing well or poorly.

\vspace{1mm}
\noindent \textbf{Notation.} At each time step $t$, each worker $k$ independently maintains its own vector of model parameters\footnote{The terms ``model'' and "model parameters" are used interchangeably, as is common in the literature.},
denoted as $\Vec{w}_t^{(k)} \in \mathbb{R}^d$. Let $\Vec{w}_t$ represent the $K\times d$ tensor of all local model vectors, and $\OVec{w}_t$ be the average model vector (this notation applies to all vector quantities):
\begin{align*}
    \Vec{w}_t = \mybrackets{\Vec{w}_t^{(1)}, \ldots , \Vec{w}_t^{(K)}} \quad , \quad \OVec{w}_t = \frac{1}{K} \sum_{k=1}^K \Vec{w}_t^{(k)}
\end{align*}
Furthermore, let $\textsc{Optimize}(\Vec{w} , \mathcal{B})$ be the updated model~\cite{DeepLearningGoodfellow} computed by some optimization algorithm (e.g., SGD, Adam) using the model $\Vec{w}$, and the batch $\mathcal{B}$ of training data. It incorporates the learning rate, loss function and relevant gradients. 
During step $t$, each worker $k$ first applies the update:
\begin{align*}
    \Vec{w}^{(k)}_t = \textsc{Optimize}(\Vec{w}_{t-1}^{(k)}\, , \, \mathcal{B}_t^{(k)})
\end{align*}
Moreover, operation $\textsc{AllReduce}(\Vec{w}_{t}^{(k)})$ computes and returns the average model vector \cite{li2020pytorchdistributed}:
\begin{align*}
    \OVec{w}_t = \textsc{AllReduce}(\Vec{w}_{t}^{(k)})
\end{align*}
Workers \emph{synchronize} by executing $\textsc{AllReduce}(\Vec{w}_{t}^{(k)})$, thereby setting $\Vec{w}_{t}^{(k)} := \OVec{w}_t$. 
If synchronization is not performed at step $t$, each worker continues training with its locally updated model. A comprehensive list of the notation used throughout this section is provided in Table \ref{tab:notation}.

\begin{table}
  \caption{Notation}
  \label{tab:notation}
  \begin{tabular}{ll}
    \toprule
    Symbol & Meaning\\
    \midrule
    $\dotp{\cdot}{\cdot}$ & Dot product \\
    $t$ & Time step index \\
    $K$ & Number of workers\\
    $d$ & Model dimension\\
     $\mathcal{D}_k$  & Training data of worker $k$ \\
     $\mathcal{B}^{(k)}_t$ & A batch sampled from $\mathcal{D}_k$  \\
    $\Vec{w}_t^{(k)} \in \mathbb{R}^d$ & Model of worker $k$\\
    $\Vec{w}_t = [\Vec{w}_t^{(1)}, \ldots , \Vec{w}_t^{(K)}]$ & Tensor of local models\\
    $\OVec{w}_t = \frac{1}{K} \sum_{k=1}^K \Vec{w}_t^{(k)}$ & Average model (global model) \\
    $\OVec{w}_{t_0}$ & Model after most recent sync. \\
    $\OVec{w}_{t_{-1}}$ & Model after 2nd most recent sync. \\
    $\Vec{u}_t^{(k)} = \Vec{w}_t^{(k)} - \OVec{w}_{t_0}$ & Local model drift \\
    $\OVec{u}_t = \frac{1}{K} \sum_{k=1}^K \Vec{u}_t^{(k)}$ & Average model drift (global drift)\\
    $\Var{\Vec{w}_t}$ & Model variance\\
    $\Theta$ & Model variance threshold\\
    $\Vec{S}_t^{(k)}$ & State of worker $k$\\
    $\OVec{S}_t = \frac{1}{K} \sum_{k=1}^K \Vec{S}_t^{(k)}$ & Average state \\
     $H(\cdot)$ & Function for variance estimation \\
    $\sk(\cdot) : \mathbb{R}^d \to \mathbb{R}^{l \times m}$ & AMS sketch operator (\S\ref{section:Sketch}) \\
    $\mathcal{M}_2(\cdot) : \mathbb{R}^{l \times m} \to \mathbb{R}$ & $L^2$ norm squared estimate (\S\ref{section:Sketch}) \\
    $\epsilon$ & Error of sketch estimate (\S\ref{section:Sketch}) \\
    $(1-\delta)$ & Confidence of approximation (\S\ref{section:Sketch}) \\
    $l = \mathcal{O}(\log 1 / \delta)$ & \#Rows of sketch matrix (\S\ref{section:Sketch}) \\
    $m = \mathcal{O}(1 / \epsilon^2)$ & \#Columns of sketch matrix (\S\ref{section:Sketch}) \\
    $\xi = \frac{\OVec{w}_{t_0} - \OVec{w}_{t_{-1}}}{ \norm{\OVec{w}_{t_0} - \OVec{w}_{t_{-1}}} }$ &  Heuristic vec. for 
    \textsc{LinearFDA} (\S\ref{section:Linear}) \\
  \bottomrule
\end{tabular}
\end{table}

\vspace{1mm}
\noindent \textbf{Model Variance and FDA.} The \textit{model variance} quantifies the dispersion or spread of worker models around the average model: 
\begin{align}
    \Var{\Vec{w}_t} &= \frac{1}{K} \sum_{k=1}^K \normsq{\Vec{w}_t^{(k)} - \OVec{w}_t} 
\end{align}
This measure provides insight into how closely aligned the workers' models are at any given time. High variance indicates that the models are widely spread out, essentially drifting apart, leading to a lack of cohesion in the aggregated model. Conversely, a moderate or low variance suggests that the workers' models are closely aligned, working collectively towards the shared objective. 

The FDA algorithm (Algorithm~\ref{alg:FDA}) is based on the premise that, as long as the variance is below a threshold $\Theta$, synchronization is not needed. Thus, we introduce the \emph{Round Invariant} (RI):
\begin{equation}\label{eq:RI}
    \Var{\Vec{w}_t} \leq \Theta
\end{equation}

To preserve the RI, our FDA algorithm maintains (Lines~4-6 of Algorithm~\ref{alg:FDA}) at each worker $k$ a local (low-dimensional) state-vector $\Vec{S}_t^{(k)}$, which is computed based on $\Vec{w}_t^{(k)}$. These state vectors are vital for the subsequent estimation of the model variance, and underpin the two variants of the \textsc{FDA} algorithm (provided in Sections~\ref{section:Sketch} and~\ref{section:Linear}, respectively). Our estimation techniques begin by performing $\textsc{AllReduce}$ on the states $\Vec{S}_t^{(k)}$, consolidating them into the average state $\OVec{S}_t$ (Line~7). Importantly, this communication step requires significantly less bandwidth and resources than transmitting the full models $\Vec{w}^{(k)}_t$.

For each FDA variant, we also define a (different) function $H(\OVec{S}_t)$ that overestimates the variance, i.e., it ensures that as long as $H(\OVec{S}_t) \leq \Theta$ then the variance is bounded by $\Theta$. This guarantee is probabilistic for the Sketch-based variant of FDA, and deterministic for its Linear counterpart. 
Consequently, if $H(\OVec{S}_t) > \Theta$ then synchronization is performed (Lines~8-9) --- the RI invariant cannot be guaranteed. After synchronization, the model variance is zero.

\vspace{1mm}
\noindent \textbf{Efficiently Monitoring the RI.} Estimating model variance efficiently is at the heart of FDA.  To this end, we first introduce the \emph{local model drift}, $\Vec{u}_t^{(k)}$, and \emph{average drift}, $\OVec{u}_t$, defined as follows:
\begin{align*}
    \Vec{u}_t^{(k)} = \Vec{w}_t^{(k)} - \OVec{w}_{t_0} \quad , \quad \OVec{u}_t = \frac{1}{K} \sum_{k=1}^K \Vec{u}_t^{(k)} 
\end{align*}
Here, $\OVec{w}_{t_0}$ denotes the model vector after the most recent synchronization. Subsequently, the model variance can be written as:
\begin{equation}\label{eq:variance_def}
    \Var{\Vec{w}_t} = \parenthesis{ \frac{1}{K} \sum_{k=1}^K \normsq{\IVec{u}{k}_t}} - \normsq{\OVec{u}_t}
\end{equation}
\begin{proof} Adding an offset ($-\OVec{w}_{t_0}$) to each $\Vec{w}_t^{(k)}$ does not alter the variance, therefore:
\begin{align*}
    \Var{\Vec{w}_t} &= \Var{\Vec{w}_t-\OVec{w}_{t_0}} = \Var{\Vec{u}_t} = \frac{1}{K} \sum_{k=1}^K \normsq{\Vec{u}_t^{(k)} - \OVec{u}_t} \\
    &\hspace{-24pt}= \frac{1}{K} \sum_{k=1}^K \parenthesis{ \normsq{\Vec{u}_t^{(k)}} - 2 \dotp{\Vec{u}_t^{(k)}}{\OVec{u}_t} + \normsq{\OVec{u}_t} } \\
    &\hspace{-24pt}= \parenthesis{ \frac{1}{K} \sum_{k=1}^K \normsq{\IVec{u}{k}_t}} - 2 \parenthesis{ \frac{1}{K} \sum_{k=1}^K \dotp{\Vec{u}_t^{(k)}}{\OVec{u}_t}} + \parenthesis{ \frac{1}{K} \sum_{k=1}^K \normsq{\OVec{u}_t}} \\
    &\hspace{-24pt}= \parenthesis{ \frac{1}{K} \sum_{k=1}^K \normsq{\IVec{u}{k}_t}} - 2 \dotp{\parenthesis{ \frac{1}{K} \sum_{k=1}^K \Vec{u}_t^{(k)}}}{\OVec{u}_t} + \normsq{\OVec{u}_t} \\
    &\hspace{-24pt}= \parenthesis{ \frac{1}{K} \sum_{k=1}^K \normsq{\IVec{u}{k}_t}} - 2 \dotp{\OVec{u}_t}{\OVec{u}_t} + \normsq{\OVec{u}_t} \\
    &\hspace{-24pt}= \parenthesis{ \frac{1}{K} \sum_{k=1}^K \normsq{\IVec{u}{k}_t}} - 2 \normsq{\OVec{u}_t} + \normsq{\OVec{u}_t} \\
    &\hspace{-24pt}= \parenthesis{ \frac{1}{K} \sum_{k=1}^K \normsq{\IVec{u}{k}_t}} - \normsq{\OVec{u}_t}
\end{align*}
\end{proof}

\begin{algorithm}[t]
\caption{Federated Dynamic Averaging - \textsc{FDA}}
\label{alg:FDA}
\begin{algorithmic}[1]
\Statex \textbf{Require:} $K$: The number of workers indexed by $k$
\Statex \textbf{Require:} $\Theta$: The model variance threshold
\Statex \textbf{Require:} $b$: The local mini-batch size
\Statex
\State Initialize $\Vec{w}_0^{(k)} = \OVec{w}_0 \in \mathbb{R}^d$
\State \textbf{for} each step $t = 1, 2, \dots$ \textbf{do}
\State \Indent \textbf{for} each worker $k = 1, \dots, K$ \textbf{in parallel do}
\State  \Indent \Indent $\mathcal{B}_t^{(k)} \leftarrow$ (sample a batch of size $b$ from $\mathcal{D}_k$)
\State \Indent \Indent $\Vec{w}^{(k)}_{t} \leftarrow \textsc{Optimize}(\Vec{w}_{t-1}^{(k)} \, , \, \mathcal{B}_t^{(k)})$ 
\State \Indent \Indent Update $\Vec{S}_t^{(k)}$
\State \Indent \Indent $\OVec{S}_t \leftarrow \textsc{AllReduce}(\Vec{S}_t^{(k)})$
\State \Indent \Indent \textbf{if} $H(\OVec{S}_t) > \Theta$  \textbf{then}
\State \Indent \Indent \Indent $\Vec{w}_t^{(k)} \leftarrow \textsc{AllReduce}(\Vec{w}^{(k)}_{t})$ \Comment{In-place}
\end{algorithmic} 
\end{algorithm}

Conceptually, following Eq~\eqref{eq:variance_def}, to precisely monitor the variance, we need to calculate two quantities: (1) $\frac{1}{K} \sum_{k=1}^K \|\IVec{u}{k}_t\|_2^2$, and (2) $\|\OVec{u}_t\|_2^2$. The first quantity requires an $\textsc{AllReduce}$ operation on the squared norm of the worker drifts, which involves minimal overhead since these values are scalar. In contrast, the second quantity necessitates an $\textsc{AllReduce}$ operation on the worker drifts themselves, which are of model dimension, thus incurring a high communication cost. In fact, this operation is equivalent to synchronization, which is exactly what we aim to avoid in the first place. Thus, it becomes evident that communication-efficient model variance estimation hinges on estimating $\|\OVec{u}_t\|_2^2$ efficiently.

Upcoming sections will detail two techniques for communication efficient variance estimation (which primarily involves estimating $\|\OVec{u}_t\|_2^2$): \textsc{SketchFDA} and \textsc{LinearFDA}. To present them uniformly, we introduce the \emph{local state} $\Vec{S}_t^{(k)}$, a tensor which contains: (1) the scalar value $\|\IVec{u}{k}_t\|_2^2$ for precisely calculating the first quantity, and (2) a low-dimensional summary of $\Vec{u}_t^{(k)}$, different for each technique, for estimating the second quantity. For each technique we define an estimation function $H(\cdot)$ that calculates the current variance estimate from \emph{average state} $\OVec{S}_t = \frac{1}{K} \sum_{k=1}^K \Vec{S}_t^{(k)}$ (obtained via $\textsc{AllReduce}$).

\subsection{\textsc{SketchFDA}: Sketch-based Estimation}\label{section:Sketch}

An optimal estimator for $\normsq{\Dtm}$ can be obtained through the utilization and properties of AMS sketches, as detailed in~\cite{cormode2005sketch}. An AMS sketch of a vector $\Vec{v} \in \mathbb{R}^d$ is an $l \times m$ real matrix:
\begin{displaymath}
    \sk \parenthesis{\Vec{v}} = \begin{bmatrix}
        \psi_1 & \psi_2 & \dots & \psi_l
    \end{bmatrix}^\top \in \mathbb{R}^{l \times m} \; \; , \; \; l \cdot m \ll d
\end{displaymath}
An estimate for squared-norm $\normsq{\Vec{v}}$ is provided by the formula
\begin{displaymath}
    \mathcal{M}_2 \parenthesis{ \sk(\Vec{v}) } = \mathrm{median} \left \{ \normsq{ \psi_i} \; , \; i = 1, \dots, l \right \}
\end{displaymath}

\noindent The quality of estimation depends on the size of the sketch. For chosen $\epsilon, \delta > 0$,  where sketch dimensions are given by $l = \mathcal{O}\parenthesis{\log 1 / \delta}$ and $m = \mathcal{O}\parenthesis{1 / \epsilon^2}$, we have the following probabilistic guarantee:  with confidence at least $1-\delta$,
\[ \mathcal{M}_2 (\sk (\Vec{v})) \in (1\pm \epsilon) \normsq{\Vec{v}} \]
Notably, observe that the accuracy ($\epsilon$) and confidence ($1-\delta$) only depend on the size of the sketch and not on the dimensionality of vector $\Vec{v}$.

Two crucial properties of the AMS sketch are that (a) it is a linear transformation, i.e., for $\alpha_1, \alpha_2 \in \mathbb{R}$ and $\Vec{v}_1, \Vec{v}_2 \in \mathbb{R} ^d$, 
\[
    \sk( \alpha_1 \Vec{v}_1 + \alpha_2 \Vec{v}_2 ) = 
    \alpha_1 \sk( \Vec{v}_1 )
    + \alpha_2 \sk(\Vec{v}_2 )
\]
and (b) can be computed efficiently in time $O(l \cdot d)$. 

In the \textsc{SketchFDA} approach, the salient idea is to employ AMS sketches
$\sk (\Dt{k}) \in \mathbb{R}^{l \times m}$
as a low-dimensional representation of the local drifts $\Dt{k}$.

\begin{theorem} Let $l = \mathcal{O}(\log \frac{1}{\delta})$ and $m = \mathcal{O}(\frac{1}{ \epsilon^2})$. Define the local state as
\begin{displaymath}
    \Vec{S}_t^{(k)} = \parenthesis{ \normsq{\Dt{k}} \, , \, \sk \parenthesis{\Dt{k}} } \in \mathbb{R} \times \mathbb{R}^{l \times m}
\end{displaymath}
and the approximation function as
\begin{displaymath}
    H\parenthesis{\OVec{S}_t } = \frac{1}{K}\sum_k \normsq{\Vec{u}_t^{(k)}} - \frac{1}{1+\epsilon} \mathcal{M}_2 \parenthesis{ \frac{1}{K} \sum_{k=1}^K  \sk \parenthesis{\Dt{k}} }.
\end{displaymath}
Then, the condition $H\large(\OVec{S}_t\large) \leq \Theta$ implies $\Var{\Vec{w}_t} \leq \Theta$ with probability at least ($1-\delta$).
\end{theorem}

\begin{proof}
\begin{align*}
    H\parenthesis{\OVec{S}_t} &= \frac{1}{K}\sum_{k=1}^K \normsq{\Vec{u}_t^{(k)}} - \frac{1}{1+\epsilon} \mathcal{M}_2 \parenthesis{ \frac{1}{K} \sum_{i=1}^K \sk \parenthesis{\Dt{k}} } \\
    \overset{\text{(lin.)}}&{=} \frac{1}{K}\sum_{k=1}^K \normsq{\Vec{u}_t^{(k)}} - \frac{1}{1+\epsilon} \mathcal{M}_2 \parenthesis{\sk \parenthesis{ \frac{1}{K} \sum_{i=1}^K \Dt{k}} } \\
    &= \frac{1}{K}\sum_{k=1}^K \normsq{\Vec{u}_t^{(k)}} - \frac{1}{1+\epsilon} \mathcal{M}_2 \parenthesis{\sk \parenthesis{ \OVec{u}_t } } \\
    \overset{\text{($\epsilon$-err.)}}&{\geq} \frac{1}{K}\sum_{k=1}^K \normsq{\Vec{u}_t^{(k)}} - \norm{\OVec{u}_t}^2 \quad \text{with prob. at least ($1-\delta$)} \\
    &= \Var{\Vec{w}_t}
\end{align*}
We proved that $H(\OVec{S}_t) \geq \Var{\Vec{w}_t}$ with probability at least ($1-\delta$), i.e., we overestimate the model variance with probability at least ($1-\delta$), completing the proof.
\end{proof}

In Section~\ref{section:discussion}, we discuss the empirical basis for choosing the values of $l$ and $m$, and how they practically impact the quality of the sketch approximation.

\subsection{\textsc{LinearFDA}: Linear Approximation}\label{section:Linear}

Although AMS sketches provide good estimates for variance, their dimension is  in the several hundreds, and the communication cost of $\textsc{AllReduce}$
on sketches, performed at each step, may be non-negligible. 
Therefore, we also introduce a low-cost, ad-hoc estimation
variant.

In this approach, 
instead of an AMS sketch, each local state contains the scalar value $\dotpN{\xi}{\Dt{k}} \in \mathbb{R}$, where $\xi \in \mathbb{R}^d$ is a unit vector, known to all workers.

\begin{theorem}
Define the local state as 
\begin{displaymath}
    \Vec{S}_t^{(k)} = \parenthesis{\normsq{\Dt{k}} \, , \, \dotp{\xi}{\Dt{k}}} \in \mathbb{R} \times \mathbb{R} \; \; , \; \; \left\Vert \xi \right\Vert_2 = 1
\end{displaymath}
and the approximation function as
\begin{displaymath}
    H\parenthesis{\OVec{S}_t} = \frac{1}{K}\sum_{k=1}^K \normsq{\Vec{u}_t^{(k)}} - \abs{\frac{1}{K} \sum_{i=1}^K \dotp{\xi}{\Dt{k}} }^2
\end{displaymath}
Then, the condition $H\large(\OVec{S}_t\large) \leq \Theta$ implies $\Var{\Vec{w}_t} \leq \Theta$.
\end{theorem}
\begin{proof}
\begin{align*}
    H\parenthesis{\OVec{S}_t} &= \frac{1}{K}\sum_{k=1}^K \normsq{\Vec{u}_t^{(k)}} - \abs{\frac{1}{K} \sum_{i=1}^K \dotp{\xi}{\Dt{k}} }^2 \\
    &= \frac{1}{K}\sum_{k=1}^K \normsq{\Vec{u}_t^{(k)}} - 
    \abs{\dotp{\xi}{\frac{1}{K} \sum_{i=1}^K \Dt{k}} }^2 \\
    &= \frac{1}{K}\sum_{k=1}^K \normsq{\Vec{u}_t^{(k)}} - \abs{\dotp{\xi}{\OVec{u}_t}}^2 \\
    &\geq  \frac{1}{K} \sum_{k=1}^K \normsq{\Dt{k}} - \norm{\xi}^2 \norm{\OVec{u}_t}^2 \\
    &= \frac{1}{K} \sum_{k=1}^K \normsq{\Dt{k}} - \norm{\OVec{u}_t}^2 \\
    &= \Var{\Vec{w}_t}
\end{align*}
We proved that $H(\OVec{S}_t) \geq \Var{\Vec{w}_t}$, i.e., we always overestimate the model variance, completing the proof.
\end{proof}

An arbitrary choice of $\xi$ (e.g., a random vector) is likely to estimate 
$\norm{\OVec{u}_t}^2$ poorly; if $\Vec{\xi}$ is uncorrelated to $\Dtm$, then $\abs{\dotp{\xi}{\OVec{u}_t}}^2$ will likely be close to zero.
A heuristic choice that might be correlated to $\Dtm$ is the 
(normalized) value of $\OVec{u}_{t_0}$, the global drift vector right at the time of last synchronization. All nodes can compute it independently without extra communication, if they take the difference of the models of the last two synchronizations:
 \begin{align*}
     \Vec{\xi} = 
     \frac{\OVec{u}_{t_0}}{\norm{\OVec{u}_{t_0}}}
     = \frac{\OVec{w}_{t_0} - \OVec{w}_{t_{-1}}}{ \norm{\OVec{w}_{t_0} - \OVec{w}_{t_{-1}}} }
 \end{align*}
 



\subsection{Discussion}\label{section:discussion}

\vspace{1mm}

\noindent \textbf{\textsc{FDA}: Intuition.} The main intuition for FDA is summarized in making the decision to synchronize dynamic, based on \emph{model variance} during training. This metric is designed to capture the collective state of the training process. In what follows, we provide intuition on why this is the case. It is important to remember that the global model $\OVec{w}_t$ and, by extension, the global drift $\OVec{u}_t$, are ultimately what we care about and evaluate.

Model variance, as defined in  Equation~\eqref{eq:variance_def}, is the difference between the average of the squared local drifts $\frac{1}{K} \sum \|\Vec{u}_t^{(k)}\|_2^2$ and the squared global drift $\normsq{\OVec{u}_t}$. The first term reflects how far the individual worker models have moved--essentially, how much each worker has learned. The second term indicates how much of this learning is retained in the global model after aggregation. 

The interplay between these two quantities is crucial. For example, when the local drifts are high but the global drift is low, the variance increases, signaling the need for synchronization. This scenario suggests that while individual workers have made significant progress (as indicated by high local drifts), this progress is not being effectively captured in the global model (indicated by the low global drift). In other words, the worker models have moved significantly, but the global model has remained relatively stationary in this high-dimensional space. This misalignment indicates that training is no longer progressing optimally, as the workers are moving towards disparate and conflicting local minima, making it crucial to synchronize and realign them. Conversely, when both the local and global drifts are either low or high, synchronization is not necessary, and the variance naturally remains low.

Neither the average of the local drifts nor the global drift alone provides a complete picture of the collective training progress. Relying solely on one or the other would lead to suboptimal synchronization decisions and likely prove ineffective. In \textsc{FDA}, it is the relationship between these quantities, as captured by the model variance, that offers valuable insights and guides the crucial decision of when to synchronize.

\begin{figure}[t]
    \centering
    \begin{subfigure}[b]{0.49\textwidth}
        \centering
        \includegraphics[width=0.9\textwidth]{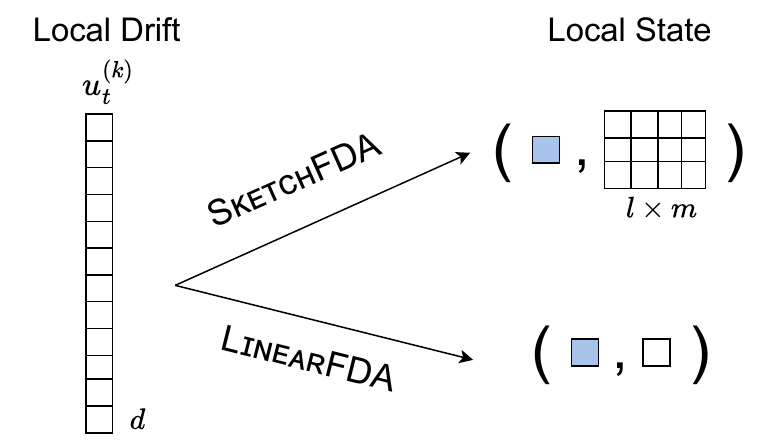}
    \end{subfigure}
    \caption{
    \textsc{SketchFDA} \& \textsc{LinearFDA}: 
    Local State structure.}
    \label{fig:sketch_linear_illustration}
\end{figure}

\vspace{1mm}
\noindent {\textbf{\textsc{SketchFDA} vs. \textsc{LinearFDA}}}: 
Both methods send the squared norm of the drift $\|\Vec{u}_t^{(k)}\|_2^2$, but differ in the additional accompanying lower-dimensional representation they transmit (Figure \ref{fig:sketch_linear_illustration}):
\begin{enumerate}[leftmargin=20pt]
    \item \textsc{SketchFDA}: An AMS sketch of the local drift.
    
    \item \textsc{LinearFDA}: The dot product of a vector and the local drift.
\end{enumerate}

\noindent The key difference between these two variants lies in the fidelity of approximation of the model variance. While both methods conservatively overestimate the variance, \textsc{SketchFDA} provides a provably accurate estimation, which is expected to lead to fewer synchronizations.  \textsc{LinearFDA} requires less computational effort and bandwidth to create and communicate the local states, but may overestimate variance by too much, causing unnecessary synchronizations.

\vspace{1mm}
\noindent \textbf{\textsc{SketchFDA}: Choice of $l$ and $m$}. We empirically
measured the approximation achieved with sketch dimensions of $l=5$ rows and $m=250$ columns (as defined in Section~\ref{section:Sketch}): these settings yield an error bound of $\epsilon \approx 6\%$ and a probabilistic confidence of $(1-\delta) \approx 95\%$. Based on our experiments, we have adopted these values in our experiments and recommend them. Using these values, the byte-size of a sketch is $l \cdot m \cdot 4\,\mathrm{bytes} = 5\,\mathrm{kB}$, significantly smaller than the size of all our models.  Sketches of smaller size could be used, albeit 
weakening the approximation of the variance. However, given that \textsc{LinearFDA}
similarly weakens approximation and avoids using AMS sketches, in the interest of space we do not explore varying AMS sketch sizes in this paper.

\vspace{1mm}
\noindent \textbf{\textsc{FDA}: Asynchronous Operation.} As mentioned in Section~\ref{section:rel_work}, FDA can be readily modified to operate asynchronously. In this setup, one worker-node acts as a coordinator, aggregating local states and determining whether synchronization is needed each time a local state is received. This decision is based on the most recent local states from all workers. It is important to note that, since local states are small in size, asynchronous operation is unlikely to alleviate bandwidth issues. The primary advantage is that it allows training to continue even in the presence of stragglers. Asynchronous operation might also be beneficial in rare cases where the overhead of initializing communication dominates the actual transmission time.  

\section{Experiments}\label{section:exps}

\subsection{Setup}

Table \ref{tab:summary_experiments} provides a comprehensive overview of our experiments. For each experiment, we detail the Neural Network (NN) architecture, its parameter count ($d$), and the dataset used for training. The table also specifies key hyper-parameters: the batch size ($b$), the number of workers ($K$), and the \textsc{FDA}-specific variance threshold ($\Theta$). Additionally, we indicate the chosen optimizer (as detailed in Section~\ref{section:FDA}) and the training algorithms employed for each configuration.

\vspace{1.1mm}
\noindent \textbf{Platform.} We employ TensorFlow~\cite{abadi2015tensorflow}, integrated with Keras~\cite{chollet2015keras}, as the platform for conducting our experiments. We used TensorFlow to implement our FDA variants and all competitive algorithms. All relevant code, figures, and data of this study are available in \url{https://github.com/miketheologitis/FedL-Sync-FDA}.

\vspace{1.1mm}
\noindent \textbf{Hardware \& Infrastructure.} We conducted our experiments on the ARIS High performance computing (HPC) environment\footnote{\url{https://www.hpc.grnet.gr/en/hardware-2/}}, utilizing a cluster of 44 GPU-accelerated worker-nodes. Each worker is equipped with two NVIDIA Tesla K40m GPUs and interconnected via an InfiniBand FDR14 network, providing up to 56 GB/s of bandwidth. Crucially, our evaluation remains agnostic to the underlying infrastructure of the specific workers.

\begin{table*}[t]
  \caption{Summary of Experiments}
  \label{tab:summary_experiments}
  \begin{adjustbox}{width=\textwidth,center}
  \begin{tabular}{|c|c|c|c|c|c|c|c|}
    \cline{4-8}
    \multicolumn{3}{c|}{} & \multicolumn{3}{c|}{Hyper-Parameters} & \multicolumn{2}{c|}{Training} \\
    \hline
    NN & d & Dataset & \(\Theta\) & b & K & Optimizer & Algorithms \\
    \hline
    LeNet-5 & 62K & MNIST & $\{0.5, 1, 1.5, 2, 3, 5, 7\}$ & 32 & \{ 5, 10, \dots, 60 \} & Adam & \textsc{FDA}, \textsc{Synchronous}, \textsc{FedAdam} \\
    VGG16* &  2.6M & MNIST & $\{ 20, 25, 30, 50, 75, 90, 100 \}$ & 32 & \{ 5, 10, \dots, 60 \} & Adam & \textsc{FDA}, \textsc{Synchronous}, \textsc{FedAdam} \\
    DenseNet121 & 6.9M & CIFAR-10 & $\{ 200, 250, 275, 300, 325, 350, 400 \}$ & 32 & \{ 5, 10, \dots, 30 \} & SGD-NM & \textsc{FDA}, \textsc{Synchronous}, \textsc{FedAvgM} \\
    DenseNet201 & 18M & CIFAR-10 & $\{ 350, 500, 600, 700, 800, 850, 900 \}$ & 32 & \{ 5, 10, \dots, 30 \} & SGD-NM & \textsc{FDA}, \textsc{Synchronous}, \textsc{FedAvgM} \\
    \hline
    \begin{tabular}{@{}c@{}} {\small (fine-tuning)} \\ ConvNeXtLarge \end{tabular} & 198M & CIFAR-100 & $\{ 25, 50, 100, 150\}$ & 32 & \{ 3, 5 \} & AdamW & \textsc{FDA}, \textsc{Synchronous} \\
    \hline
  \end{tabular}%
  \end{adjustbox}
\end{table*}

\vspace{1.1mm}
\noindent \textbf{Datasets \& Models.} The core experiments involve training Convolutional Neural Networks (CNNs) of varying sizes and complexities on two datasets: MNIST~\cite{deng2012mnist} and CIFAR-10~\cite{krizhevsky2012cifar10}. For the MNIST dataset, we employ LeNet-5~\cite{lecun1998lenet}, composed of approximately 62 thousand parameters, and a modified version of VGG16~\cite{simonyan2014vgg}, denoted as VGG16*, consisting of 2.6 million parameters. VGG16* was specifically adapted for the MNIST dataset, a less demanding learning problem compared to ImageNet~\cite{russakovsky2015imagenet}, for which VGG16 was designed. In VGG16*, we omitted the $512$-channel convolutional blocks and downscaled the final two fully connected (FC) layers from $4096$ to $512$ units each. Both models use Glorot uniform initialization~\cite{glorot2010uniform}. For CIFAR-10, we utilize DenseNet121 and DenseNet201~\cite{huang2016densenet}, as implemented in Keras~\cite{chollet2015keras}, with the addition of dropout regularization layers at rate 0.2 and weight decay of $10^{-4}$, as prescribed in~\cite{huang2016densenet}. The DenseNet121 and DenseNet201 models have 6.9 million and 18 million parameters, respectively, and are both initialized with He normal~\cite{he2015normal}. 

Lastly, we explore a transfer learning scenario on the dataset \mbox{CIFAR-100}~\cite{krizhevsky2012cifar10}, a choice reflecting the DL community's growing preference of using pre-trained models in such downstream tasks~\cite{han2021ptm}. For example, a pre-trained visual transformer (ViT) on ImageNet, transferred to classify CIFAR-100, is currently on par with the state-of-the-art results for this task~\cite{dosovitskiy2020vit}. We adopt this exact transfer learning scenario, leveraging the more powerful \mbox{ConvNeXtLarge} model, pre-trained on ImageNet, with $198$ million parameters~\cite{liu2022convenet, chollet2015keras}. Following the feature extraction step~\cite{DeepLearningGoodfellow}, the testing accuracy on CIFAR-100 stands at $60\%$. Subsequently, we employ and evaluate our \textsc{FDA} algorithms in the arduous fine-tuning stage, where the entirety of the model is trained~\cite{sigmod2022nautilus_tfl}.

\vspace{1.1mm}
\noindent \textbf{Algorithms.} We consider five distributed deep learning algorithms: \textsc{LinearFDA}, \textsc{SketchFDA}, \textsc{Synchronous} \footnote{The name was derived from the Bulk Synchronous Parallel approach; can be understood as a special case of the \textsc{FDA} Algorithm~\ref{alg:FDA} where $\Theta$ is set to zero.}, \textsc{FedAdam}~\cite{reddi2021fedadam}, and \textsc{FedAvgM}~\cite{hsu2019fedAvgM}; the first three are standard in all experiments. Depending on the local optimizer, Adam~\cite{kingma2017adam} or SGD with Nesterov momentum (SGD-NM)~\cite{suts2013nesterov_mom}, we also include their communication-efficient federated counterparts \textsc{FedAdam} or \textsc{FedAvgM}, respectively.

\vspace{1.1mm}
\noindent \textbf{Evaluation Methodology.} Comparing DDL algorithms is not straightforward.
For example, comparing DDL algorithms based on the average cost of a training epoch can be misleading, as it does not consider the effects on the trained model's quality.
To achieve a comprehensive performance assessment of \textsc{FDA}, 
we define a \emph{training run} as the process of executing the DDL algorithm under evaluation, on (a) a specific DL model and
training dataset, and (b) until a final epoch in which the trained model achieves a specific \emph{testing accuracy} (termed as \textit{Accuracy Target} in figures).
Based on this definition, we focus on two performance metrics:
\begin{enumerate}[leftmargin=20pt]

    \item \textbf{Communication cost}, which is the total data (in bytes) transmitted by all workers. Notably, communication cost is unaffected by the training data volume since only model updates (when synchronizing) and local states (at each step), but not training data, are transmitted. Thus, the communication cost mainly depends on the complexity (number of parameters) of the used model. Translating the communication cost to \emph{wall-clock time} (i.e., the total time required for the computation and communication of the DDL) depends on the network infrastructure connecting the workers and on the overhead of establishing and initializing communication. Its impact is larger in FL scenarios, where workers often use slower Wi-Fi connections.
    
    \item \textbf{Computation cost}, which is the number of mini-batch steps (termed as \textit{In-Parallel Learning Steps} in figures) performed by each worker. Translating this cost to \emph{wall-clock time} is determined by the mini-batch size and the computational resources of the worker-nodes. Its impact is larger for workers with lower computational resources.
\end{enumerate}

\vspace{1.1mm}
\noindent \textbf{Hyper-Parameters \& Optimizers.} Hyper-parameters unique to each training dataset and model are detailed in Table~\ref{tab:summary_experiments}; $\Theta$ is pertinent to \textsc{FDA} algorithms and not applicable to others. Notably, a guideline for setting the parameter $\Theta$ is provided in Section~\ref{section:Results}. For experiments involving \textsc{FedAvgM} and \textsc{FedAdam}, we use $E = 1$ local epochs, following~\cite{reddi2021fedadam}. For experiments with LeNet-5 and VGG16*, local optimization employs Adam, using the default settings as per~\cite{kingma2017adam}. In these cases, \textsc{FedAdam} also adheres to the default settings for both local and server optimization~\cite{reddi2021fedadam, chollet2015keras}. For DenseNet121 and DenseNet201, local optimization is performed using SGD with Nesterov momentum (SGD-NM), setting the momentum parameter at $0.9$ and learning rate at $0.1$~\cite{huang2016densenet}. For \textsc{FedAvgM}, local optimization is conducted with default settings~\cite{hsu2019fedAvgM, chollet2015keras}, while server optimization employs SGD with momentum, setting the momentum parameter and learning rate to $0.9$ and $0.316$, respectively~\cite{reddi2021fedadam}. Lastly, for the transfer learning experiments, local optimization leverages AdamW~\cite{losh2019adamw}, with the hyper-parameters used for fine-tuning ConvNeXtLarge in the original study~\cite{liu2022convenet}.

\vspace{1.1mm}
\noindent \textbf{Data Distribution.} In all experiments, the training dataset is divided into approximately equal parts among the workers. To assess the impact of data heterogeneity, we explore three scenarios:
\begin{enumerate}[leftmargin=20pt]

    \item \textbf{IID} --- Independent and identically distributed.
    
    \item \textbf{Non-IID:} $X$\textbf{\%} --- A portion $X\%$ of the dataset is sorted by label and sequentially allocated to workers, with the remainder distributed in an IID fashion.
    
    \item \textbf{Non-IID: Label} $Y$ --- All samples from label $Y$ are assigned to a few workers, while the rest are distributed in an IID manner.
\end{enumerate}

\begin{figure}[t]
    \centering

    \begin{subfigure}[b]{0.49\textwidth}
        \centering
        \includegraphics[width=0.851\textwidth]{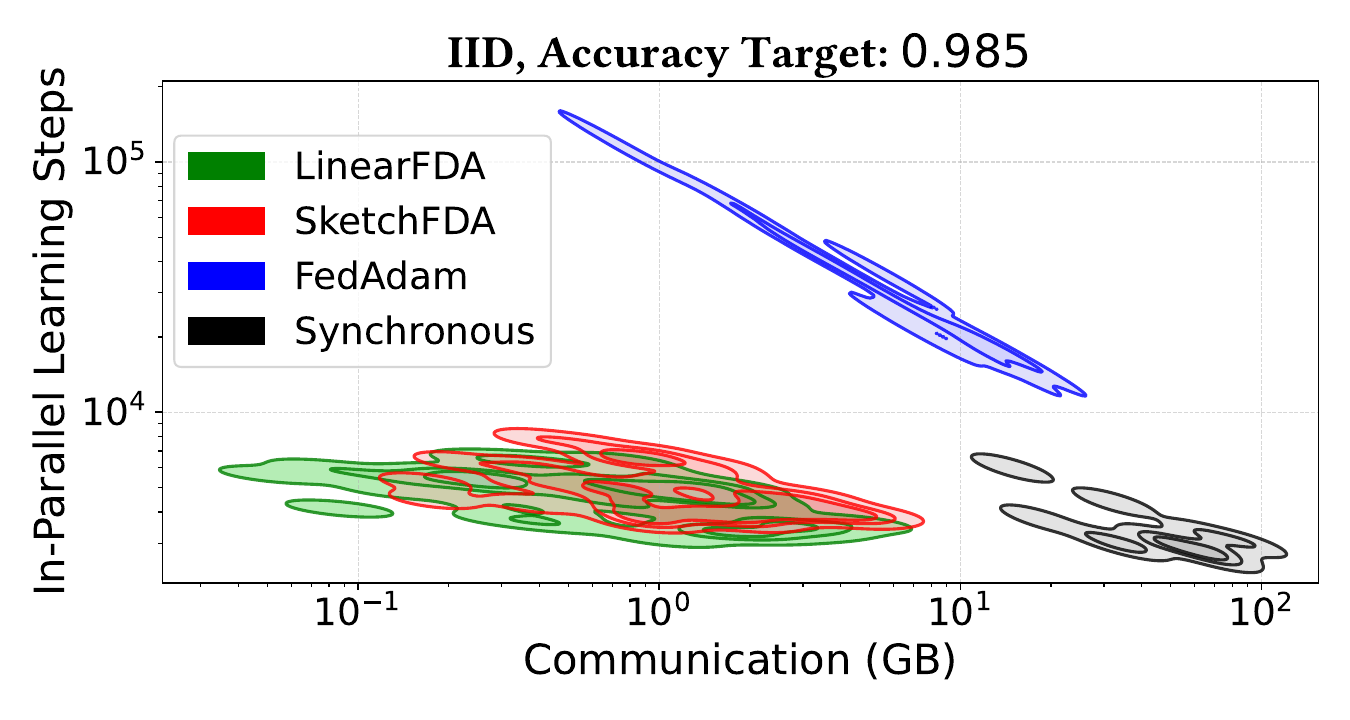}
    \end{subfigure}
    \hfill
    \begin{subfigure}[b]{0.49\textwidth}
        \centering
        \includegraphics[width=0.851\textwidth]{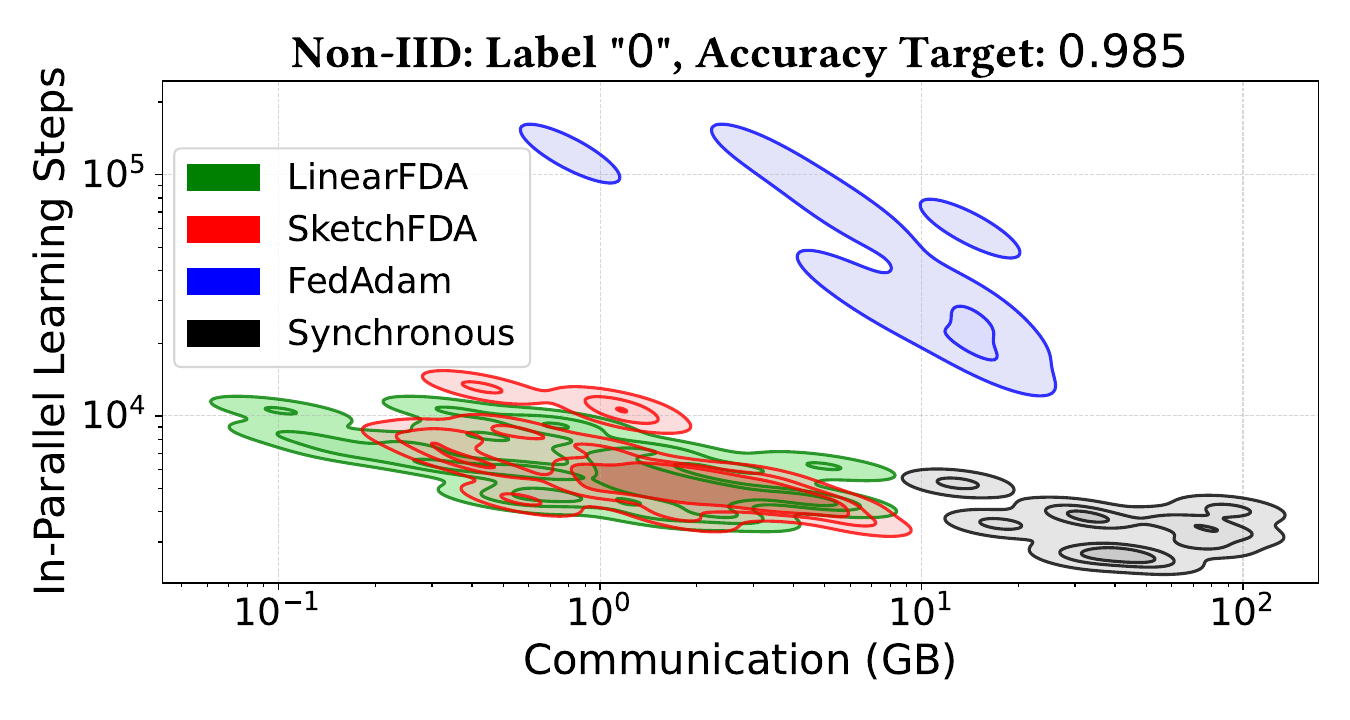}
    \end{subfigure}
    \hfill
    \begin{subfigure}[b]{0.49\textwidth}
        \centering
        \includegraphics[width=0.851\textwidth]{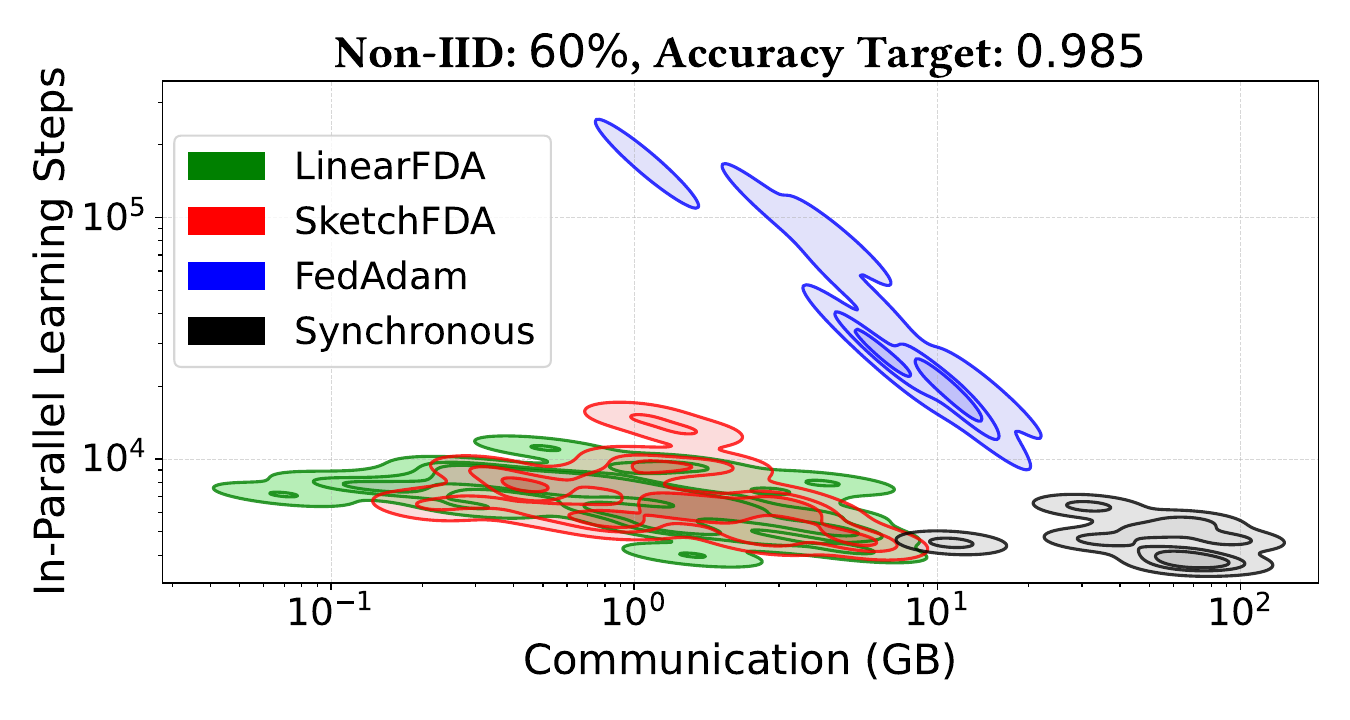}
    \end{subfigure}
    \caption{LeNet-5 on MNIST. At Non-IID: Label "0", the samples of Label "0" are assigned to few workers. At Non-IID: 60\%, 60\% of the dataset is sorted and allocated to workers, causing some workers to receive many samples from the same label}
    \label{fig:lenet5_joint_kde}
\end{figure}

\begin{figure*}[t]
    \centering
    \begin{subfigure}[b]{0.49\textwidth}
        \centering
        \includegraphics[width=0.851\textwidth]{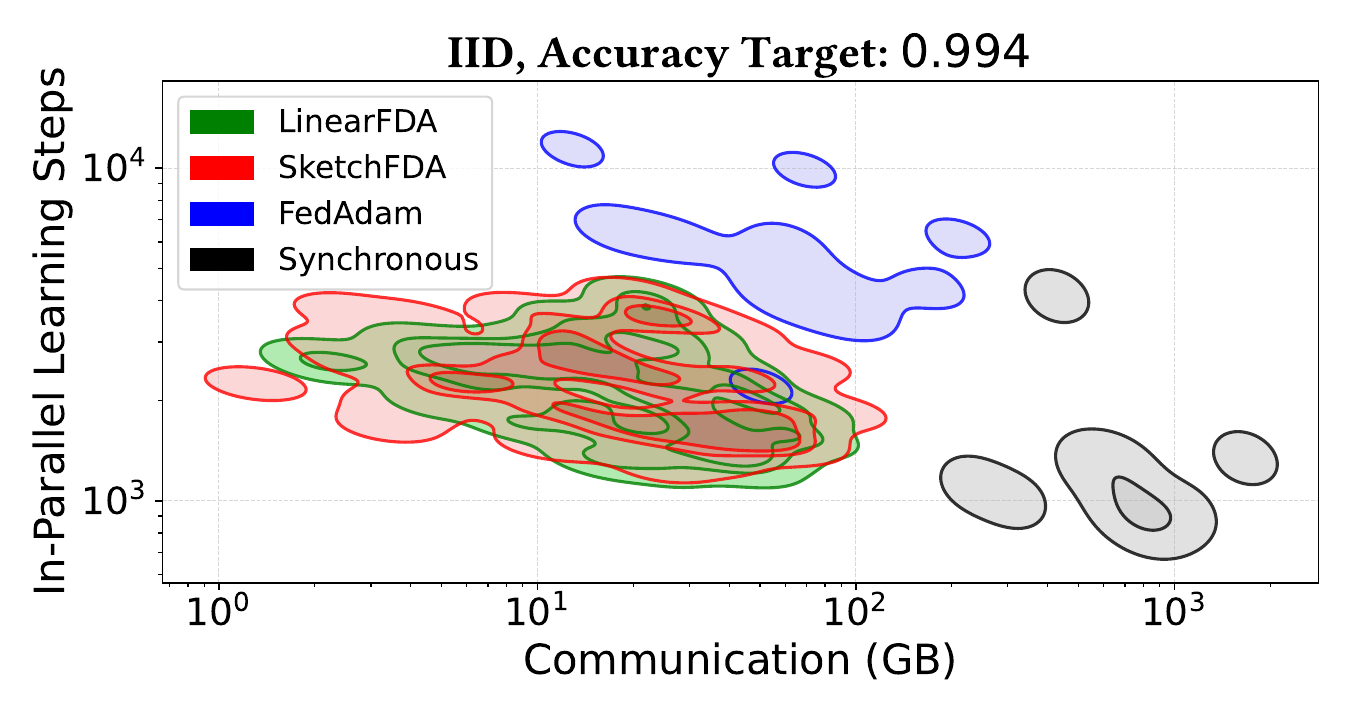}
    \end{subfigure}
    \hfill
    \begin{subfigure}[b]{0.49\textwidth}
        \centering
        \includegraphics[width=0.851\textwidth]{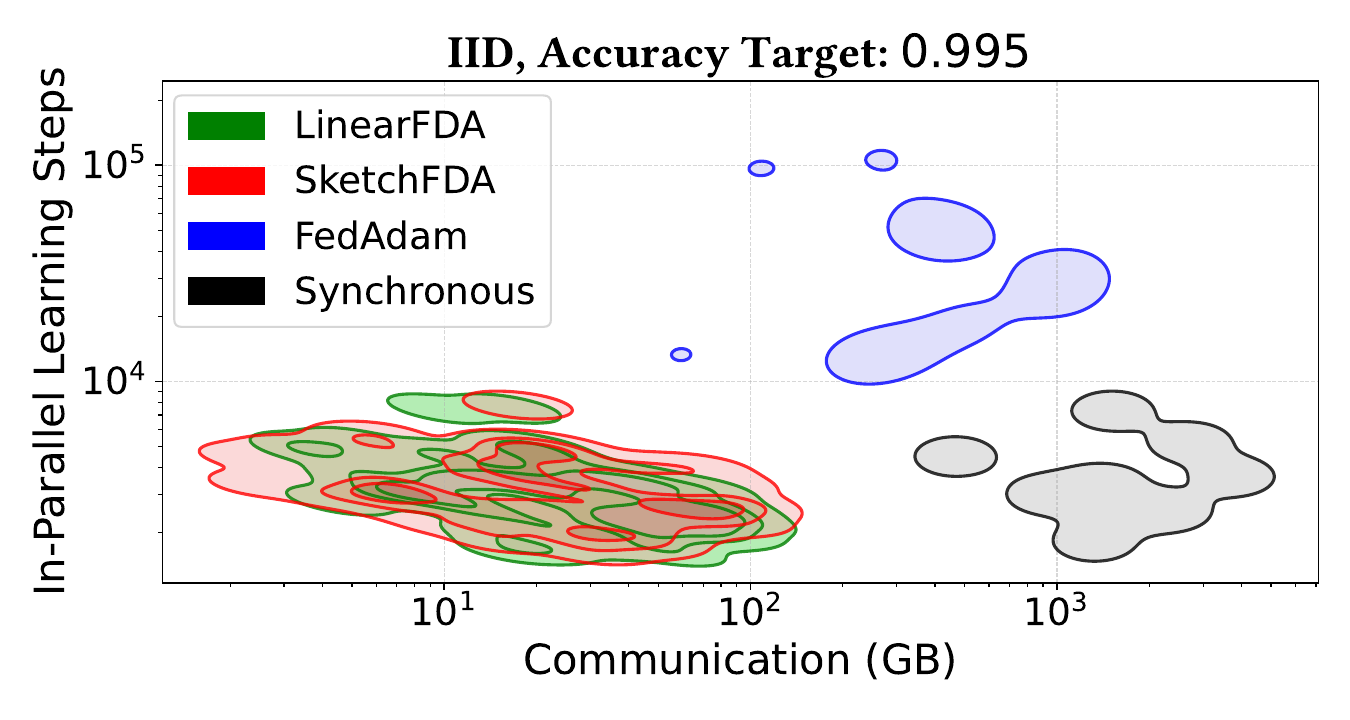}
    \end{subfigure}

    \begin{subfigure}[b]{0.49\textwidth}
        \centering
        \includegraphics[width=0.851\textwidth]{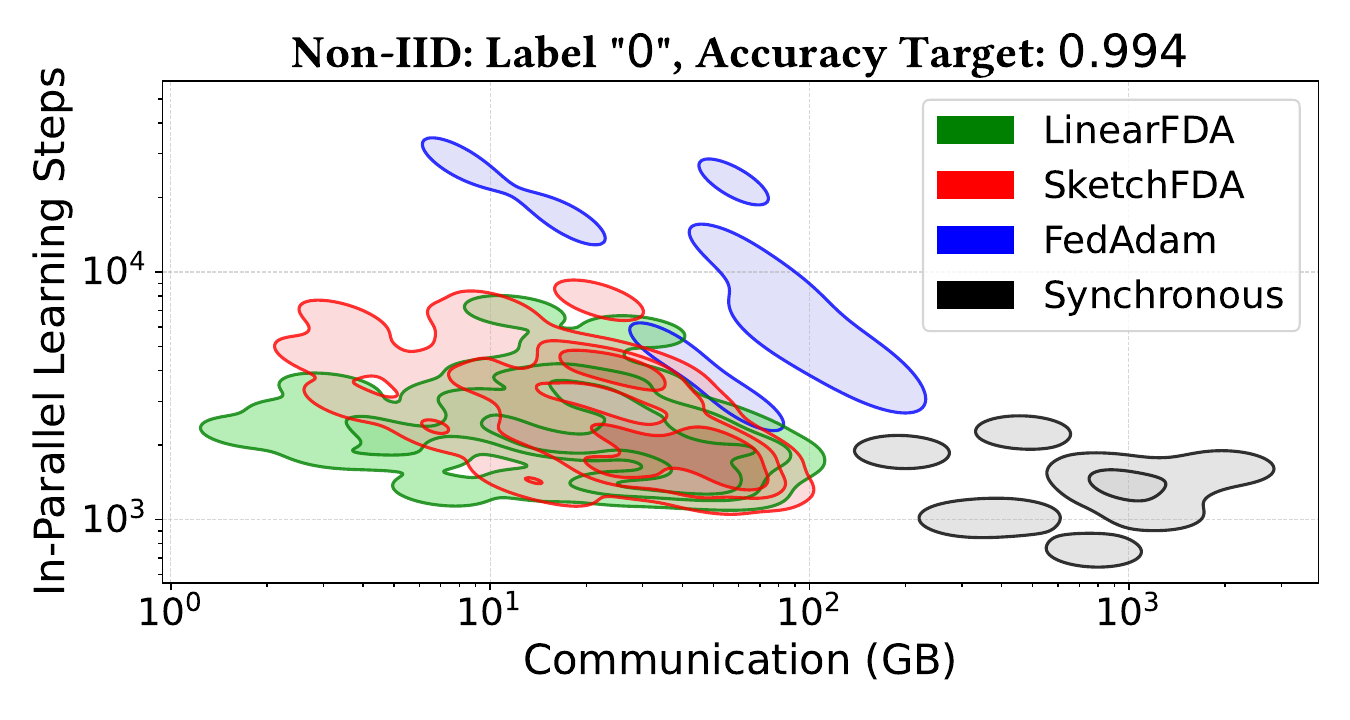}
    \end{subfigure}
    \hfill
    \begin{subfigure}[b]{0.49\textwidth}
        \centering
        \includegraphics[width=0.851\textwidth]{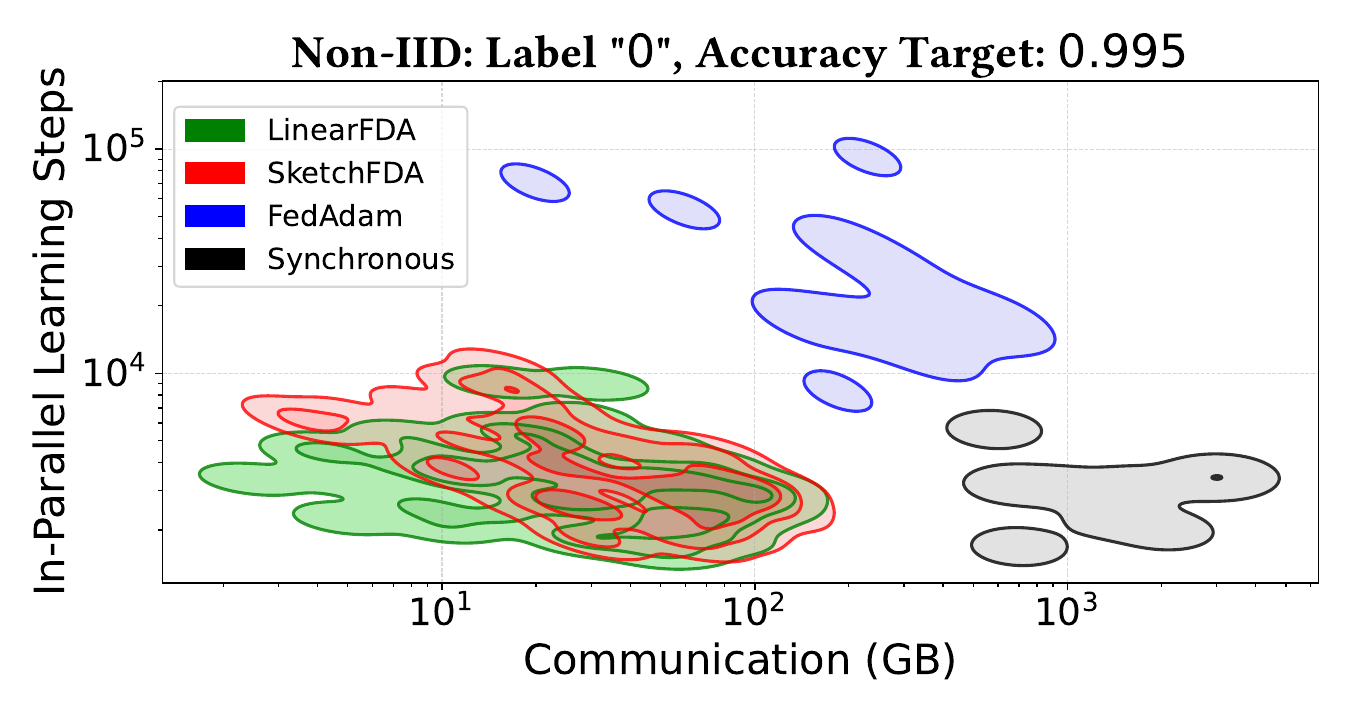}
    \end{subfigure}

    \begin{subfigure}[b]{0.49\textwidth}
        \centering
        \includegraphics[width=0.851\textwidth]{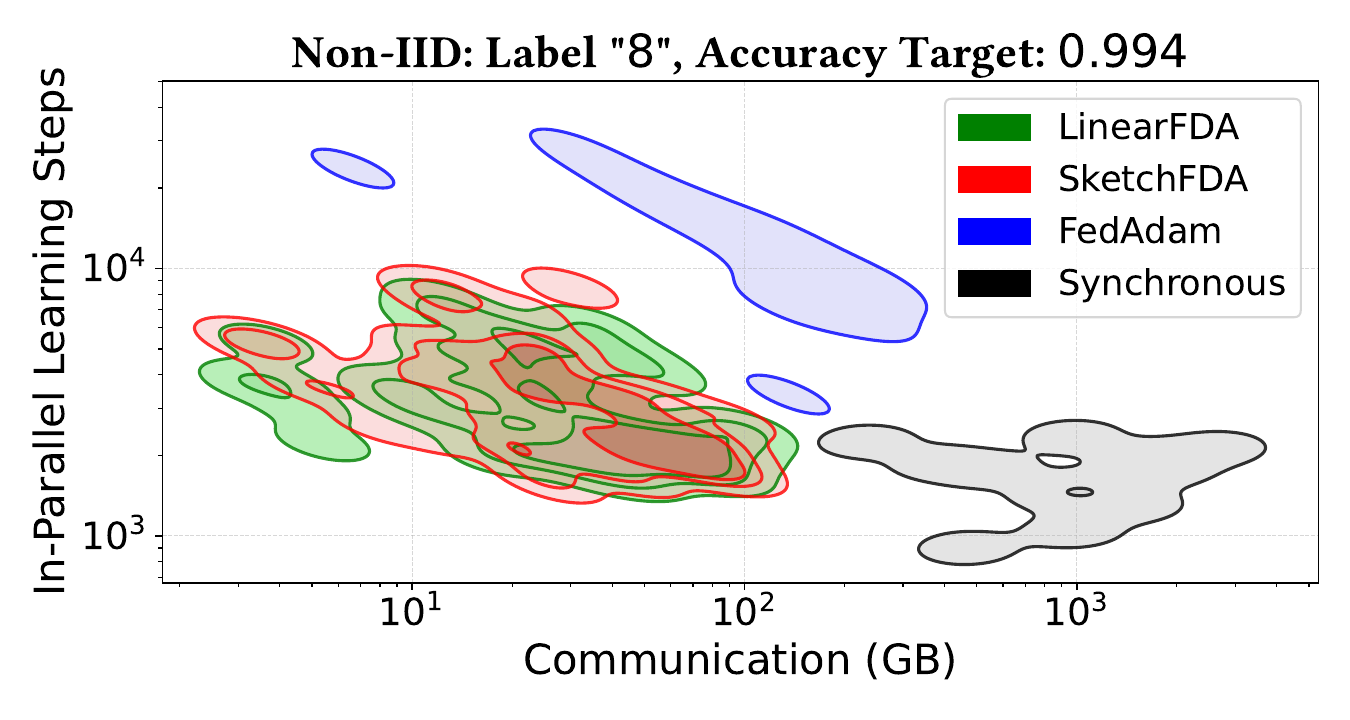}
    \end{subfigure}
    \hfill
    \begin{subfigure}[b]{0.49\textwidth}
        \centering
        \includegraphics[width=0.851\textwidth]{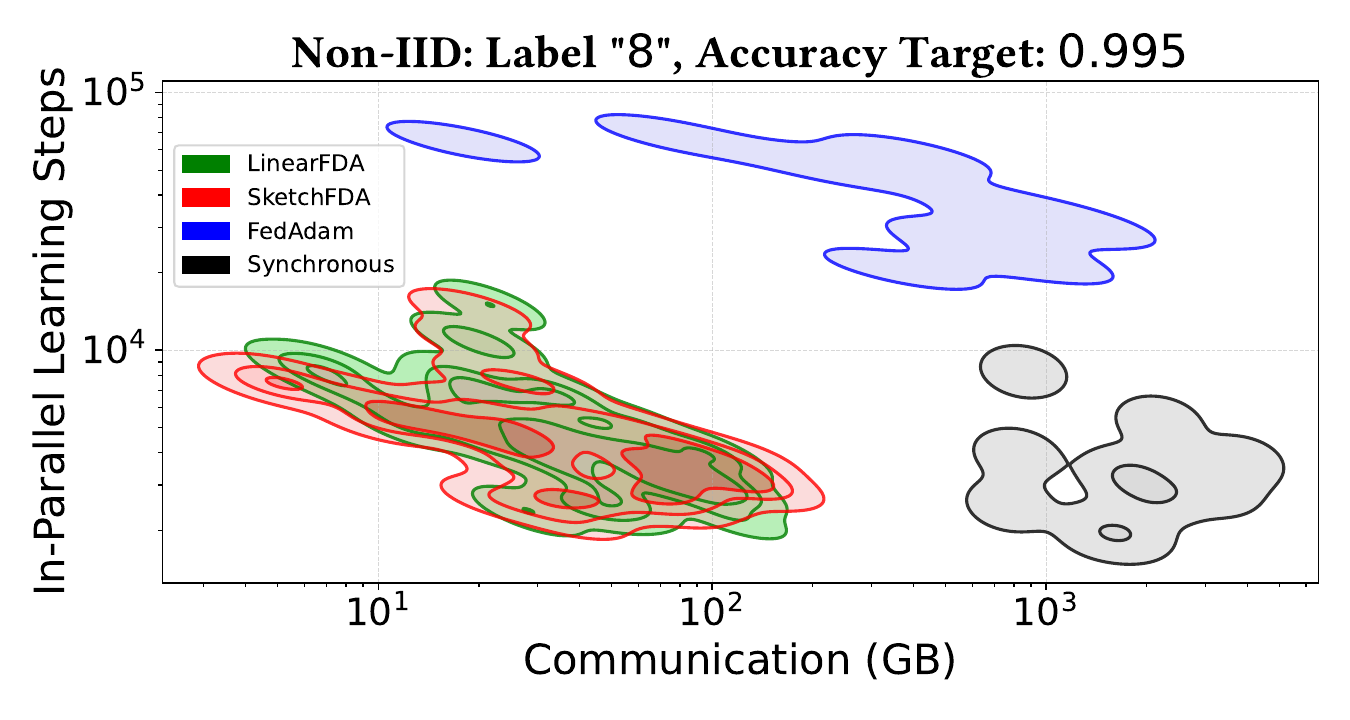}
    \end{subfigure}

    \caption{VGG16* on MNIST}
    \label{fig:vgg16_joint_kde}
\end{figure*}

\subsection{Main Findings} 
The main findings of our experimental analyses are:
\begin{enumerate}[leftmargin=20pt]

    \item \textsc{LinearFDA} and \textsc{SketchFDA} outperform the \textsc{Synchronous},   \textsc{FedAdam} and \textsc{FedAvgM} techniques (their use depends on the local optimizer choice) by 1-2 orders of magnitude in communication,
while maintaining equivalent model performance. 

    \item \textsc{LinearFDA} and \textsc{SketchFDA} also significantly outperform the \textsc{FedAdam} and \textsc{FedAvgM} techniques in terms of computation.
    
    \item The performance of \textsc{LinearFDA} and \textsc{SketchFDA} is comparable in most experiments. \textsc{SketchFDA} provides a more accurate estimator of the variance and leads to fewer synchronizations than \textsc{LinearFDA}, but has a larger communication overhead for its local state (a sketch, compared to two numbers). \textsc{SketchFDA} significantly outperforms \textsc{LinearFDA} at the transfer learning scenario.
    
    \item The FDA variants remain robust at various data
    heterogeneity settings, maintaining comparable performance to
    the IID case.
\end{enumerate}

\subsection{Results}\label{section:Results}
Due to the extensive set of unique experiments (over $1000$), as detailed in Table~\ref{tab:summary_experiments}, we leverage Kernel Density Estimation (KDE) plots~\cite{waskom2021seaborn} to visualize the bivariate distribution of computation and communication costs incurred by each strategy for attaining the \emph{Accuracy Target}. These KDE plots provide a high-level overview of the cost trade-off for training accurate models. The varying levels of opacity in the filled areas of the KDE plots represent the density of the underlying data points: higher opacity indicates areas with a greater concentration of data, whereas lower opacity signifies less dense areas.

As an illustrative example, Figure~\ref{fig:lenet5_joint_kde} depicts the strategies' bivariate distribution for the LeNet-5 model trained on MNIST with different data heterogeneity setups. In these plots, the \textsc{SketchFDA} distribution is generated from experiments across all hyper-parameter combinations ($\Theta$ and $K$ in Table~\ref{tab:summary_experiments}) that attained the \emph{Accuracy Target} of $0.985$. The observed high variance in the method's distribution stems from the varying $K$ and $\Theta$ values. In subsequent subsections, we elucidate how these hyper-parameters influence the communication and computation costs.


\begin{figure*}[t]
    \centering

    \begin{minipage}[b]{0.49\textwidth}
        \centering
        \includegraphics[width=0.851\textwidth]{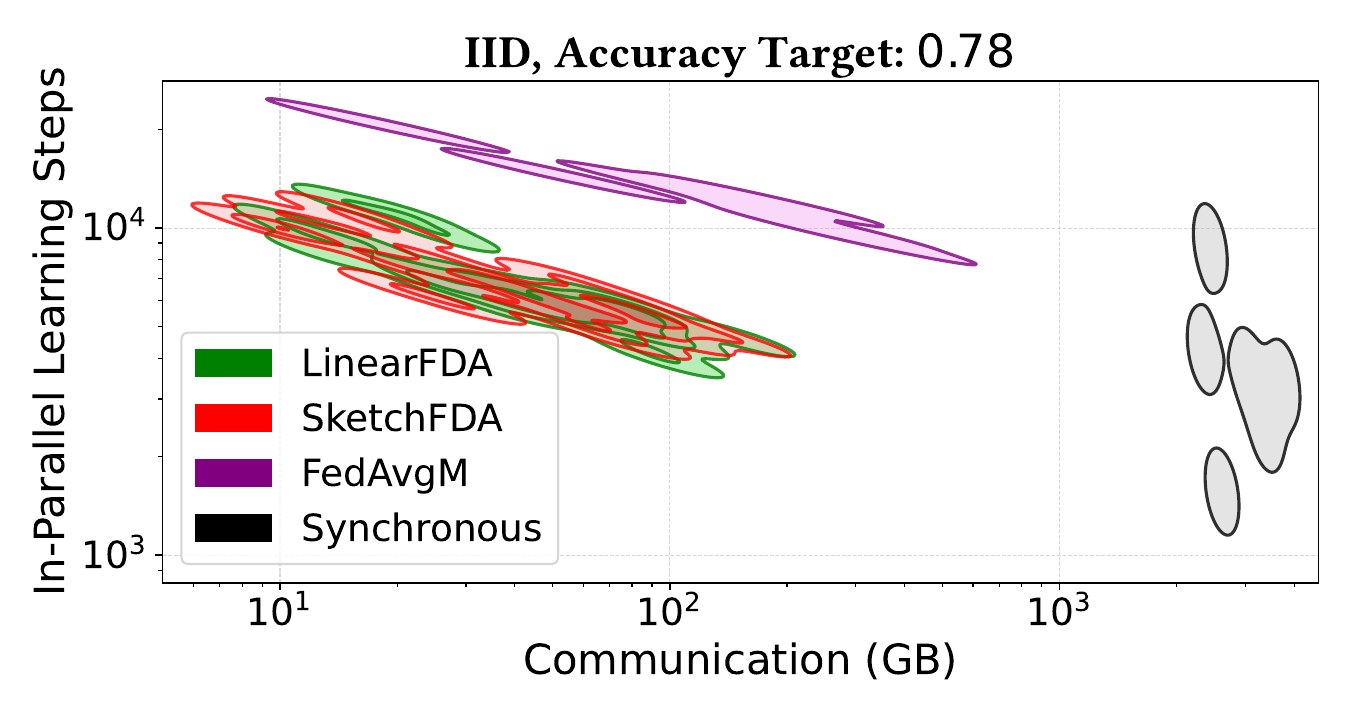}
        \includegraphics[width=0.851\textwidth]{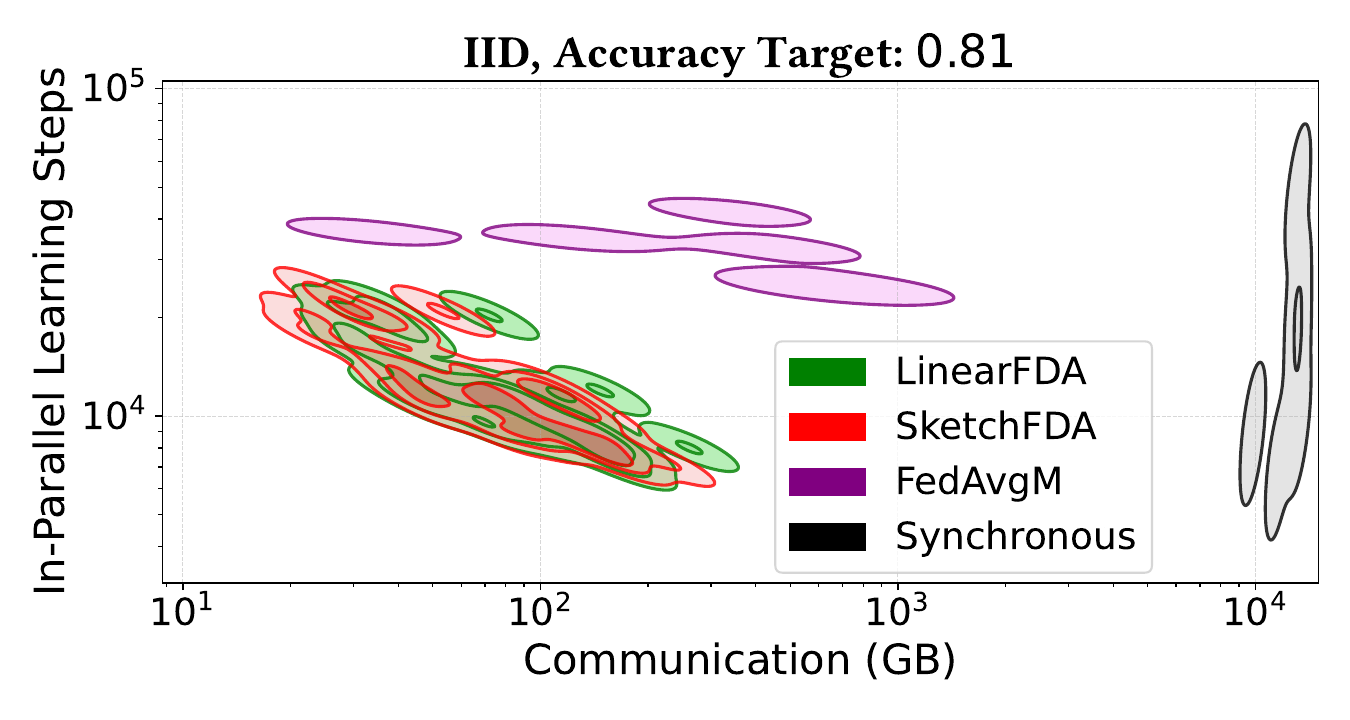}
        \caption{DenseNet121 on CIFAR-10}
        \label{fig:densenet121_joint_kde}
    \end{minipage}
    \hfill
    \begin{minipage}[b]{0.49\textwidth}
        \centering
        \includegraphics[width=0.851\textwidth]{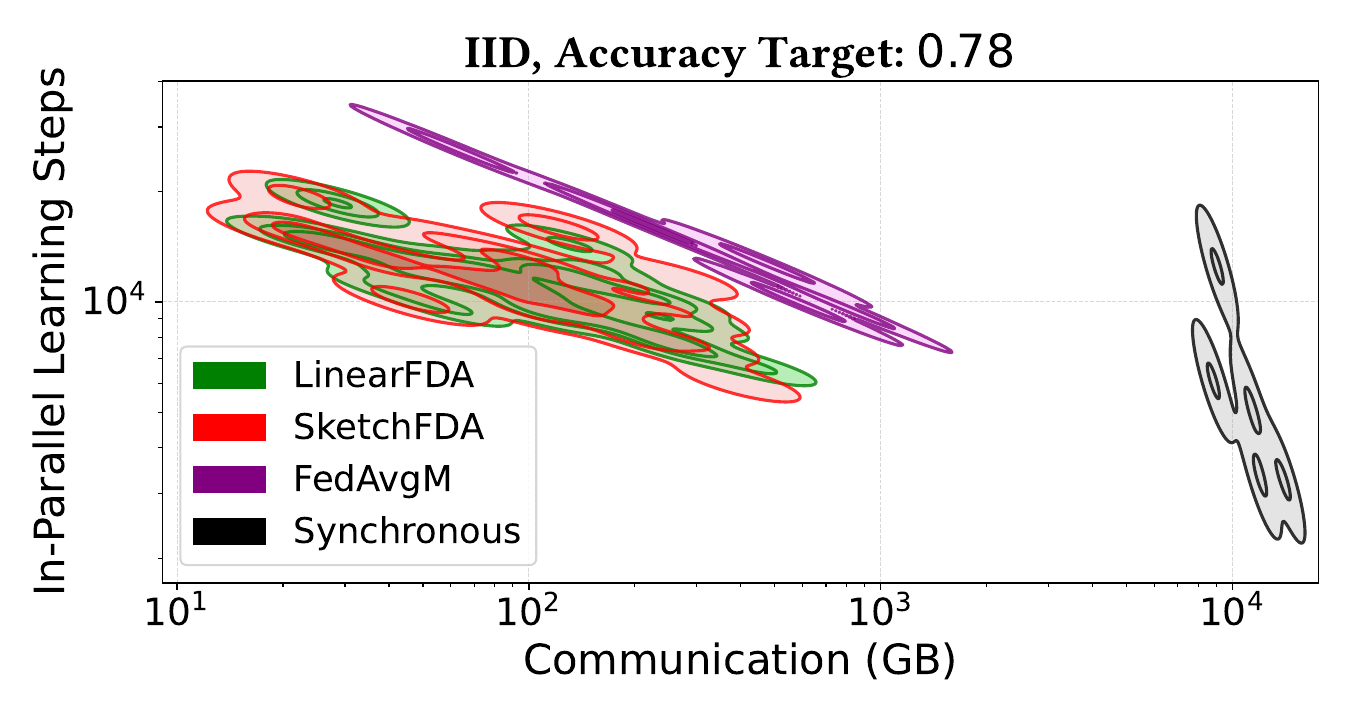}
        \includegraphics[width=0.851\textwidth]{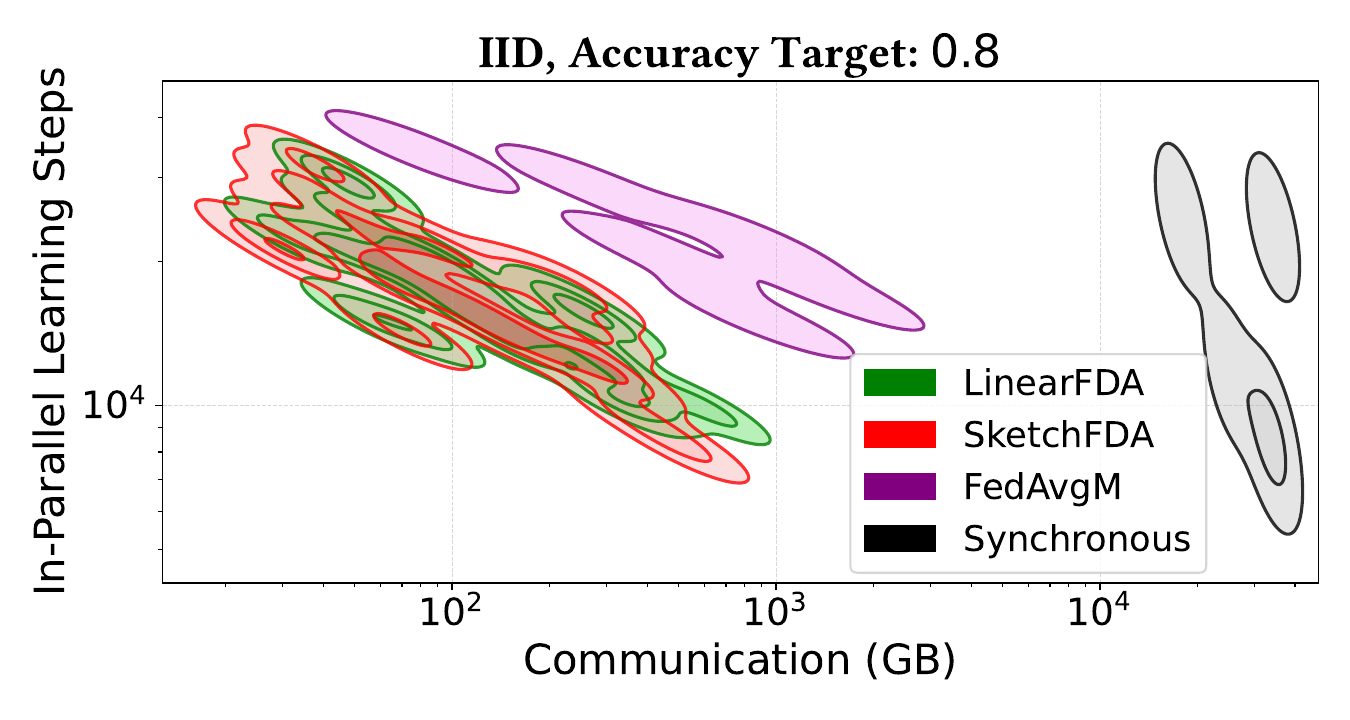}
        \caption{DenseNet201 on CIFAR-10}
        \label{fig:densenet201_joint_kde}
    \end{minipage}
\end{figure*}

\begin{figure}[t]
    \centering
    \begin{subfigure}[b]{0.49\textwidth}
        \centering
        \includegraphics[width=0.91\textwidth]{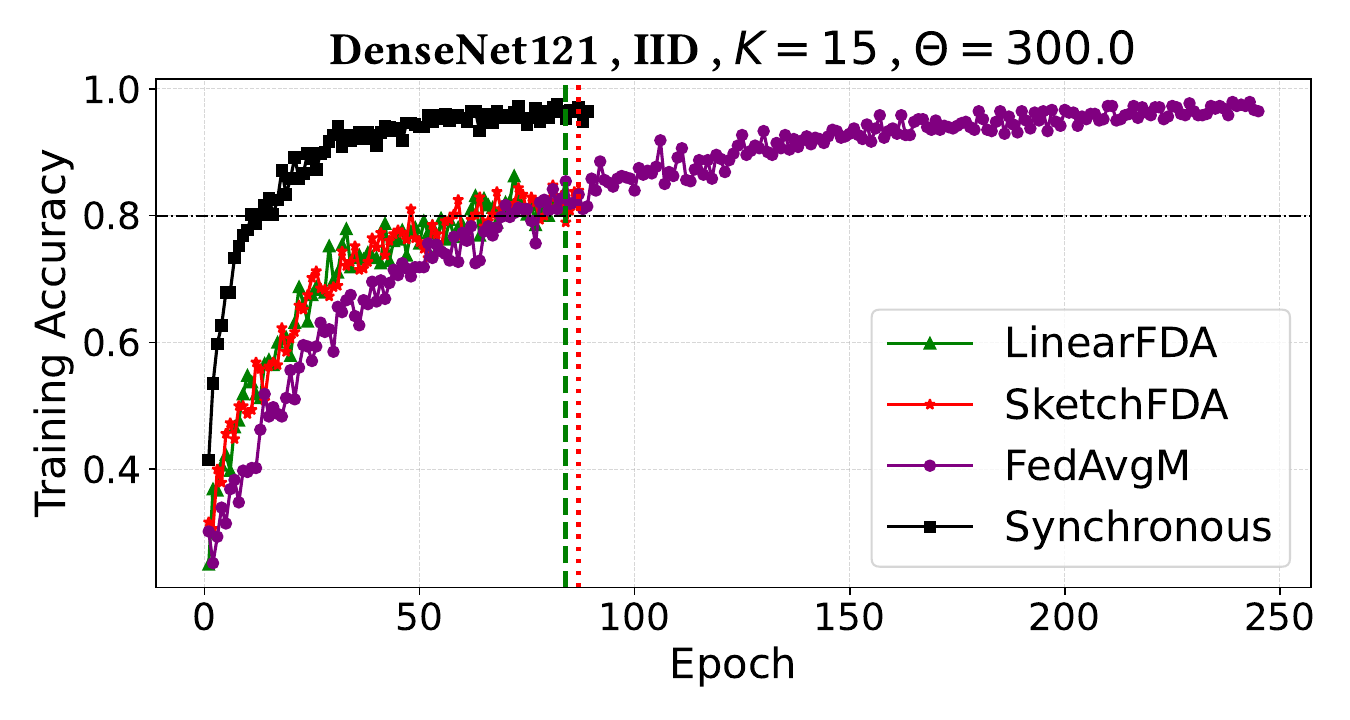}
    \end{subfigure}
    
    \begin{subfigure}[b]{0.49\textwidth}
        \centering
        \includegraphics[width=0.91\textwidth]{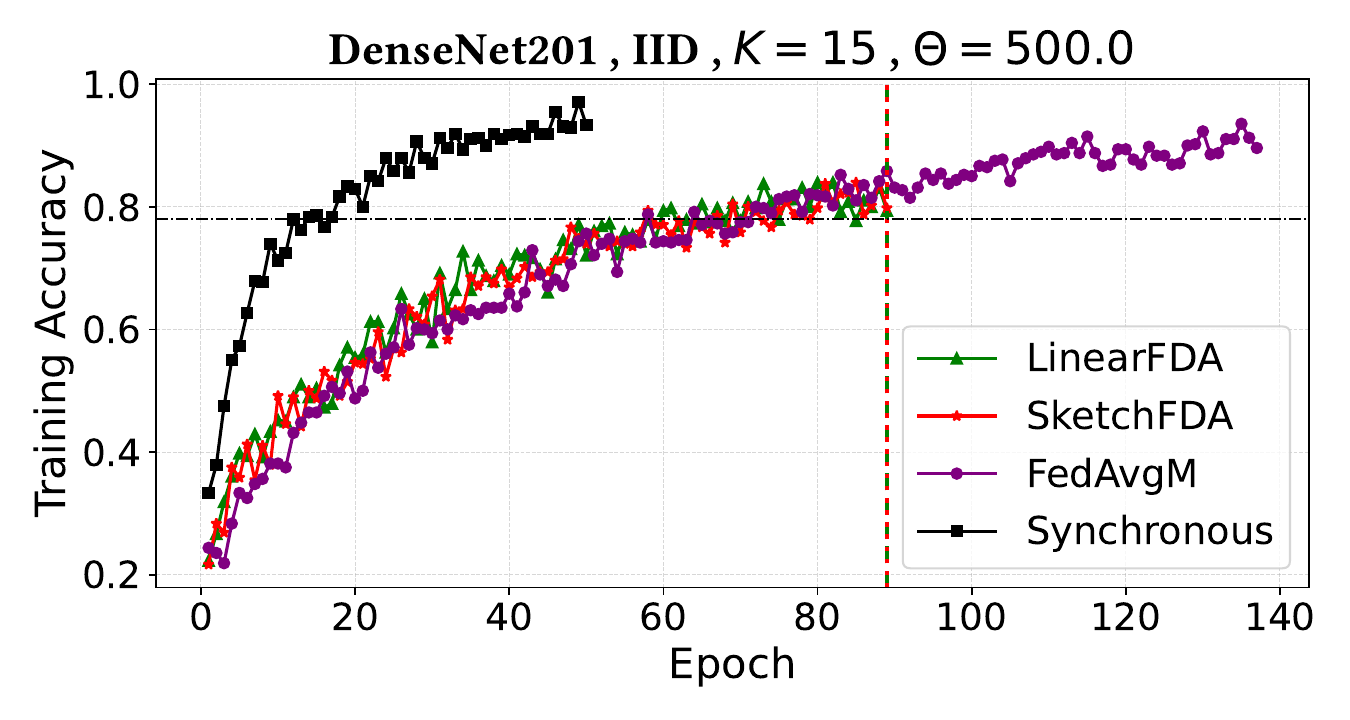}
    \end{subfigure}
    
    \caption{Training accuracy progression with a test accuracy target (horizontal line) of $0.8$ (top), and $0.78$ (bottom). Dashed and doted lines indicate when \textsc{LinearFDA} and \textsc{SketchFDA} attain the target accuracy, respectively. A smaller final gap between training and target accuracy indicates less overfitting, i.e., better generalization capabilities of the trained model}
    \label{fig:generalization_figure}
\end{figure}

\vspace{1.1mm}
\noindent \textbf{FDA balances Communication vs. Computation.} DDL algorithms face a fundamental challenge: balancing the competing demands of computation and communication. Frequent communication accelerates convergence and potentially improves model performance, but incurs higher network overhead, an overhead that may be prohibitive when workers communicate through lower speed connections. Conversely, reducing communication saves bandwidth but risks hindering, or even stalling, convergence. Traditional DDL approaches, like \textsc{Synchronous}, require synchronizing model parameters after every learning step, leading to significant communication overhead but facilitating faster convergence (lower computation cost). This is evident in Figures~\ref{fig:lenet5_joint_kde},~\ref{fig:vgg16_joint_kde},~\ref{fig:densenet121_joint_kde}, and~\ref{fig:densenet201_joint_kde} (where \textsc{Synchronous} appears in the bottom right --- low computation, very high communication). Conversely, Federated Optimization (\textsc{FedOpt}) methods~\cite{reddi2021fedadam} are designed to be communication-efficient, reducing communication between devices (workers) at the expense of increased local computation. Indeed, as shown in Figures~\ref{fig:lenet5_joint_kde}-\ref{fig:densenet201_joint_kde}, \textsc{FedAvgM} and \textsc{FedAdam} reduce communication by orders of magnitude but at the price of a corresponding increase in computation. Our two proposed \textsc{FDA} strategies achieve the best of both worlds: the low computation cost of traditional methods and the communication efficiency of \textsc{FedOpt} approaches, as seen in Figures~\ref{fig:lenet5_joint_kde},~\ref{fig:vgg16_joint_kde}, ~\ref{fig:densenet121_joint_kde}, and~\ref{fig:densenet201_joint_kde}. In fact, they significantly outperform \textsc{FedAvgM} and \textsc{FedAdam} in their element, that is, communication-efficiency. Across all experiments, the \textsc{FDA} methods' distributions lie in the desired bottom left quadrant --- low computation, very low communication.

\vspace{1mm}
\noindent \textbf{FDA counters diminishing returns.} The phenomenon of \textit{diminishing returns} states that as a DL model nears its learning limits for a given dataset and architecture, each additional increment in accuracy may necessitate a disproportionate increase in training time, tuning, and resources~\cite{DeepLearningGoodfellow, thompson2021dr}. We first clearly notice this with VGG16* on MNIST in Figure~\ref{fig:vgg16_joint_kde} for all three data heterogeneity settings. For a 0.001 increase in accuracy (effectively 10 misclassified testing images), \textsc{FedAdam} needs approximately 2-7$\times$ more communication and 3-7$\times$ more computation, respectively. This can be seen by comparing the figures at the left column of Figure~\ref{fig:vgg16_joint_kde} with the corresponding ones in the right column. Similarly, \textsc{Synchronous} requires comparable increases in computation and approximately half an order of magnitude more in communication. On the other hand, the \textsc{FDA} methods suffer a slight (if any) increase in computation and communication for this accuracy enhancement. For DenseNet121 and DenseNet201 on CIFAR-10 (Figures ~\ref{fig:densenet121_joint_kde}, and~\ref{fig:densenet201_joint_kde}), \textsc{FedAvgM} and \textsc{Synchronous} require half an order of magnitude more computation and communication to achieve the final marginal accuracy gains ($0.78$ to $0.81$ for DenseNet121, and $0.78$ to $0.8$ for DenseNet201). In contrast, the \textsc{FDA} methods have almost no increase in communication and comparable increase in computation.

\vspace{1mm}
\noindent \textbf{FDA is resilient to data heterogeneity.} In DDL, data heterogeneity is a prevalent challenge, reflecting the complexity of real-world applications where the IID assumption often does not hold. The ability of DDL algorithms to maintain consistent performance in the face of non-IID data is a critical metric for their effectiveness and adaptability. Our empirical investigation reveals the \textsc{FDA} methods' noteworthy resilience in such heterogeneous environments. For LeNet-5 on MNIST, as illustrated in Figure~\ref{fig:lenet5_joint_kde}, the computation and communication costs required to attain a test accuracy of $0.985$ show negligible differences across the IID and the two Non-IID settings (Label "$0$", $60\%$). Similarly, for VGG16* on MNIST, Figure~\ref{fig:vgg16_joint_kde} demonstrates that achieving a test accuracy of $0.995$ incurs comparable computation and communication costs across the IID and the two Non-IID settings (Label "$0$", Label "$8$"); while overall costs are aligned, the distributions of the computation costs exhibit greater variability, yet remain closely consistent with the IID scenario.

\vspace{1mm}
\noindent \textbf{FDA has a lower generalization gap.} The factors determining how well a DL algorithm performs are its ability to: (1) make the training accuracy high, and (2) make the gap between training and test accuracy small. These two factors correspond to the two central challenges in DL: underfitting and overfitting~\cite{DeepLearningGoodfellow}. For DenseNet121 on CIFAR-10, with a test accuracy target of $0.8$, as illustrated in Figure~\ref{fig:generalization_figure}, \textsc{Synchronous} and \textsc{FedAvgM} exhibit overfitting, with a noticeable discrepancy between training and test accuracy. In stark contrast, the \textsc{FDA} methods have an almost zero accuracy gap. Please note that \textsc{LinearFDA} and \textsc{SketchFDA} reach the test accuracy target of $0.8$ much earlier (at epochs 86 and 91, respectively). 
Turning our focus to DenseNet201 on CIFAR-10, with a test accuracy target of $0.78$, \textsc{Synchronous} again tends towards overfitting, while \textsc{FedAvgM} shows a slight improvement but still does not match the \textsc{FDA} methods, which continue to exhibit exceptional generalization capabilities, evidenced by a minimal training-test accuracy gap, as shown at Figure~\ref{fig:generalization_figure}. Notably, given the necessity to fix hyper-parameters $\Theta$ and $K$ for the training accuracy plots, we selected two representative examples. The patterns of performance we highlighted are consistent across most of the conducted tests.

\begin{figure}[t]
    \centering
    \begin{subfigure}[t]{0.49\textwidth}
        \centering
        \includegraphics[width=0.911\textwidth]{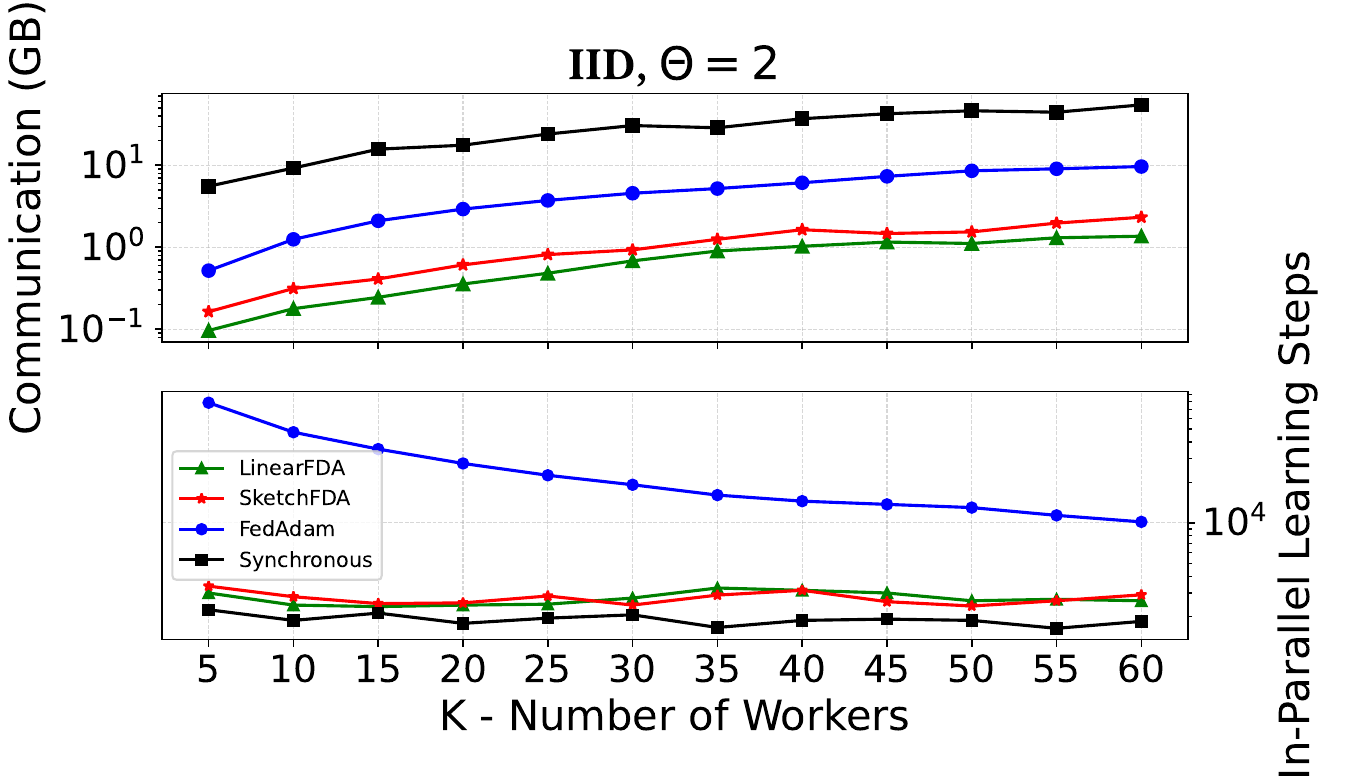}
    \end{subfigure}
    \begin{subfigure}[t]{0.49\textwidth}
        \centering
        \includegraphics[width=0.911\textwidth]{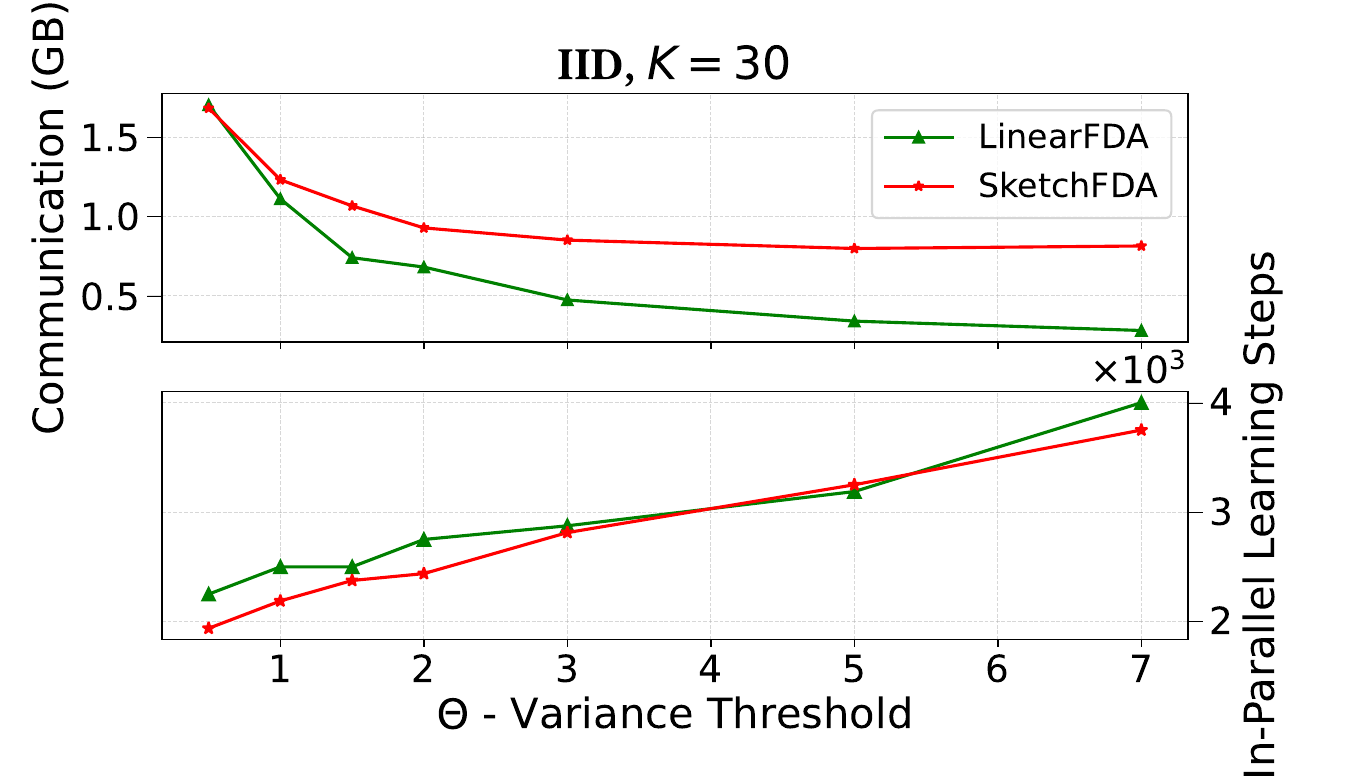}
    \end{subfigure}

    \caption{LeNet-5 on MNIST: Varying the Number of Workers and $\Theta$ --- Accuracy Target: $0.98$}
    \label{fig:lenet5_theta_K}
\end{figure}

\begin{figure}[t]
    \centering
    \begin{subfigure}[t]{0.49\textwidth}
        \centering
        \includegraphics[width=0.911\textwidth]{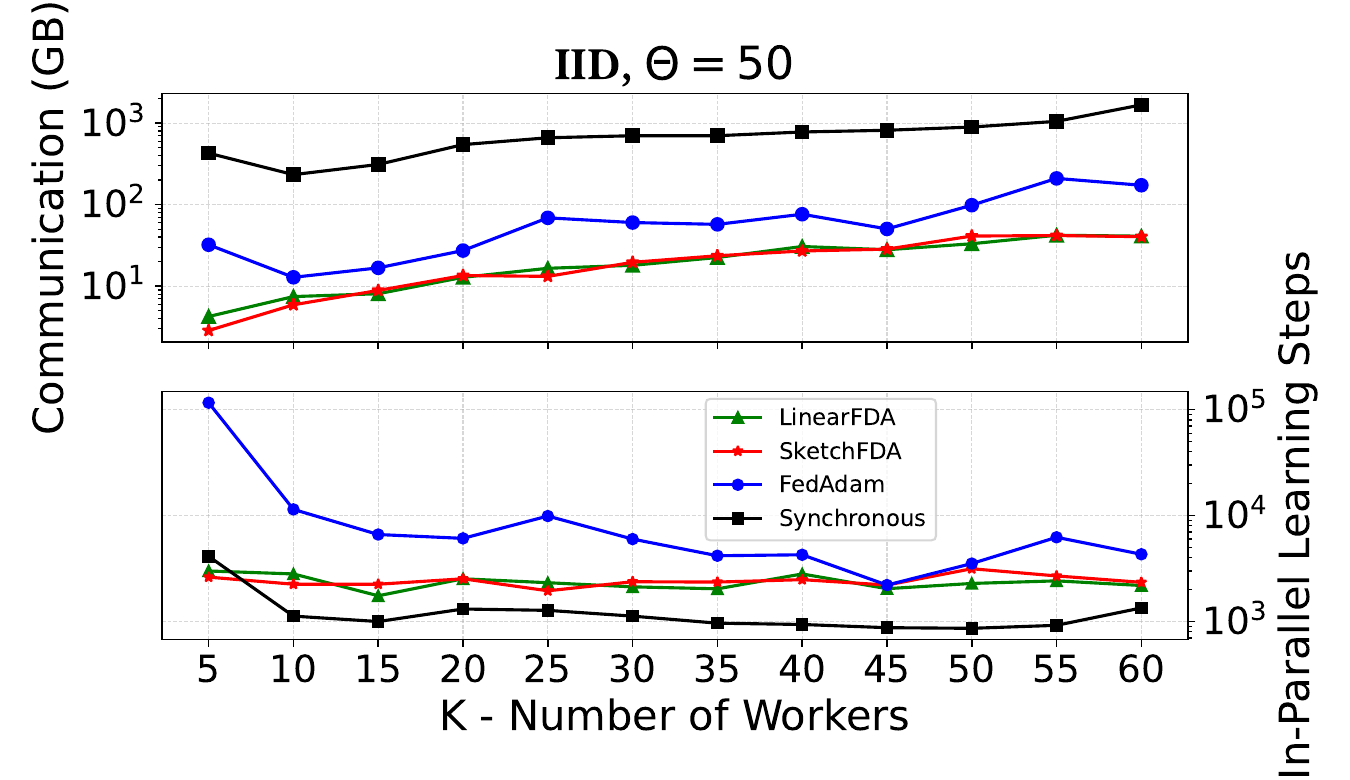}
    \end{subfigure}
    \begin{subfigure}[t]{0.49\textwidth}
        \centering
        \includegraphics[width=0.911\textwidth]{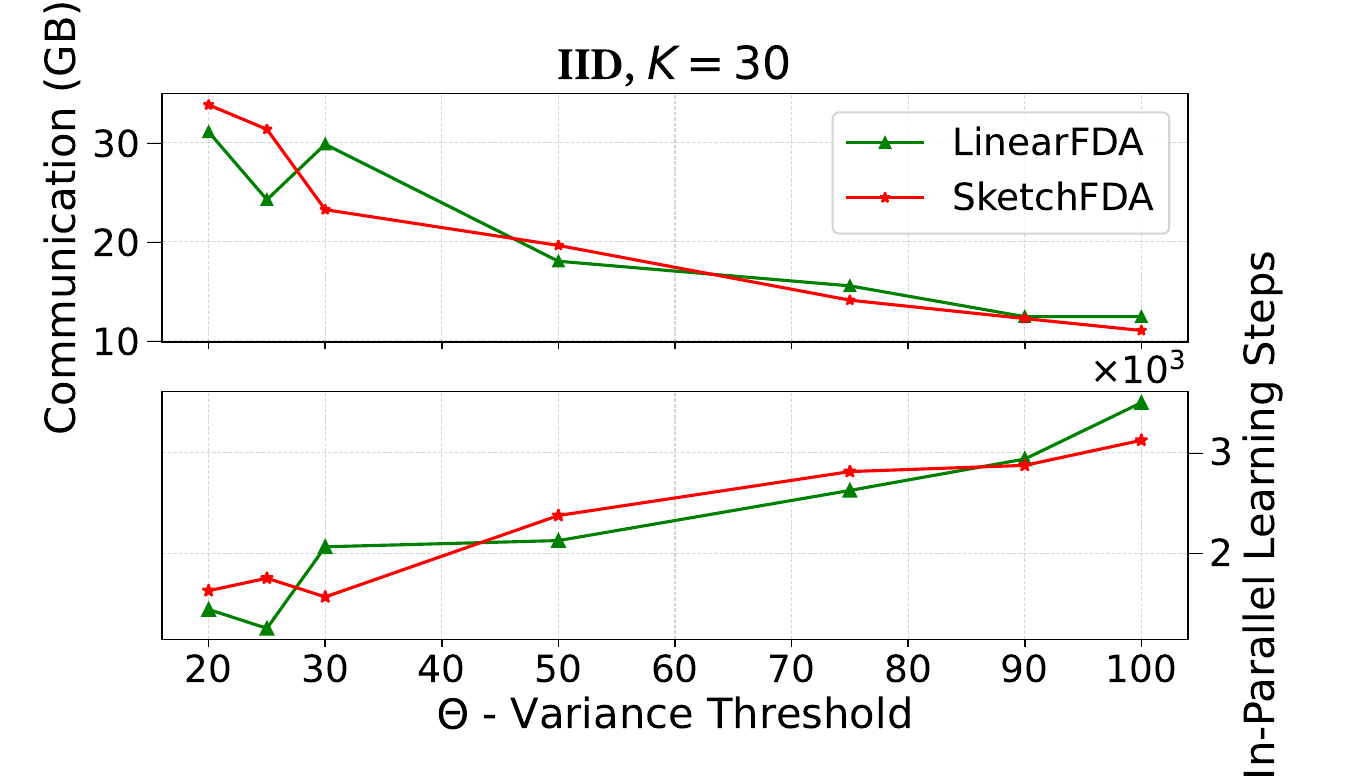}
    \end{subfigure}

    \caption{VGG16* on MNIST: Varying the Number of Workers and $\Theta$ --- Accuracy Target: $0.994$}
    \label{fig:vgg16_theta_K}
\end{figure}

\begin{figure}[t]
    \centering
    \begin{subfigure}[t]{0.49\textwidth}
        \centering
        \includegraphics[width=0.911\textwidth]{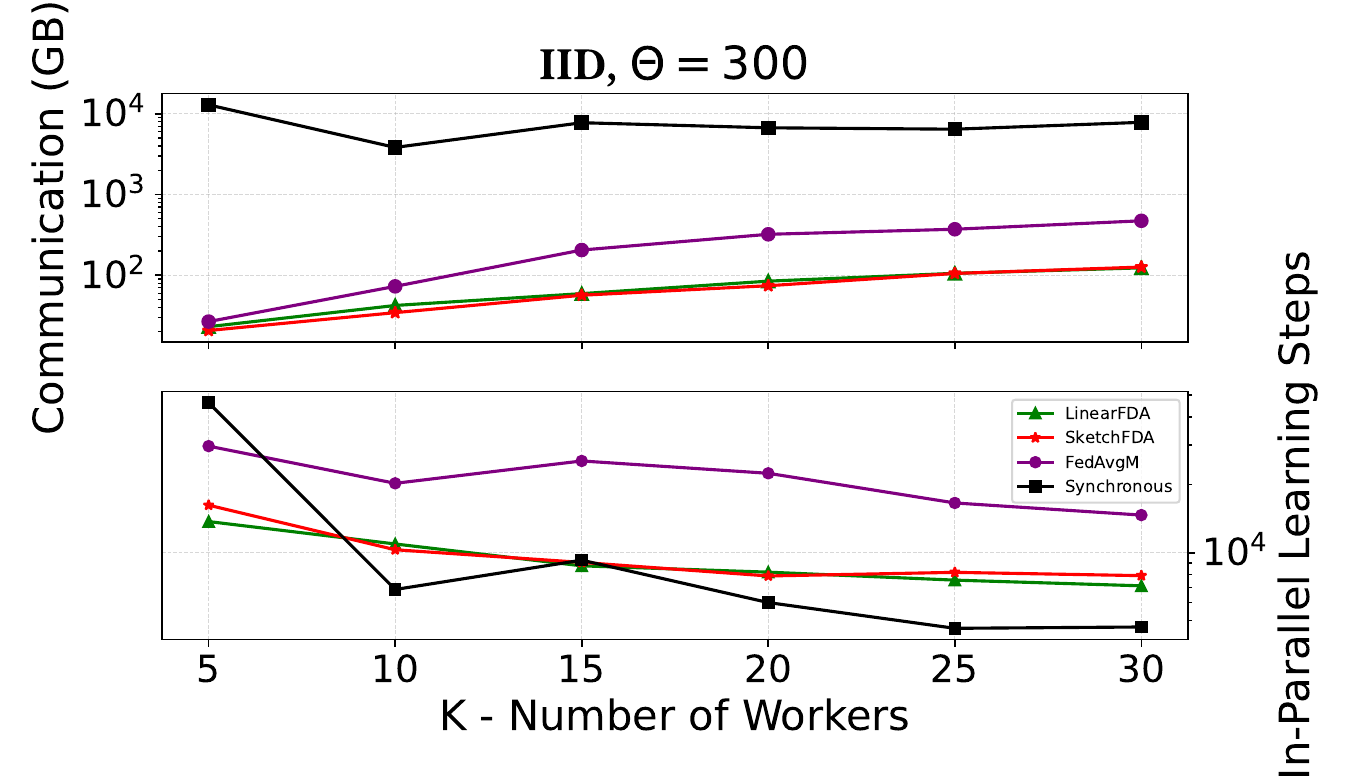}
    \end{subfigure}
    \begin{subfigure}[t]{0.49\textwidth}
        \centering
        \includegraphics[width=0.911\textwidth]{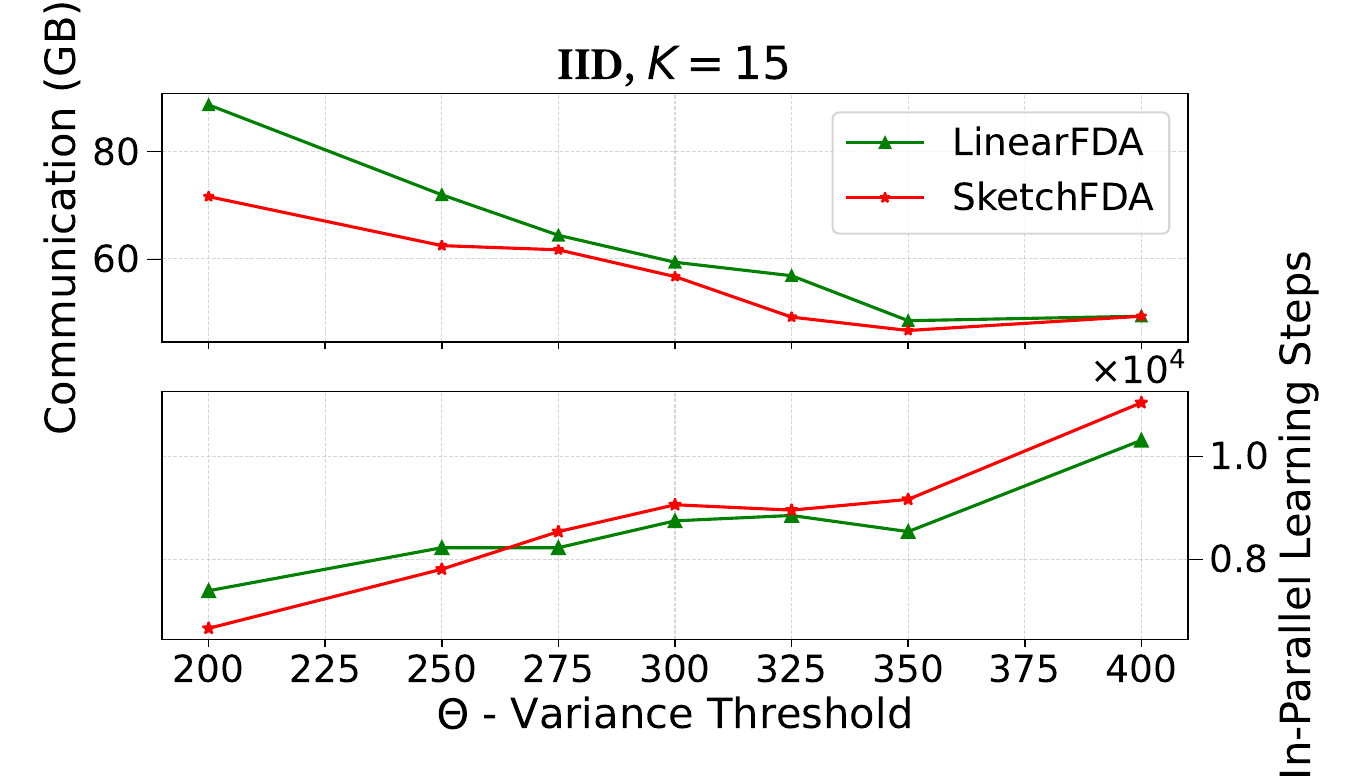}
    \end{subfigure}

    \caption{DenseNet121 on CIFAR-10: Varying the Number of Workers and $\Theta$ --- Accuracy Target: $0.8$}
    \label{fig:densenet121_theta_K}
\end{figure}

\begin{figure}[t]
    \centering
    \begin{subfigure}[t]{0.49\textwidth}
        \centering
        \includegraphics[width=0.911\textwidth]{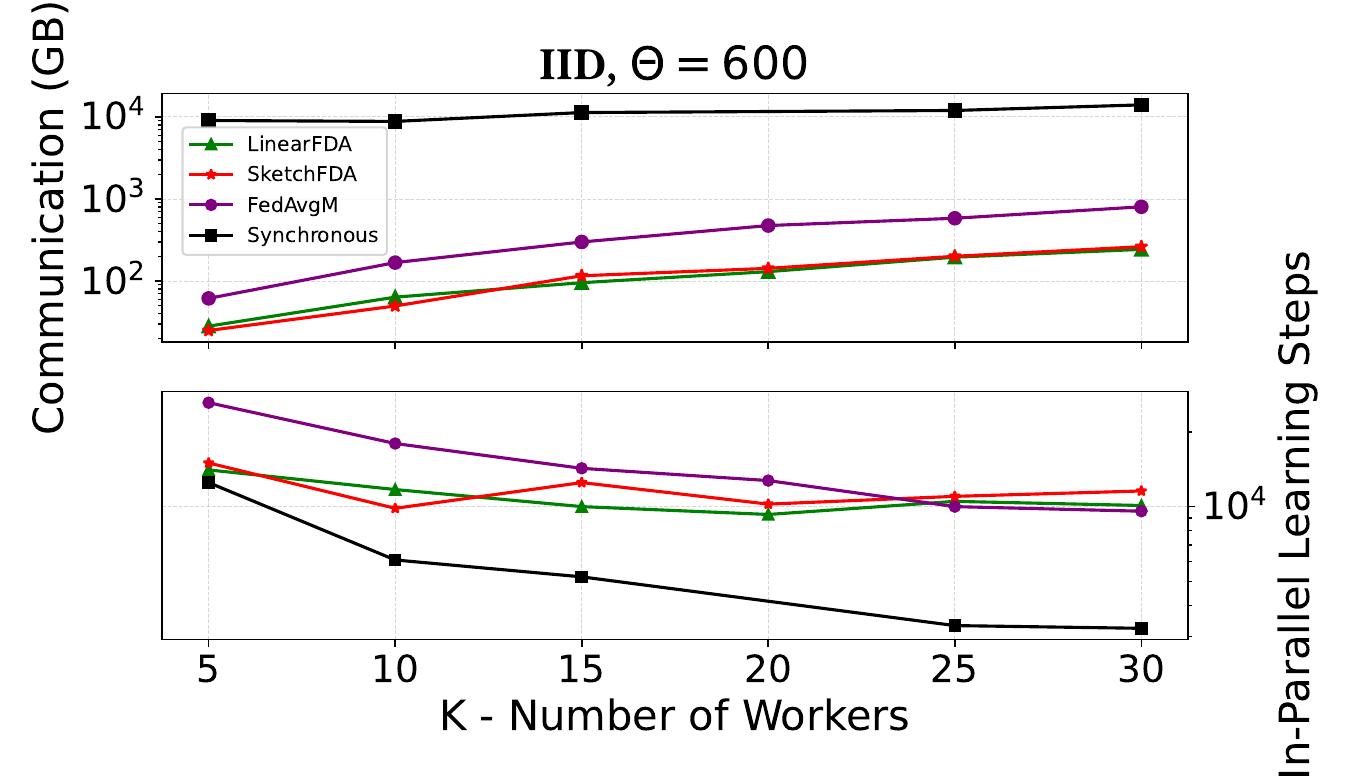}
    \end{subfigure}
    \begin{subfigure}[t]{0.49\textwidth}
        \centering
        \includegraphics[width=0.911\textwidth]{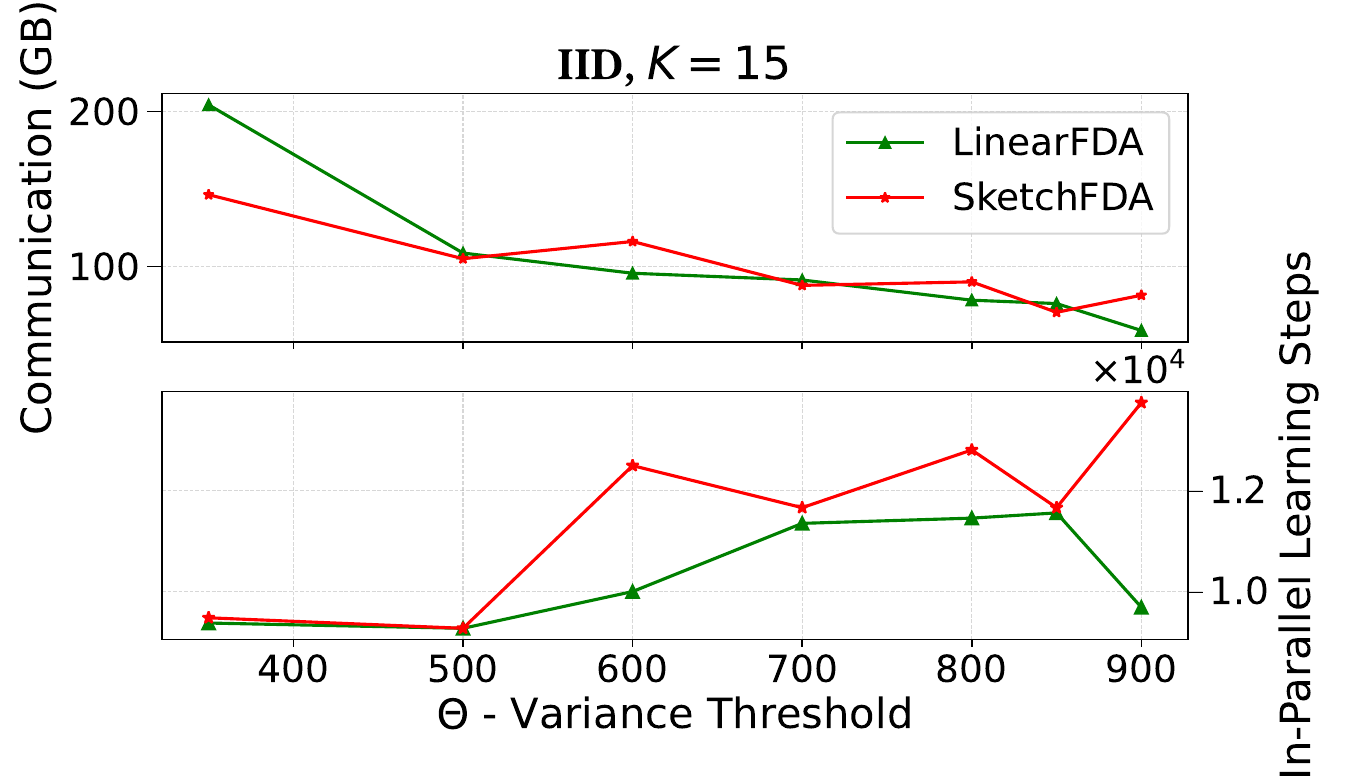}
    \end{subfigure}
    \caption{DenseNet201 on CIFAR-10: Varying the Number of Workers and $\Theta$ --- Accuracy Target: $0.78$}
    \label{fig:densenet201_theta_K}
\end{figure}

\vspace{1.1mm}
\noindent \textbf{Dependence on $K$.} In distributed computing, scaling up typically results in proportional speed improvements. In DDL, however, scalability is less predictable due to the nuanced interplay of computation and communication costs with convergence, complicating the expected linear speedup \cite{yu2018parallel}. This unpredictability is starkly illustrated with LeNet-5 and VGG16* on the MNIST dataset across all data heterogeneity settings and all strategies. Figures~\ref{fig:lenet5_theta_K}, and \ref{fig:vgg16_theta_K} (top) demonstrate that increasing the number of workers does not decrease computation -- except for \textsc{FedAdam} which begins with significantly high computation -- but rather exacerbates communication. These findings are troubling, as they reveal scaling up only hampers training speed and wastes resources. However, for more complex learning tasks like training DenseNet-121 and DenseNet-201 on CIFAR-10 (top part of Figures~\ref{fig:densenet121_theta_K},~\ref{fig:densenet201_theta_K}), the expected behavior starts to emerge. Especially for DenseNet-121, scaling up ($K$ increase) leads to a decrease in computation cost for all strategies. Communication cost, however, increases with $K$ for all methods except \textsc{Synchronous}, which maintains constant communication irrespective of worker count, but at the expense of orders of magnitude higher communication overhead. Notably, while our findings might, in some cases, suggest potential speed benefits of not scaling up (smaller $K$), DDL is increasingly conducted within federated settings, where there is no other choice but to utilize the high number of workers. Our FDA variants consistently outperform \textsc{FedAdam}, \textsc{FedAvgM}, and \textsc{Synchronous} in communication efficiency, as demonstrated across all experiments in Figures~\ref{fig:lenet5_theta_K}-\ref{fig:densenet201_theta_K}. Specifically, they require up to 30 times less communication than \textsc{FedAdam}, 4 times less than \textsc{FedAvgM}, and up to 2.5 orders of magnitude less than \textsc{Synchronous}.

\vspace{2mm}
\noindent \textbf{FDA: Dependence on $\Theta$.} The variance threshold $\Theta$ can be seen as a lever in balancing communication and computation; essentially, it calibrates the trade-off between these two costs. A higher $\Theta$ allows for greater model divergence before synchronization, reducing communication at the cost of potentially increased computation to achieve convergence. This impact of $\Theta$ is consistently observed across both FDA strategies, and all learning tasks, and data heterogeneity settings (Figures~\ref{fig:lenet5_theta_K}-\ref{fig:densenet201_theta_K}). Interestingly, for more complex models like DenseNet121 and DenseNet201 on CIFAR-10, increasing the variance threshold ($\Theta$) does not lead to a significant rise in computation cost, as illustrated in Figures~\ref{fig:densenet121_theta_K} and~\ref{fig:densenet201_theta_K}. It suggests that the \textsc{FDA} methods, by strategically timing synchronizations (monitoring the variance), substantially reduce the number of necessary synchronizations without a proportional increase in computation for the same model performance; this is particularly promising for complex DDL tasks.

\vspace{1mm}
\noindent \textbf{FDA: Choice of $\Theta$.} The experimental results suggest that selecting any $\Theta$ within a specific order of magnitude (e.g., between $10^2$ and $10^3$ for DenseNet201) ensures convergence, as demonstrated in Figures~\ref{fig:lenet5_theta_K}, ~\ref{fig:vgg16_theta_K},~\ref{fig:densenet121_theta_K}, and~\ref{fig:densenet201_theta_K}. Therefore, identifying this range becomes crucial. To this end, we conducted extensive exploratory testing to estimate the $\Theta$ ranges for each learning task which are predominantly influenced by the number of parameters $d$ of the DNN. Within this context, $\Theta$ values outside the desirable range exhibit notable effects: below this range, the training process mimics \textsc{Synchronous} or Local-SGD approaches with small $\tau$, while exceeding it leads to non-convergence. Subsequently, having identified the optimal ranges for $\Theta$, we selected diverse values within them for our experimental evaluation (Table~\ref{tab:summary_experiments}), thereby investigating different computation and communication trade-offs. For instance, in the ARIS-HPC environment with an InfiniBand connection (up to 56 Gb/s), experiments show that training wall-time (the total time required for the computation and the communication of the DDL) is predominantly influenced by the computation cost, rendering communication concerns negligible. In such contexts, lower $\Theta$ values are favored due to their computational efficiency. On the contrary, in FL settings, where communication typically poses the greater challenge, opting for higher $\Theta$ values proves advantageous; reduction in communication achieved with higher $\Theta$ values will translate in a large reduction in total wall-time.

\begin{figure}[t]
  \centering
  \includegraphics[width=0.45\textwidth]{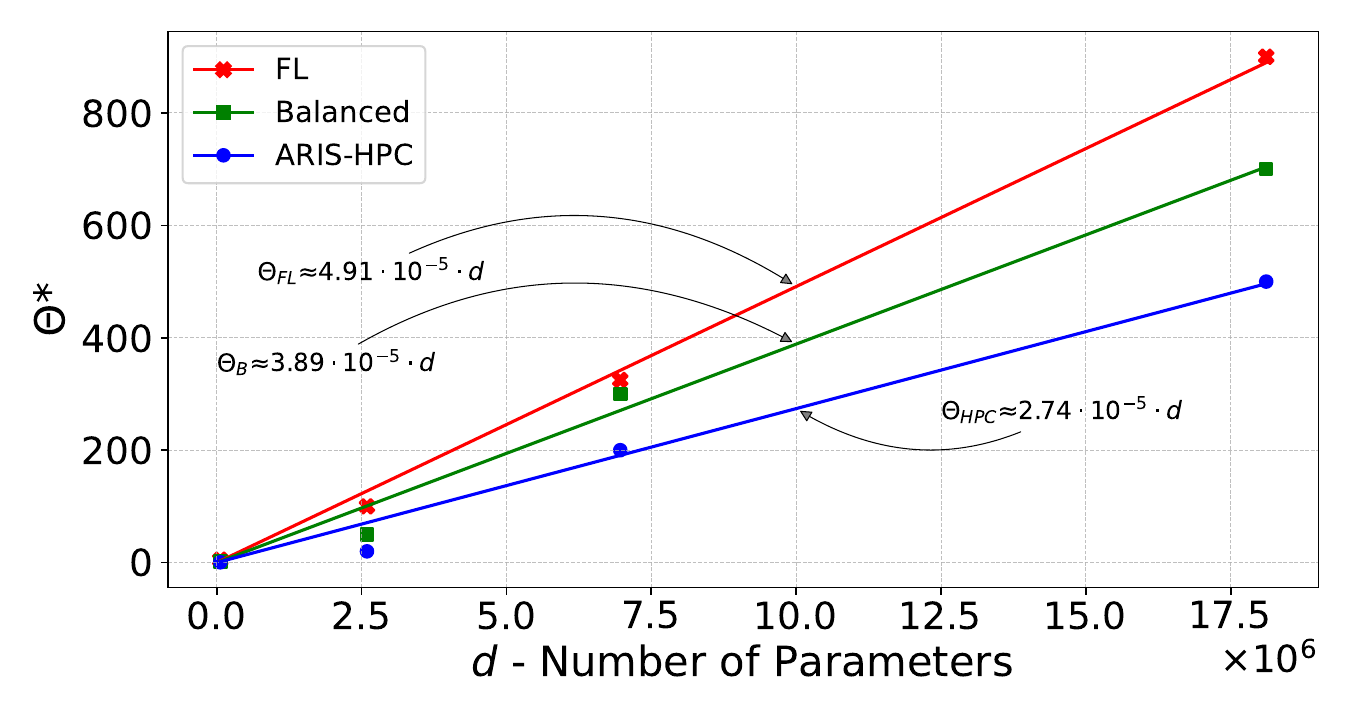}
  \caption{Empirical Estimation of the Variance Threshold}
  \label{fig:theta_estimation}
\end{figure}

\balance
To assist researchers in selecting the variance threshold, Figure~\ref{fig:theta_estimation} presents empirical estimations for $\Theta$ across three distinct learning settings: \eat{(1) FL (assuming a common channel of $0.5$Gbps), (2) Balanced communication-computation equilibrium setting, and (3) our HPC environment at the ARIS supercomputer.
\begin{align*}
    \Theta_{FL} &= 4.91 \cdot 10^{-5} \cdot d \\
    \Theta_{B} &= 3.89 \cdot 10^{-5} \cdot d \\
    \Theta_{HPC} &= 2.74 \cdot 10^{-5} \cdot d
\end{align*}
}

\begin{enumerate}[leftmargin=20pt]

    \item FL, assuming a common channel of $0.5$Gbps, where
    \begin{align*}
        \Theta_{FL} &= 4.91 \cdot 10^{-5} \cdot d
    \end{align*}

    \item Balanced communication-computation equilibrium, where
    \begin{align*}
        \Theta_{B} &= 3.89 \cdot 10^{-5} \cdot d
    \end{align*}
    
    \item Our HPC environment at the ARIS supercomputer, where
    \begin{align*}
        \Theta_{HPC} &= 2.74 \cdot 10^{-5} \cdot d
    \end{align*}
\end{enumerate}

\begin{figure}[t]           
  \centering
  \includegraphics[width=0.49\textwidth]{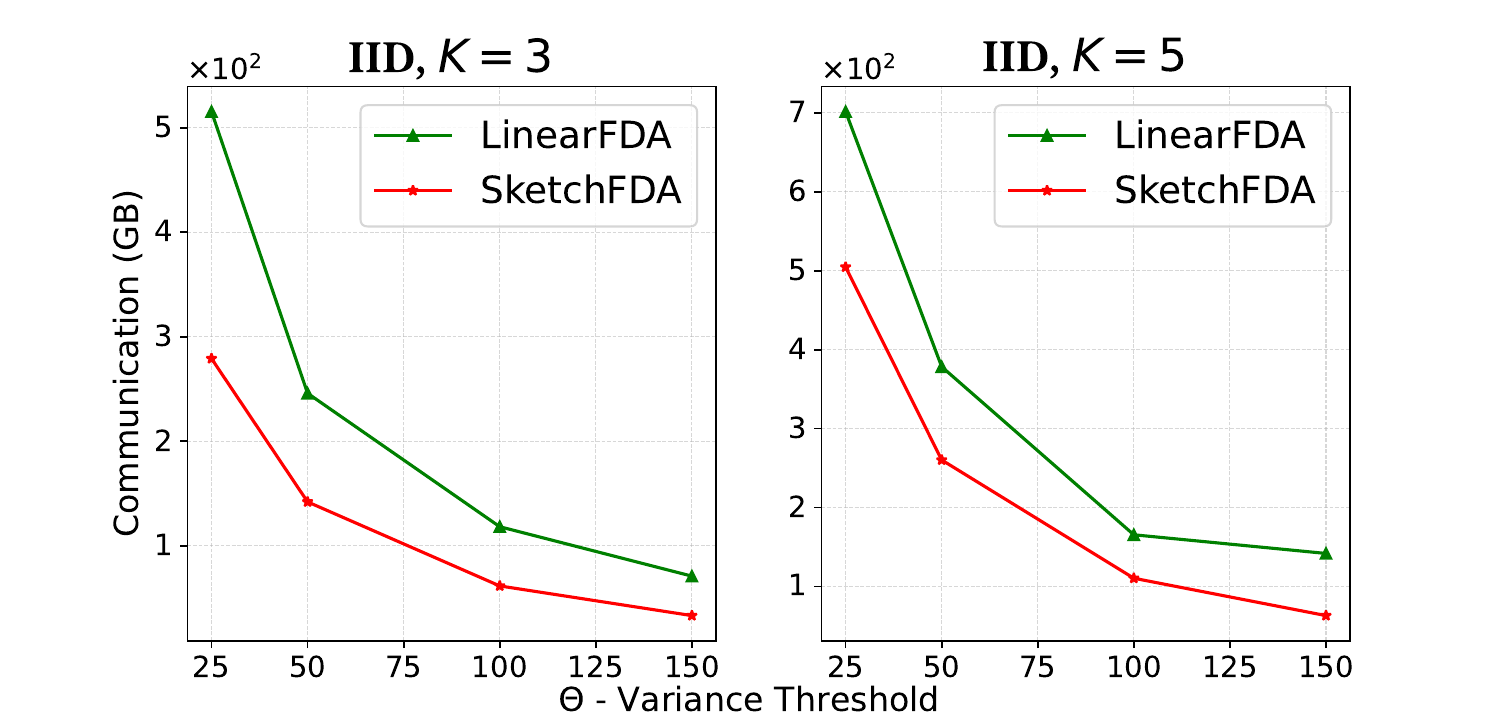}
  \caption{ConvNeXtLarge on CIFAR-100 (transfer learning from ImageNet) --- Deployment of \textsc{FDA} during the fine-tuning stage with Accuracy Target of $0.76$}
  \label{fig:convnextlarge}
\end{figure}

\vspace{1.3mm}
\noindent \textbf{FDA: Linear vs. Sketch.} In our main body of experiments, across most learning tasks and data heterogeneity settings, the two proposed FDA methods exhibit comparable performance, as illustrated in Figures~\ref{fig:lenet5_joint_kde},~\ref{fig:vgg16_joint_kde},~\ref{fig:densenet121_joint_kde}, and~\ref{fig:densenet201_joint_kde}. This suggests that the precision of the variance approximation is not critical. However, in all experiments within the more intricate transfer learning scenario, \textsc{LinearFDA} requires approximately $1.5$ times more communication than \textsc{SketchFDA} to fine-tune the deep ConvNeXtLarge model to equivalent performance levels (Figure~\ref{fig:convnextlarge}). 
In light of these findings, we conclude the following: for straightforward and less demanding tasks, \textsc{LinearFDA} is the recommended option due to its simplicity and lower complexity per local state computation. On the other hand, for intricate learning tasks and deeper models, \textsc{SketchFDA} becomes the preferred choice, if communication-efficiency is paramount.

\section{Conclusions and Future Work}\label{section:conclusion}
In this paper, we introduced Federated Dynamic Averaging (\textsc{FDA}), an innovative, adaptive and communication-efficient algorithm for distributed deep learning. Essentially, \textsc{FDA} makes informed, dynamic decisions on when to synchronize the local models based on approximations of the model variance. Through extensive experiments across diverse datasets and learning tasks, we demonstrated that \textsc{FDA}  significantly reduces communication overhead (often by orders of magnitude) without a corresponding increase in computation or compromise in model performance---contrary to the typical trade-offs encountered in the literature. Furthermore, we showed that \textsc{FDA} is robust to data heterogeneity and inherently mitigates over-fitting. Our results push the limits of modern communication-efficient distributed deep learning, paving the way for more scalable, dynamic, and broadly applicable strategies.

An interesting direction for future work is whether the value of $\Theta$ can be dynamically adjusted in order to achieve (or not to exceed) a target average bandwidth consumption. Since the expected behavior is that the communication cost decreases when $\Theta$ increases, such an approach seems feasible (i.e., increasing $\Theta$ when the bandwidth consumption is higher than what is desired), especially by using statistics. We plan to look into this extension in the future.

\begin{acks}
We wish to thank the anonymous reviewers and the meta-reviewer for their insightful comments and suggestions. This work was supported by the EU project CREXDATA under Horizon Europe agreement No. 101092749. Moreover, this work was supported by computational time granted from the National Infrastructures for Research and Technology S.A. (GRNET S.A.) in the National HPC facility - ARIS - under project ID pa230902-fda1.
\end{acks}

%% file: main.bbl

\begin{thebibliography}{69}


\ifx \showCODEN    \undefined \def \showCODEN     #1{\unskip}     \fi
\ifx \showDOI      \undefined \def \showDOI       #1{#1}\fi
\ifx \showISBNx    \undefined \def \showISBNx     #1{\unskip}     \fi
\ifx \showISBNxiii \undefined \def \showISBNxiii  #1{\unskip}     \fi
\ifx \showISSN     \undefined \def \showISSN      #1{\unskip}     \fi
\ifx \showLCCN     \undefined \def \showLCCN      #1{\unskip}     \fi
\ifx \shownote     \undefined \def \shownote      #1{#1}          \fi
\ifx \showarticletitle \undefined \def \showarticletitle #1{#1}   \fi
\ifx \showURL      \undefined \def \showURL       {\relax}        \fi
\providecommand\bibfield[2]{#2}
\providecommand\bibinfo[2]{#2}
\providecommand\natexlab[1]{#1}
\providecommand\showeprint[2][]{arXiv:#2}

\bibitem[\protect\citeauthoryear{Abadi, Barham, Chen, Chen, Davis, Dean, Devin, Ghemawat, Irving, Isard, Kudlur, Levenberg, Monga, Moore, Murray, Steiner, Tucker, Vasudevan, Warden, Wicke, Yu, and Zheng}{Abadi et~al\mbox{.}}{2016}]%
        {abadi2015tensorflow}
\bibfield{author}{\bibinfo{person}{Mart\'{\i}n Abadi}, \bibinfo{person}{Paul Barham}, \bibinfo{person}{Jianmin Chen}, \bibinfo{person}{Zhifeng Chen}, \bibinfo{person}{Andy Davis}, \bibinfo{person}{Jeffrey Dean}, \bibinfo{person}{Matthieu Devin}, \bibinfo{person}{Sanjay Ghemawat}, \bibinfo{person}{Geoffrey Irving}, \bibinfo{person}{Michael Isard}, \bibinfo{person}{Manjunath Kudlur}, \bibinfo{person}{Josh Levenberg}, \bibinfo{person}{Rajat Monga}, \bibinfo{person}{Sherry Moore}, \bibinfo{person}{Derek~G. Murray}, \bibinfo{person}{Benoit Steiner}, \bibinfo{person}{Paul Tucker}, \bibinfo{person}{Vijay Vasudevan}, \bibinfo{person}{Pete Warden}, \bibinfo{person}{Martin Wicke}, \bibinfo{person}{Yuan Yu}, {and} \bibinfo{person}{Xiaoqiang Zheng}.} \bibinfo{year}{2016}\natexlab{}.
\newblock \showarticletitle{TensorFlow: a system for large-scale machine learning}. In \bibinfo{booktitle}{\emph{Proceedings of the 12th USENIX Conference on Operating Systems Design and Implementation}} \emph{(\bibinfo{series}{OSDI'16})}. \bibinfo{publisher}{USENIX Association}, \bibinfo{address}{USA}, \bibinfo{pages}{265–283}.
\newblock
\showISBNx{9781931971331}


\bibitem[\protect\citeauthoryear{Acar, Zhao, Matas, Mattina, Whatmough, and Saligrama}{Acar et~al\mbox{.}}{2021}]%
        {acar2021fedDyn}
\bibfield{author}{\bibinfo{person}{Durmus Alp~Emre Acar}, \bibinfo{person}{Yue Zhao}, \bibinfo{person}{Ramon Matas}, \bibinfo{person}{Matthew Mattina}, \bibinfo{person}{Paul Whatmough}, {and} \bibinfo{person}{Venkatesh Saligrama}.} \bibinfo{year}{2021}\natexlab{}.
\newblock \showarticletitle{Federated Learning Based on Dynamic Regularization}. In \bibinfo{booktitle}{\emph{International Conference on Learning Representations}}.
\newblock


\bibitem[\protect\citeauthoryear{Aji and Heafield}{Aji and Heafield}{2017}]%
        {aji2017sparsification}
\bibfield{author}{\bibinfo{person}{Alham~Fikri Aji} {and} \bibinfo{person}{Kenneth Heafield}.} \bibinfo{year}{2017}\natexlab{}.
\newblock \showarticletitle{Sparse Communication for Distributed Gradient Descent}. In \bibinfo{booktitle}{\emph{Proceedings of the 2017 Conference on Empirical Methods in Natural Language Processing}}, \bibfield{editor}{\bibinfo{person}{Martha Palmer}, \bibinfo{person}{Rebecca Hwa}, {and} \bibinfo{person}{Sebastian Riedel}} (Eds.). \bibinfo{publisher}{Association for Computational Linguistics}, \bibinfo{address}{Copenhagen, Denmark}, \bibinfo{pages}{440--445}.
\newblock


\bibitem[\protect\citeauthoryear{Basu, Data, Karakus, and Diggavi}{Basu et~al\mbox{.}}{2019}]%
        {basu2019qsparselocalsgd}
\bibfield{author}{\bibinfo{person}{Debraj Basu}, \bibinfo{person}{Deepesh Data}, \bibinfo{person}{Can Karakus}, {and} \bibinfo{person}{Suhas Diggavi}.} \bibinfo{year}{2019}\natexlab{}.
\newblock \bibinfo{booktitle}{\emph{Qsparse-local-SGD: distributed SGD with quantization, sparsification, and local computations}}.
\newblock \bibinfo{publisher}{Curran Associates Inc.}, \bibinfo{address}{Red Hook, NY, USA}.
\newblock


\bibitem[\protect\citeauthoryear{Chen, Giannakis, Sun, and Yin}{Chen et~al\mbox{.}}{2018}]%
        {chen2018lag}
\bibfield{author}{\bibinfo{person}{Tianyi Chen}, \bibinfo{person}{Georgios~B. Giannakis}, \bibinfo{person}{Tao Sun}, {and} \bibinfo{person}{Wotao Yin}.} \bibinfo{year}{2018}\natexlab{}.
\newblock \showarticletitle{LAG: lazily aggregated gradient for communication-efficient distributed learning}. In \bibinfo{booktitle}{\emph{Proceedings of the 32nd International Conference on Neural Information Processing Systems}} \emph{(\bibinfo{series}{NIPS'18})}. \bibinfo{publisher}{Curran Associates Inc.}, \bibinfo{address}{Red Hook, NY, USA}, \bibinfo{pages}{5055–5065}.
\newblock


\bibitem[\protect\citeauthoryear{Chilimbi, Suzue, Apacible, and Kalyanaraman}{Chilimbi et~al\mbox{.}}{2014}]%
        {chilimbi2014distributed}
\bibfield{author}{\bibinfo{person}{Trishul Chilimbi}, \bibinfo{person}{Yutaka Suzue}, \bibinfo{person}{Johnson Apacible}, {and} \bibinfo{person}{Karthik Kalyanaraman}.} \bibinfo{year}{2014}\natexlab{}.
\newblock \showarticletitle{Project Adam: Building an Efficient and Scalable Deep Learning Training System}. In \bibinfo{booktitle}{\emph{11th USENIX Symposium on Operating Systems Design and Implementation (OSDI 14)}}. \bibinfo{publisher}{USENIX Association}, \bibinfo{address}{Broomfield, CO}, \bibinfo{pages}{571--582}.
\newblock
\showISBNx{978-1-931971-16-4}


\bibitem[\protect\citeauthoryear{Chollet et~al\mbox{.}}{Chollet et~al\mbox{.}}{2015}]%
        {chollet2015keras}
\bibfield{author}{\bibinfo{person}{Fran\c{c}ois Chollet} {et~al\mbox{.}}} \bibinfo{year}{2015}\natexlab{}.
\newblock \bibinfo{title}{Keras}.
\newblock
\newblock


\bibitem[\protect\citeauthoryear{Cormode and Garofalakis}{Cormode and Garofalakis}{2005}]%
        {cormode2005sketch}
\bibfield{author}{\bibinfo{person}{Graham Cormode} {and} \bibinfo{person}{Minos Garofalakis}.} \bibinfo{year}{2005}\natexlab{}.
\newblock \showarticletitle{Sketching Streams through the Net: Distributed Approximate Query Tracking}. In \bibinfo{booktitle}{\emph{Proceedings of the 31st International Conference on Very Large Data Bases}} \emph{(\bibinfo{series}{VLDB '05})}. \bibinfo{publisher}{VLDB Endowment}, \bibinfo{pages}{13–24}.
\newblock
\showISBNx{1595931546}


\bibitem[\protect\citeauthoryear{Cormode, Markov, and Srinivas}{Cormode et~al\mbox{.}}{2024}]%
        {CormodeMS24private}
\bibfield{author}{\bibinfo{person}{Graham Cormode}, \bibinfo{person}{Igor~L. Markov}, {and} \bibinfo{person}{Harish Srinivas}.} \bibinfo{year}{2024}\natexlab{}.
\newblock \showarticletitle{Private and Efficient Federated Numerical Aggregation}. In \bibinfo{booktitle}{\emph{Proceedings 27th International Conference on Extending Database Technology, EDBT 2024, Paestum, Italy, March 25 - March 28}}, \bibfield{editor}{\bibinfo{person}{Letizia Tanca}, \bibinfo{person}{Qiong~Luo 0001}, \bibinfo{person}{Giuseppe Polese}, \bibinfo{person}{Loredana Caruccio}, \bibinfo{person}{Xavier Oriol}, {and} \bibinfo{person}{Donatella Firmani}} (Eds.). \bibinfo{pages}{734--742}.
\newblock


\bibitem[\protect\citeauthoryear{Davitkova, Gjurovski, and 0001}{Davitkova et~al\mbox{.}}{2024}]%
        {DavitkovaGM24}
\bibfield{author}{\bibinfo{person}{Angjela Davitkova}, \bibinfo{person}{Damjan Gjurovski}, {and} \bibinfo{person}{Sebastian~Michel 0001}.} \bibinfo{year}{2024}\natexlab{}.
\newblock \showarticletitle{Learning over Sets for Databases}. In \bibinfo{booktitle}{\emph{Proceedings 27th International Conference on Extending Database Technology, EDBT 2024, Paestum, Italy, March 25 - March 28}}, \bibfield{editor}{\bibinfo{person}{Letizia Tanca}, \bibinfo{person}{Qiong~Luo 0001}, \bibinfo{person}{Giuseppe Polese}, \bibinfo{person}{Loredana Caruccio}, \bibinfo{person}{Xavier Oriol}, {and} \bibinfo{person}{Donatella Firmani}} (Eds.). \bibinfo{publisher}{OpenProceedings.org}, \bibinfo{pages}{68--80}.
\newblock
\showISBNx{978-3-89318-091-2}


\bibitem[\protect\citeauthoryear{Dean, Corrado, Monga, Chen, Devin, Mao, Ranzato, Senior, Tucker, Yang, Le, and Ng}{Dean et~al\mbox{.}}{2012}]%
        {Dean2012distributed}
\bibfield{author}{\bibinfo{person}{Jeffrey Dean}, \bibinfo{person}{Greg Corrado}, \bibinfo{person}{Rajat Monga}, \bibinfo{person}{Kai Chen}, \bibinfo{person}{Matthieu Devin}, \bibinfo{person}{Mark Mao}, \bibinfo{person}{Marc\textquotesingle~aurelio Ranzato}, \bibinfo{person}{Andrew Senior}, \bibinfo{person}{Paul Tucker}, \bibinfo{person}{Ke Yang}, \bibinfo{person}{Quoc Le}, {and} \bibinfo{person}{Andrew Ng}.} \bibinfo{year}{2012}\natexlab{}.
\newblock \showarticletitle{Large Scale Distributed Deep Networks}. In \bibinfo{booktitle}{\emph{Advances in Neural Information Processing Systems}}, \bibfield{editor}{\bibinfo{person}{F.~Pereira}, \bibinfo{person}{C.J. Burges}, \bibinfo{person}{L.~Bottou}, {and} \bibinfo{person}{K.Q. Weinberger}} (Eds.), Vol.~\bibinfo{volume}{25}. \bibinfo{publisher}{Curran Associates, Inc.}
\newblock


\bibitem[\protect\citeauthoryear{Deng}{Deng}{2012}]%
        {deng2012mnist}
\bibfield{author}{\bibinfo{person}{Li Deng}.} \bibinfo{year}{2012}\natexlab{}.
\newblock \showarticletitle{The mnist database of handwritten digit images for machine learning research}.
\newblock \bibinfo{journal}{\emph{IEEE Signal Processing Magazine}} \bibinfo{volume}{29}, \bibinfo{number}{6} (\bibinfo{year}{2012}), \bibinfo{pages}{141--142}.
\newblock


\bibitem[\protect\citeauthoryear{Dosovitskiy, Beyer, Kolesnikov, Weissenborn, Zhai, Unterthiner, Dehghani, Minderer, Heigold, Gelly, Uszkoreit, and Houlsby}{Dosovitskiy et~al\mbox{.}}{2021}]%
        {dosovitskiy2020vit}
\bibfield{author}{\bibinfo{person}{Alexey Dosovitskiy}, \bibinfo{person}{Lucas Beyer}, \bibinfo{person}{Alexander Kolesnikov}, \bibinfo{person}{Dirk Weissenborn}, \bibinfo{person}{Xiaohua Zhai}, \bibinfo{person}{Thomas Unterthiner}, \bibinfo{person}{Mostafa Dehghani}, \bibinfo{person}{Matthias Minderer}, \bibinfo{person}{Georg Heigold}, \bibinfo{person}{Sylvain Gelly}, \bibinfo{person}{Jakob Uszkoreit}, {and} \bibinfo{person}{Neil Houlsby}.} \bibinfo{year}{2021}\natexlab{}.
\newblock \showarticletitle{An Image is Worth 16x16 Words: Transformers for Image Recognition at Scale}. In \bibinfo{booktitle}{\emph{9th International Conference on Learning Representations, {ICLR} 2021, Virtual Event, Austria, May 3-7, 2021}}. \bibinfo{publisher}{OpenReview.net}.
\newblock


\bibitem[\protect\citeauthoryear{Fu, Miao, Jiang, Xue, and Cui}{Fu et~al\mbox{.}}{2022}]%
        {fu2022stale}
\bibfield{author}{\bibinfo{person}{Fangcheng Fu}, \bibinfo{person}{Xupeng Miao}, \bibinfo{person}{Jiawei Jiang}, \bibinfo{person}{Huanran Xue}, {and} \bibinfo{person}{Bin Cui}.} \bibinfo{year}{2022}\natexlab{}.
\newblock \showarticletitle{Towards communication-efficient vertical federated learning training via cache-enabled local updates}.
\newblock  \bibinfo{volume}{15}, \bibinfo{number}{10} (\bibinfo{date}{jun} \bibinfo{year}{2022}), \bibinfo{pages}{2111–2120}.
\newblock
\showISSN{2150-8097}


\bibitem[\protect\citeauthoryear{Glorot and Bengio}{Glorot and Bengio}{2010}]%
        {glorot2010uniform}
\bibfield{author}{\bibinfo{person}{Xavier Glorot} {and} \bibinfo{person}{Yoshua Bengio}.} \bibinfo{year}{2010}\natexlab{}.
\newblock \showarticletitle{Understanding the difficulty of training deep feedforward neural networks}. In \bibinfo{booktitle}{\emph{Proceedings of the Thirteenth International Conference on Artificial Intelligence and Statistics}} \emph{(\bibinfo{series}{Proceedings of Machine Learning Research})}, \bibfield{editor}{\bibinfo{person}{Yee~Whye Teh} {and} \bibinfo{person}{Mike Titterington}} (Eds.), Vol.~\bibinfo{volume}{9}. \bibinfo{publisher}{PMLR}, \bibinfo{address}{Chia Laguna Resort, Sardinia, Italy}, \bibinfo{pages}{249--256}.
\newblock


\bibitem[\protect\citeauthoryear{Goodfellow, Bengio, and Courville}{Goodfellow et~al\mbox{.}}{2016}]%
        {DeepLearningGoodfellow}
\bibfield{author}{\bibinfo{person}{Ian~J. Goodfellow}, \bibinfo{person}{Yoshua Bengio}, {and} \bibinfo{person}{Aaron Courville}.} \bibinfo{year}{2016}\natexlab{}.
\newblock \bibinfo{booktitle}{\emph{Deep Learning}}.
\newblock \bibinfo{publisher}{MIT Press}, \bibinfo{address}{Cambridge, MA, USA}.
\newblock


\bibitem[\protect\citeauthoryear{Haddadpour, Kamani, Mahdavi, and Cadambe}{Haddadpour et~al\mbox{.}}{2019}]%
        {haddadpour2019localSGD}
\bibfield{author}{\bibinfo{person}{Farzin Haddadpour}, \bibinfo{person}{Mohammad~Mahdi Kamani}, \bibinfo{person}{Mehrdad Mahdavi}, {and} \bibinfo{person}{Viveck~R. Cadambe}.} \bibinfo{year}{2019}\natexlab{}.
\newblock \showarticletitle{Local SGD with Periodic Averaging: Tighter Analysis and Adaptive Synchronization}. In \bibinfo{booktitle}{\emph{Neural Information Processing Systems}}.
\newblock


\bibitem[\protect\citeauthoryear{Han, Zhang, Ding, Gu, Liu, Huo, Qiu, Yao, Zhang, Zhang, Han, Huang, Jin, Lan, Liu, Liu, Lu, Qiu, Song, Tang, Wen, Yuan, Zhao, and Zhu}{Han et~al\mbox{.}}{2021}]%
        {han2021ptm}
\bibfield{author}{\bibinfo{person}{Xu Han}, \bibinfo{person}{Zhengyan Zhang}, \bibinfo{person}{Ning Ding}, \bibinfo{person}{Yuxian Gu}, \bibinfo{person}{Xiao Liu}, \bibinfo{person}{Yuqi Huo}, \bibinfo{person}{Jiezhong Qiu}, \bibinfo{person}{Yuan Yao}, \bibinfo{person}{Ao Zhang}, \bibinfo{person}{Liang Zhang}, \bibinfo{person}{Wentao Han}, \bibinfo{person}{Minlie Huang}, \bibinfo{person}{Qin Jin}, \bibinfo{person}{Yanyan Lan}, \bibinfo{person}{Yang Liu}, \bibinfo{person}{Zhiyuan Liu}, \bibinfo{person}{Zhiwu Lu}, \bibinfo{person}{Xipeng Qiu}, \bibinfo{person}{Ruihua Song}, \bibinfo{person}{Jie Tang}, \bibinfo{person}{Ji-Rong Wen}, \bibinfo{person}{Jinhui Yuan}, \bibinfo{person}{Wayne~Xin Zhao}, {and} \bibinfo{person}{Jun Zhu}.} \bibinfo{year}{2021}\natexlab{}.
\newblock \showarticletitle{Pre-trained models: Past, present and future}.
\newblock \bibinfo{journal}{\emph{AI Open}}  \bibinfo{volume}{2} (\bibinfo{year}{2021}), \bibinfo{pages}{225--250}.
\newblock
\showISSN{2666-6510}


\bibitem[\protect\citeauthoryear{He, Zhang, Ren, and Sun}{He et~al\mbox{.}}{2015}]%
        {he2015normal}
\bibfield{author}{\bibinfo{person}{Kaiming He}, \bibinfo{person}{Xiangyu Zhang}, \bibinfo{person}{Shaoqing Ren}, {and} \bibinfo{person}{Jian Sun}.} \bibinfo{year}{2015}\natexlab{}.
\newblock \showarticletitle{Delving Deep into Rectifiers: Surpassing Human-Level Performance on ImageNet Classification}. In \bibinfo{booktitle}{\emph{2015 IEEE International Conference on Computer Vision (ICCV)}}. \bibinfo{pages}{1026--1034}.
\newblock


\bibitem[\protect\citeauthoryear{Hoffer, Hubara, and Soudry}{Hoffer et~al\mbox{.}}{2017}]%
        {hoffer2018genGap}
\bibfield{author}{\bibinfo{person}{Elad Hoffer}, \bibinfo{person}{Itay Hubara}, {and} \bibinfo{person}{Daniel Soudry}.} \bibinfo{year}{2017}\natexlab{}.
\newblock \showarticletitle{Train longer, generalize better: closing the generalization gap in large batch training of neural networks}. In \bibinfo{booktitle}{\emph{Proceedings of the 31st International Conference on Neural Information Processing Systems}} \emph{(\bibinfo{series}{NIPS'17})}. \bibinfo{publisher}{Curran Associates Inc.}, \bibinfo{address}{Red Hook, NY, USA}, \bibinfo{pages}{1729–1739}.
\newblock
\showISBNx{9781510860964}


\bibitem[\protect\citeauthoryear{Hsu, Qi, and Brown}{Hsu et~al\mbox{.}}{2019}]%
        {hsu2019fedAvgM}
\bibfield{author}{\bibinfo{person}{Tzu{-}Ming~Harry Hsu}, \bibinfo{person}{Hang Qi}, {and} \bibinfo{person}{Matthew Brown}.} \bibinfo{year}{2019}\natexlab{}.
\newblock \showarticletitle{Measuring the Effects of Non-Identical Data Distribution for Federated Visual Classification}.
\newblock \bibinfo{journal}{\emph{CoRR}} (\bibinfo{year}{2019}).
\newblock


\bibitem[\protect\citeauthoryear{Huang, Liu, Maaten, and Weinberger}{Huang et~al\mbox{.}}{2017}]%
        {huang2016densenet}
\bibfield{author}{\bibinfo{person}{G. Huang}, \bibinfo{person}{Z. Liu}, \bibinfo{person}{L.~Van~Der Maaten}, {and} \bibinfo{person}{K.~Q. Weinberger}.} \bibinfo{year}{2017}\natexlab{}.
\newblock \showarticletitle{Densely Connected Convolutional Networks}. In \bibinfo{booktitle}{\emph{2017 IEEE Conference on Computer Vision and Pattern Recognition (CVPR)}}. \bibinfo{publisher}{IEEE Computer Society}, \bibinfo{address}{Los Alamitos, CA, USA}, \bibinfo{pages}{2261--2269}.
\newblock
\showISSN{1063-6919}


\bibitem[\protect\citeauthoryear{Kairouz, McMahan, Avent, Bellet, Bennis, Nitin~Bhagoji, Bonawitz, Charles, Cormode, Cummings, D’Oliveira, Eichner, El~Rouayheb, Evans, Gardner, Garrett, Gasc\'{o}n, Ghazi, Gibbons, Gruteser, Harchaoui, He, He, Huo, Hutchinson, Hsu, Jaggi, Javidi, Joshi, Khodak, Konecn\'{y}, Korolova, Koushanfar, Koyejo, Lepoint, Liu, Mittal, Mohri, Nock, \"{O}zg\"{u}r, Pagh, Qi, Ramage, Raskar, Raykova, Song, Song, Stich, Sun, Suresh, Tram\`{e}r, Vepakomma, Wang, Xiong, Xu, Yang, Yu, Yu, and Zhao}{Kairouz et~al\mbox{.}}{2021}]%
        {karouz2019advancementsFL}
\bibfield{author}{\bibinfo{person}{Peter Kairouz}, \bibinfo{person}{H.~Brendan McMahan}, \bibinfo{person}{Brendan Avent}, \bibinfo{person}{Aur\'{e}lien Bellet}, \bibinfo{person}{Mehdi Bennis}, \bibinfo{person}{Arjun Nitin~Bhagoji}, \bibinfo{person}{Kallista Bonawitz}, \bibinfo{person}{Zachary Charles}, \bibinfo{person}{Graham Cormode}, \bibinfo{person}{Rachel Cummings}, \bibinfo{person}{Rafael G.~L. D’Oliveira}, \bibinfo{person}{Hubert Eichner}, \bibinfo{person}{Salim El~Rouayheb}, \bibinfo{person}{David Evans}, \bibinfo{person}{Josh Gardner}, \bibinfo{person}{Zachary Garrett}, \bibinfo{person}{Adri\`{a} Gasc\'{o}n}, \bibinfo{person}{Badih Ghazi}, \bibinfo{person}{Phillip~B. Gibbons}, \bibinfo{person}{Marco Gruteser}, \bibinfo{person}{Zaid Harchaoui}, \bibinfo{person}{Chaoyang He}, \bibinfo{person}{Lie He}, \bibinfo{person}{Zhouyuan Huo}, \bibinfo{person}{Ben Hutchinson}, \bibinfo{person}{Justin Hsu}, \bibinfo{person}{Martin Jaggi}, \bibinfo{person}{Tara Javidi}, \bibinfo{person}{Gauri Joshi},
  \bibinfo{person}{Mikhail Khodak}, \bibinfo{person}{Jakub Konecn\'{y}}, \bibinfo{person}{Aleksandra Korolova}, \bibinfo{person}{Farinaz Koushanfar}, \bibinfo{person}{Sanmi Koyejo}, \bibinfo{person}{Tancr\`{e}de Lepoint}, \bibinfo{person}{Yang Liu}, \bibinfo{person}{Prateek Mittal}, \bibinfo{person}{Mehryar Mohri}, \bibinfo{person}{Richard Nock}, \bibinfo{person}{Ayfer \"{O}zg\"{u}r}, \bibinfo{person}{Rasmus Pagh}, \bibinfo{person}{Hang Qi}, \bibinfo{person}{Daniel Ramage}, \bibinfo{person}{Ramesh Raskar}, \bibinfo{person}{Mariana Raykova}, \bibinfo{person}{Dawn Song}, \bibinfo{person}{Weikang Song}, \bibinfo{person}{Sebastian~U. Stich}, \bibinfo{person}{Ziteng Sun}, \bibinfo{person}{Ananda~Theertha Suresh}, \bibinfo{person}{Florian Tram\`{e}r}, \bibinfo{person}{Praneeth Vepakomma}, \bibinfo{person}{Jianyu Wang}, \bibinfo{person}{Li Xiong}, \bibinfo{person}{Zheng Xu}, \bibinfo{person}{Qiang Yang}, \bibinfo{person}{Felix~X. Yu}, \bibinfo{person}{Han Yu}, {and} \bibinfo{person}{Sen Zhao}.}
  \bibinfo{year}{2021}\natexlab{}.
\newblock \showarticletitle{Advances and Open Problems in Federated Learning}.
\newblock \bibinfo{journal}{\emph{Foundations and Trends in Machine Learning}} \bibinfo{volume}{14}, \bibinfo{number}{1–2} (\bibinfo{date}{jun} \bibinfo{year}{2021}), \bibinfo{pages}{1–210}.
\newblock
\showISSN{1935-8237}


\bibitem[\protect\citeauthoryear{Karimireddy, Jaggi, Kale, Mohri, Reddi, Stich, and Suresh}{Karimireddy et~al\mbox{.}}{2021}]%
        {karimireddy2021mime}
\bibfield{author}{\bibinfo{person}{Sai~Praneeth Karimireddy}, \bibinfo{person}{Martin Jaggi}, \bibinfo{person}{Satyen Kale}, \bibinfo{person}{Mehryar Mohri}, \bibinfo{person}{Sashank~J. Reddi}, \bibinfo{person}{Sebastian~U Stich}, {and} \bibinfo{person}{Ananda~Theertha Suresh}.} \bibinfo{year}{2021}\natexlab{}.
\newblock \bibinfo{title}{Mime: Mimicking Centralized Stochastic Algorithms in Federated Learning}.
\newblock
\newblock


\bibitem[\protect\citeauthoryear{Karimireddy, Kale, Mohri, Reddi, Stich, and Suresh}{Karimireddy et~al\mbox{.}}{2020}]%
        {kale2019scaffold}
\bibfield{author}{\bibinfo{person}{Sai~Praneeth Karimireddy}, \bibinfo{person}{Satyen Kale}, \bibinfo{person}{Mehryar Mohri}, \bibinfo{person}{Sashank Reddi}, \bibinfo{person}{Sebastian Stich}, {and} \bibinfo{person}{Ananda~Theertha Suresh}.} \bibinfo{year}{2020}\natexlab{}.
\newblock \showarticletitle{{SCAFFOLD}: Stochastic Controlled Averaging for Federated Learning}. In \bibinfo{booktitle}{\emph{Proceedings of the 37th International Conference on Machine Learning}} \emph{(\bibinfo{series}{Proceedings of Machine Learning Research})}, \bibfield{editor}{\bibinfo{person}{Hal~Daumé III} {and} \bibinfo{person}{Aarti Singh}} (Eds.), Vol.~\bibinfo{volume}{119}. \bibinfo{publisher}{PMLR}, \bibinfo{pages}{5132--5143}.
\newblock


\bibitem[\protect\citeauthoryear{Kingma and Ba}{Kingma and Ba}{2015}]%
        {kingma2017adam}
\bibfield{author}{\bibinfo{person}{Diederik Kingma} {and} \bibinfo{person}{Jimmy Ba}.} \bibinfo{year}{2015}\natexlab{}.
\newblock \showarticletitle{Adam: A Method for Stochastic Optimization}. In \bibinfo{booktitle}{\emph{International Conference on Learning Representations (ICLR)}}. \bibinfo{address}{San Diego, CA, USA}.
\newblock


\bibitem[\protect\citeauthoryear{Krizhevsky}{Krizhevsky}{2012}]%
        {krizhevsky2012cifar10}
\bibfield{author}{\bibinfo{person}{Alex Krizhevsky}.} \bibinfo{year}{2012}\natexlab{}.
\newblock \showarticletitle{Learning Multiple Layers of Features from Tiny Images}.
\newblock \bibinfo{journal}{\emph{University of Toronto}} (\bibinfo{date}{05} \bibinfo{year}{2012}).
\newblock


\bibitem[\protect\citeauthoryear{Lai, Zolaktaf, Milani, AlOmeir, Cao, and Pottinger}{Lai et~al\mbox{.}}{2023}]%
        {lai2023queryDBdeeplearning}
\bibfield{author}{\bibinfo{person}{Eugenie~Yujing Lai}, \bibinfo{person}{Zainab Zolaktaf}, \bibinfo{person}{Mostafa Milani}, \bibinfo{person}{Omar AlOmeir}, \bibinfo{person}{Jianhao Cao}, {and} \bibinfo{person}{Rachel Pottinger}.} \bibinfo{year}{2023}\natexlab{}.
\newblock \showarticletitle{Workload-Aware Query Recommendation Using Deep Learning}. In \bibinfo{booktitle}{\emph{Proceedings 26th International Conference on Extending Database Technology, EDBT 2023, Ioannina, Greece, March 28-31, 2023}}, \bibfield{editor}{\bibinfo{person}{Julia Stoyanovich}, \bibinfo{person}{Jens Teubner}, \bibinfo{person}{Nikos Mamoulis}, \bibinfo{person}{Evaggelia Pitoura}, {and} \bibinfo{person}{Jan Mühlig}} (Eds.). \bibinfo{publisher}{OpenProceedings.org}, \bibinfo{pages}{53--65}.
\newblock
\showISBNx{978-3-89318-088-2}


\bibitem[\protect\citeauthoryear{Lecun, Bottou, Bengio, and Haffner}{Lecun et~al\mbox{.}}{1998}]%
        {lecun1998lenet}
\bibfield{author}{\bibinfo{person}{Y. Lecun}, \bibinfo{person}{L. Bottou}, \bibinfo{person}{Y. Bengio}, {and} \bibinfo{person}{P. Haffner}.} \bibinfo{year}{1998}\natexlab{}.
\newblock \showarticletitle{Gradient-based learning applied to document recognition}.
\newblock \bibinfo{journal}{\emph{Proc. IEEE}} \bibinfo{volume}{86}, \bibinfo{number}{11} (\bibinfo{year}{1998}), \bibinfo{pages}{2278--2324}.
\newblock


\bibitem[\protect\citeauthoryear{Li, Zhao, Varma, Salpekar, Noordhuis, Li, Paszke, Smith, Vaughan, Damania, and Chintala}{Li et~al\mbox{.}}{2020}]%
        {li2020pytorchdistributed}
\bibfield{author}{\bibinfo{person}{Shen Li}, \bibinfo{person}{Yanli Zhao}, \bibinfo{person}{Rohan Varma}, \bibinfo{person}{Omkar Salpekar}, \bibinfo{person}{Pieter Noordhuis}, \bibinfo{person}{Teng Li}, \bibinfo{person}{Adam Paszke}, \bibinfo{person}{Jeff Smith}, \bibinfo{person}{Brian Vaughan}, \bibinfo{person}{Pritam Damania}, {and} \bibinfo{person}{Soumith Chintala}.} \bibinfo{year}{2020}\natexlab{}.
\newblock \showarticletitle{PyTorch distributed: experiences on accelerating data parallel training}.
\newblock \bibinfo{journal}{\emph{Proc. VLDB Endow.}} \bibinfo{volume}{13}, \bibinfo{number}{12} (\bibinfo{date}{aug} \bibinfo{year}{2020}), \bibinfo{pages}{3005–3018}.
\newblock
\showISSN{2150-8097}


\bibitem[\protect\citeauthoryear{Li, Yang, Wang, and Zhang}{Li et~al\mbox{.}}{2021}]%
        {li2021communicationefficient}
\bibfield{author}{\bibinfo{person}{Xiang Li}, \bibinfo{person}{Wenhao Yang}, \bibinfo{person}{Shusen Wang}, {and} \bibinfo{person}{Zhihua Zhang}.} \bibinfo{year}{2021}\natexlab{}.
\newblock \bibinfo{title}{Communication-Efficient Local Decentralized SGD Methods}.
\newblock
\newblock
\showeprint[arxiv]{stat.ML/1910.09126}


\bibitem[\protect\citeauthoryear{Lin, Stich, Patel, and Jaggi}{Lin et~al\mbox{.}}{2020}]%
        {lin2018postLocal}
\bibfield{author}{\bibinfo{person}{Tao Lin}, \bibinfo{person}{Sebastian~U. Stich}, \bibinfo{person}{Kumar~Kshitij Patel}, {and} \bibinfo{person}{Martin Jaggi}.} \bibinfo{year}{2020}\natexlab{}.
\newblock \showarticletitle{Don't Use Large Mini-batches, Use Local {SGD}}. In \bibinfo{booktitle}{\emph{8th International Conference on Learning Representations, {ICLR} 2020, Addis Ababa, Ethiopia, April 26-30, 2020}}. \bibinfo{publisher}{OpenReview.net}.
\newblock


\bibitem[\protect\citeauthoryear{Liu, Mao, Wu, Feichtenhofer, Darrell, and Xie}{Liu et~al\mbox{.}}{2022}]%
        {liu2022convenet}
\bibfield{author}{\bibinfo{person}{Z. Liu}, \bibinfo{person}{H. Mao}, \bibinfo{person}{C. Wu}, \bibinfo{person}{C. Feichtenhofer}, \bibinfo{person}{T. Darrell}, {and} \bibinfo{person}{S. Xie}.} \bibinfo{year}{2022}\natexlab{}.
\newblock \showarticletitle{A ConvNet for the 2020s}. In \bibinfo{booktitle}{\emph{2022 IEEE/CVF Conference on Computer Vision and Pattern Recognition (CVPR)}}. \bibinfo{publisher}{IEEE Computer Society}, \bibinfo{address}{Los Alamitos, CA, USA}, \bibinfo{pages}{11966--11976}.
\newblock


\bibitem[\protect\citeauthoryear{Loshchilov and Hutter}{Loshchilov and Hutter}{2019}]%
        {losh2019adamw}
\bibfield{author}{\bibinfo{person}{Ilya Loshchilov} {and} \bibinfo{person}{Frank Hutter}.} \bibinfo{year}{2019}\natexlab{}.
\newblock \showarticletitle{Decoupled Weight Decay Regularization}. In \bibinfo{booktitle}{\emph{International Conference on Learning Representations}}.
\newblock


\bibitem[\protect\citeauthoryear{Ma, Yan, Cai, Huang, Wu, and Cheng}{Ma et~al\mbox{.}}{2023}]%
        {ma2023fec_comm}
\bibfield{author}{\bibinfo{person}{Kaihao Ma}, \bibinfo{person}{Xiao Yan}, \bibinfo{person}{Zhenkun Cai}, \bibinfo{person}{Yuzhen Huang}, \bibinfo{person}{Yidi Wu}, {and} \bibinfo{person}{James Cheng}.} \bibinfo{year}{2023}\natexlab{}.
\newblock \showarticletitle{FEC: Efficient Deep Recommendation Model Training with Flexible Embedding Communication}.
\newblock \bibinfo{journal}{\emph{Proc. ACM Manag. Data}} \bibinfo{volume}{1}, \bibinfo{number}{2}, Article \bibinfo{articleno}{165} (\bibinfo{date}{jun} \bibinfo{year}{2023}), \bibinfo{numpages}{21}~pages.
\newblock


\bibitem[\protect\citeauthoryear{McMahan, Moore, Ramage, Hampson, and Arcas}{McMahan et~al\mbox{.}}{2017}]%
        {mcmahan2017FedAvg}
\bibfield{author}{\bibinfo{person}{Brendan McMahan}, \bibinfo{person}{Eider Moore}, \bibinfo{person}{Daniel Ramage}, \bibinfo{person}{Seth Hampson}, {and} \bibinfo{person}{Blaise Aguera~y Arcas}.} \bibinfo{year}{2017}\natexlab{}.
\newblock \showarticletitle{{Communication-Efficient Learning of Deep Networks from Decentralized Data}}. In \bibinfo{booktitle}{\emph{Proceedings of the 20th International Conference on Artificial Intelligence and Statistics}} \emph{(\bibinfo{series}{Proceedings of Machine Learning Research})}, \bibfield{editor}{\bibinfo{person}{Aarti Singh} {and} \bibinfo{person}{Jerry Zhu}} (Eds.), Vol.~\bibinfo{volume}{54}. \bibinfo{publisher}{PMLR}, \bibinfo{pages}{1273--1282}.
\newblock


\bibitem[\protect\citeauthoryear{Miao, Nie, Shao, Yang, Jiang, Ma, and Cui}{Miao et~al\mbox{.}}{2021}]%
        {miao2021partial_async}
\bibfield{author}{\bibinfo{person}{Xupeng Miao}, \bibinfo{person}{Xiaonan Nie}, \bibinfo{person}{Yingxia Shao}, \bibinfo{person}{Zhi Yang}, \bibinfo{person}{Jiawei Jiang}, \bibinfo{person}{Lingxiao Ma}, {and} \bibinfo{person}{Bin Cui}.} \bibinfo{year}{2021}\natexlab{}.
\newblock \showarticletitle{Heterogeneity-Aware Distributed Machine Learning Training via Partial Reduce}. In \bibinfo{booktitle}{\emph{Proceedings of the 2021 International Conference on Management of Data}} \emph{(\bibinfo{series}{SIGMOD '21})}. \bibinfo{publisher}{Association for Computing Machinery}, \bibinfo{address}{New York, NY, USA}, \bibinfo{pages}{2262–2270}.
\newblock
\showISBNx{9781450383431}


\bibitem[\protect\citeauthoryear{Mills, Hu, and Min}{Mills et~al\mbox{.}}{2023}]%
        {mills2023faster}
\bibfield{author}{\bibinfo{person}{Jed Mills}, \bibinfo{person}{Jia Hu}, {and} \bibinfo{person}{Geyong Min}.} \bibinfo{year}{2023}\natexlab{}.
\newblock \showarticletitle{Faster Federated Learning With Decaying Number of Local SGD Steps}.
\newblock \bibinfo{journal}{\emph{IEEE Trans. Parallel Distrib. Syst.}} \bibinfo{volume}{34}, \bibinfo{number}{7} (\bibinfo{date}{jul} \bibinfo{year}{2023}), \bibinfo{pages}{2198–2207}.
\newblock
\showISSN{1045-9219}


\bibitem[\protect\citeauthoryear{Nakandala and Kumar}{Nakandala and Kumar}{2022}]%
        {sigmod2022nautilus_tfl}
\bibfield{author}{\bibinfo{person}{Supun Nakandala} {and} \bibinfo{person}{Arun Kumar}.} \bibinfo{year}{2022}\natexlab{}.
\newblock \showarticletitle{Nautilus: An Optimized System for Deep Transfer Learning over Evolving Training Datasets}. In \bibinfo{booktitle}{\emph{Proceedings of the 2022 International Conference on Management of Data}} \emph{(\bibinfo{series}{SIGMOD '22})}. \bibinfo{publisher}{Association for Computing Machinery}, \bibinfo{address}{New York, NY, USA}, \bibinfo{pages}{506–520}.
\newblock
\showISBNx{9781450392495}
\urldef\tempurl%
\url{https://doi.org/10.1145/3514221.3517846}
\showDOI{\tempurl}


\bibitem[\protect\citeauthoryear{Qin, Etesami, and Uribe}{Qin et~al\mbox{.}}{2023}]%
        {qin2022roleofSteps}
\bibfield{author}{\bibinfo{person}{Tiancheng Qin}, \bibinfo{person}{S.~Rasoul Etesami}, {and} \bibinfo{person}{César~A. Uribe}.} \bibinfo{year}{2023}\natexlab{}.
\newblock \showarticletitle{The role of local steps in local SGD}.
\newblock \bibinfo{journal}{\emph{Optimization Methods and Software}} (\bibinfo{date}{Aug.} \bibinfo{year}{2023}), \bibinfo{pages}{1–27}.
\newblock
\showISSN{1029-4937}


\bibitem[\protect\citeauthoryear{Rauf, Freitas, and Paton}{Rauf et~al\mbox{.}}{2024}]%
        {RaufFP24}
\bibfield{author}{\bibinfo{person}{Hafiz~Tayyab Rauf}, \bibinfo{person}{André Freitas}, {and} \bibinfo{person}{Norman~W. Paton}.} \bibinfo{year}{2024}\natexlab{}.
\newblock \showarticletitle{Deep Clustering for Data Cleaning and Integration}. In \bibinfo{booktitle}{\emph{Proceedings 27th International Conference on Extending Database Technology, EDBT 2024, Paestum, Italy, March 25 - March 28}}, \bibfield{editor}{\bibinfo{person}{Letizia Tanca}, \bibinfo{person}{Qiong~Luo 0001}, \bibinfo{person}{Giuseppe Polese}, \bibinfo{person}{Loredana Caruccio}, \bibinfo{person}{Xavier Oriol}, {and} \bibinfo{person}{Donatella Firmani}} (Eds.). \bibinfo{publisher}{OpenProceedings.org}, \bibinfo{pages}{636--649}.
\newblock
\showISBNx{978-3-89318-091-2}
\urldef\tempurl%
\url{https://doi.org/10.48786/edbt.2024.55}
\showDOI{\tempurl}


\bibitem[\protect\citeauthoryear{Reddi, Charles, Zaheer, Garrett, Rush, Konečný, Kumar, and McMahan}{Reddi et~al\mbox{.}}{2021}]%
        {reddi2021fedadam}
\bibfield{editor}{\bibinfo{person}{Sashank Reddi}, \bibinfo{person}{Zachary~Burr Charles}, \bibinfo{person}{Manzil Zaheer}, \bibinfo{person}{Zachary Garrett}, \bibinfo{person}{Keith Rush}, \bibinfo{person}{Jakub Konečný}, \bibinfo{person}{Sanjiv Kumar}, {and} \bibinfo{person}{Brendan McMahan}} (Eds.). \bibinfo{year}{2021}\natexlab{}.
\newblock \bibinfo{booktitle}{\emph{Adaptive Federated Optimization}}.
\newblock


\bibitem[\protect\citeauthoryear{Rothchild, Panda, Ullah, Ivkin, Stoica, Braverman, Gonzalez, and Arora}{Rothchild et~al\mbox{.}}{2020}]%
        {rothchild2020fetchsgd}
\bibfield{author}{\bibinfo{person}{Daniel Rothchild}, \bibinfo{person}{Ashwinee Panda}, \bibinfo{person}{Enayat Ullah}, \bibinfo{person}{Nikita Ivkin}, \bibinfo{person}{Ion Stoica}, \bibinfo{person}{Vladimir Braverman}, \bibinfo{person}{Joseph Gonzalez}, {and} \bibinfo{person}{Raman Arora}.} \bibinfo{year}{2020}\natexlab{}.
\newblock \showarticletitle{FetchSGD: communication-efficient federated learning with sketching}. In \bibinfo{booktitle}{\emph{Proceedings of the 37th International Conference on Machine Learning}} \emph{(\bibinfo{series}{ICML'20})}. \bibinfo{publisher}{JMLR.org}, Article \bibinfo{articleno}{764}, \bibinfo{numpages}{13}~pages.
\newblock


\bibitem[\protect\citeauthoryear{Russakovsky, Deng, Su, Krause, Satheesh, Ma, Huang, Karpathy, Khosla, Bernstein, Berg, and Fei-Fei}{Russakovsky et~al\mbox{.}}{2015}]%
        {russakovsky2015imagenet}
\bibfield{author}{\bibinfo{person}{Olga Russakovsky}, \bibinfo{person}{Jia Deng}, \bibinfo{person}{Hao Su}, \bibinfo{person}{Jonathan Krause}, \bibinfo{person}{Sanjeev Satheesh}, \bibinfo{person}{Sean Ma}, \bibinfo{person}{Zhiheng Huang}, \bibinfo{person}{Andrej Karpathy}, \bibinfo{person}{Aditya Khosla}, \bibinfo{person}{Michael Bernstein}, \bibinfo{person}{Alexander~C. Berg}, {and} \bibinfo{person}{Li Fei-Fei}.} \bibinfo{year}{2015}\natexlab{}.
\newblock \showarticletitle{ImageNet Large Scale Visual Recognition Challenge}.
\newblock \bibinfo{journal}{\emph{International Journal of Computer Vision}} \bibinfo{volume}{115}, \bibinfo{number}{3} (\bibinfo{date}{01 Dec} \bibinfo{year}{2015}), \bibinfo{pages}{211--252}.
\newblock
\showISSN{1573-1405}


\bibitem[\protect\citeauthoryear{Sahu, Li, Sanjabi, Zaheer, Talwalkar, and Smith}{Sahu et~al\mbox{.}}{2018}]%
        {kumar2018fedProx}
\bibfield{author}{\bibinfo{person}{Anit Sahu}, \bibinfo{person}{Tian Li}, \bibinfo{person}{Maziar Sanjabi}, \bibinfo{person}{Manzil Zaheer}, \bibinfo{person}{Ameet Talwalkar}, {and} \bibinfo{person}{Virginia Smith}.} \bibinfo{year}{2018}\natexlab{}.
\newblock \showarticletitle{On the Convergence of Federated Optimization in Heterogeneous Networks}.
\newblock  (\bibinfo{date}{12} \bibinfo{year}{2018}).
\newblock


\bibitem[\protect\citeauthoryear{Shi, Tang, Chu, Liu, Wang, and Li}{Shi et~al\mbox{.}}{2021}]%
        {shi2021surveyUsingBert338mil}
\bibfield{author}{\bibinfo{person}{Shaohuai Shi}, \bibinfo{person}{Zhenheng Tang}, \bibinfo{person}{Xiaowen Chu}, \bibinfo{person}{Chengjian Liu}, \bibinfo{person}{Wei Wang}, {and} \bibinfo{person}{Bo Li}.} \bibinfo{year}{2021}\natexlab{}.
\newblock \showarticletitle{A Quantitative Survey of Communication Optimizations in Distributed Deep Learning}.
\newblock \bibinfo{journal}{\emph{IEEE Network}} \bibinfo{volume}{35}, \bibinfo{number}{3} (\bibinfo{year}{2021}), \bibinfo{pages}{230--237}.
\newblock


\bibitem[\protect\citeauthoryear{Shlezinger, Chen, Eldar, Poor, and Cui}{Shlezinger et~al\mbox{.}}{2021}]%
        {shlezinger2020quantizationFL}
\bibfield{author}{\bibinfo{person}{Nir Shlezinger}, \bibinfo{person}{Mingzhe Chen}, \bibinfo{person}{Yonina Eldar}, \bibinfo{person}{H.~Vincent Poor}, {and} \bibinfo{person}{Shuguang Cui}.} \bibinfo{year}{2021}\natexlab{}.
\newblock \showarticletitle{UVeQFed: Universal Vector Quantization for Federated Learning}.
\newblock \bibinfo{journal}{\emph{IEEE Transactions on Signal Processing}}  \bibinfo{volume}{69} (\bibinfo{date}{01} \bibinfo{year}{2021}), \bibinfo{pages}{500--514}.
\newblock


\bibitem[\protect\citeauthoryear{Simonyan and Zisserman}{Simonyan and Zisserman}{2015}]%
        {simonyan2014vgg}
\bibfield{author}{\bibinfo{person}{Karen Simonyan} {and} \bibinfo{person}{Andrew Zisserman}.} \bibinfo{year}{2015}\natexlab{}.
\newblock \showarticletitle{Very Deep Convolutional Networks for Large-Scale Image Recognition}. In \bibinfo{booktitle}{\emph{3rd International Conference on Learning Representations, {ICLR} 2015, San Diego, CA, USA, May 7-9, 2015, Conference Track Proceedings}}, \bibfield{editor}{\bibinfo{person}{Yoshua Bengio} {and} \bibinfo{person}{Yann LeCun}} (Eds.).
\newblock


\bibitem[\protect\citeauthoryear{Spring, Kyrillidis, Mohan, and Shrivastava}{Spring et~al\mbox{.}}{2019}]%
        {Spring2019grad}
\bibfield{author}{\bibinfo{person}{Ryan Spring}, \bibinfo{person}{Anastasios Kyrillidis}, \bibinfo{person}{Vijai Mohan}, {and} \bibinfo{person}{Anshumali Shrivastava}.} \bibinfo{year}{2019}\natexlab{}.
\newblock \showarticletitle{Compressing Gradient Optimizers via Count-Sketches}. In \bibinfo{booktitle}{\emph{Proceedings of the 36th International Conference on Machine Learning}} \emph{(\bibinfo{series}{Proceedings of Machine Learning Research})}, \bibfield{editor}{\bibinfo{person}{Kamalika Chaudhuri} {and} \bibinfo{person}{Ruslan Salakhutdinov}} (Eds.), Vol.~\bibinfo{volume}{97}. \bibinfo{publisher}{PMLR}, \bibinfo{pages}{5946--5955}.
\newblock


\bibitem[\protect\citeauthoryear{Stich}{Stich}{2019}]%
        {stich2019local}
\bibfield{author}{\bibinfo{person}{Sebastian~U. Stich}.} \bibinfo{year}{2019}\natexlab{}.
\newblock \showarticletitle{Local {SGD} Converges Fast and Communicates Little}. In \bibinfo{booktitle}{\emph{7th International Conference on Learning Representations, {ICLR} 2019, New Orleans, LA, USA, May 6-9, 2019}}. \bibinfo{publisher}{OpenReview.net}.
\newblock


\bibitem[\protect\citeauthoryear{Sutskever, Martens, Dahl, and Hinton}{Sutskever et~al\mbox{.}}{2013a}]%
        {sutskever2013momentumSGD}
\bibfield{author}{\bibinfo{person}{Ilya Sutskever}, \bibinfo{person}{James Martens}, \bibinfo{person}{George Dahl}, {and} \bibinfo{person}{Geoffrey Hinton}.} \bibinfo{year}{2013}\natexlab{a}.
\newblock \showarticletitle{On the importance of initialization and momentum in deep learning}. In \bibinfo{booktitle}{\emph{Proceedings of the 30th International Conference on Machine Learning}} \emph{(\bibinfo{series}{Proceedings of Machine Learning Research})}, \bibfield{editor}{\bibinfo{person}{Sanjoy Dasgupta} {and} \bibinfo{person}{David McAllester}} (Eds.), Vol.~\bibinfo{volume}{28}. \bibinfo{publisher}{PMLR}, \bibinfo{address}{Atlanta, Georgia, USA}, \bibinfo{pages}{1139--1147}.
\newblock


\bibitem[\protect\citeauthoryear{Sutskever, Martens, Dahl, and Hinton}{Sutskever et~al\mbox{.}}{2013b}]%
        {suts2013nesterov_mom}
\bibfield{author}{\bibinfo{person}{Ilya Sutskever}, \bibinfo{person}{James Martens}, \bibinfo{person}{George Dahl}, {and} \bibinfo{person}{Geoffrey Hinton}.} \bibinfo{year}{2013}\natexlab{b}.
\newblock \showarticletitle{On the importance of initialization and momentum in deep learning}. In \bibinfo{booktitle}{\emph{Proceedings of the 30th International Conference on Machine Learning}} \emph{(\bibinfo{series}{Proceedings of Machine Learning Research})}, \bibfield{editor}{\bibinfo{person}{Sanjoy Dasgupta} {and} \bibinfo{person}{David McAllester}} (Eds.), Vol.~\bibinfo{volume}{28}. \bibinfo{publisher}{PMLR}, \bibinfo{address}{Atlanta, Georgia, USA}, \bibinfo{pages}{1139--1147}.
\newblock


\bibitem[\protect\citeauthoryear{Tang, Shi, Li, and Chu}{Tang et~al\mbox{.}}{2023}]%
        {tang2023gossipfl}
\bibfield{author}{\bibinfo{person}{Zhenheng Tang}, \bibinfo{person}{Shaohuai Shi}, \bibinfo{person}{Bo Li}, {and} \bibinfo{person}{Xiaowen Chu}.} \bibinfo{year}{2023}\natexlab{}.
\newblock \showarticletitle{GossipFL: A Decentralized Federated Learning Framework With Sparsified and Adaptive Communication}.
\newblock \bibinfo{journal}{\emph{IEEE Transactions on Parallel and Distributed Systems}} \bibinfo{volume}{34}, \bibinfo{number}{3} (\bibinfo{year}{2023}), \bibinfo{pages}{909--922}.
\newblock


\bibitem[\protect\citeauthoryear{Thompson, Greenewald, Lee, and Manso}{Thompson et~al\mbox{.}}{2021}]%
        {thompson2021dr}
\bibfield{author}{\bibinfo{person}{Neil~C. Thompson}, \bibinfo{person}{Kristjan Greenewald}, \bibinfo{person}{Keeheon Lee}, {and} \bibinfo{person}{Gabriel~F. Manso}.} \bibinfo{year}{2021}\natexlab{}.
\newblock \showarticletitle{Deep Learning's Diminishing Returns: The Cost of Improvement is Becoming Unsustainable}.
\newblock \bibinfo{journal}{\emph{IEEE Spectrum}} \bibinfo{volume}{58}, \bibinfo{number}{10} (\bibinfo{year}{2021}), \bibinfo{pages}{50--55}.
\newblock


\bibitem[\protect\citeauthoryear{Um, Oh, Seo, Kweun, Kim, and Lee}{Um et~al\mbox{.}}{2023}]%
        {um2023fastflow}
\bibfield{author}{\bibinfo{person}{Taegeon Um}, \bibinfo{person}{Byungsoo Oh}, \bibinfo{person}{Byeongchan Seo}, \bibinfo{person}{Minhyeok Kweun}, \bibinfo{person}{Goeun Kim}, {and} \bibinfo{person}{Woo-Yeon Lee}.} \bibinfo{year}{2023}\natexlab{}.
\newblock \showarticletitle{FastFlow: Accelerating Deep Learning Model Training with Smart Offloading of Input Data Pipeline}.
\newblock  \bibinfo{volume}{16}, \bibinfo{number}{5} (\bibinfo{date}{jan} \bibinfo{year}{2023}), \bibinfo{pages}{1086–1099}.
\newblock
\showISSN{2150-8097}


\bibitem[\protect\citeauthoryear{Valiant}{Valiant}{1990}]%
        {valiant1990bsp}
\bibfield{author}{\bibinfo{person}{Leslie~G. Valiant}.} \bibinfo{year}{1990}\natexlab{}.
\newblock \showarticletitle{A Bridging Model for Parallel Computation}.
\newblock \bibinfo{journal}{\emph{Commun. ACM}} \bibinfo{volume}{33}, \bibinfo{number}{8} (\bibinfo{date}{aug} \bibinfo{year}{1990}), \bibinfo{pages}{103–111}.
\newblock
\showISSN{0001-0782}


\bibitem[\protect\citeauthoryear{Wang and Joshi}{Wang and Joshi}{[n.d.]}]%
        {wang2018adaptive}
\bibfield{author}{\bibinfo{person}{Jianyu Wang} {and} \bibinfo{person}{Gauri Joshi}.} \bibinfo{year}{[n.d.]}\natexlab{}.
\newblock \showarticletitle{Adaptive Communication Strategies to Achieve the Best Error-Runtime Trade-off in Local-update SGD}.
\newblock \bibinfo{journal}{\emph{Systems and Machine Learning (SysML) Conference}} (\bibinfo{year}{[n.\,d.]}).
\newblock


\bibitem[\protect\citeauthoryear{Wang and Joshi}{Wang and Joshi}{2021}]%
        {wang2021cooperativeSGD}
\bibfield{author}{\bibinfo{person}{Jianyu Wang} {and} \bibinfo{person}{Gauri Joshi}.} \bibinfo{year}{2021}\natexlab{}.
\newblock \showarticletitle{Cooperative SGD: A Unified Framework for the Design and Analysis of Local-Update SGD Algorithms}.
\newblock \bibinfo{journal}{\emph{Journal of Machine Learning Research}} \bibinfo{volume}{22}, \bibinfo{number}{213} (\bibinfo{year}{2021}), \bibinfo{pages}{1--50}.
\newblock


\bibitem[\protect\citeauthoryear{Wang, Tuor, Salonidis, Leung, Makaya, He, and Chan}{Wang et~al\mbox{.}}{2019}]%
        {wang2019adaptiveFL}
\bibfield{author}{\bibinfo{person}{Shiqiang Wang}, \bibinfo{person}{Tiffany Tuor}, \bibinfo{person}{Theodoros Salonidis}, \bibinfo{person}{Kin~K. Leung}, \bibinfo{person}{Christian Makaya}, \bibinfo{person}{Ting He}, {and} \bibinfo{person}{Kevin Chan}.} \bibinfo{year}{2019}\natexlab{}.
\newblock \showarticletitle{Adaptive Federated Learning in Resource Constrained Edge Computing Systems}.
\newblock \bibinfo{journal}{\emph{IEEE Journal on Selected Areas in Communications}} \bibinfo{volume}{37}, \bibinfo{number}{6} (\bibinfo{year}{2019}), \bibinfo{pages}{1205--1221}.
\newblock


\bibitem[\protect\citeauthoryear{Wang, Zhang, Chen, Jagadish, Ooi, and Tan}{Wang et~al\mbox{.}}{2016}]%
        {wang2016databases_and_ddl}
\bibfield{author}{\bibinfo{person}{Wei Wang}, \bibinfo{person}{Meihui Zhang}, \bibinfo{person}{Gang Chen}, \bibinfo{person}{H.~V. Jagadish}, \bibinfo{person}{Beng~Chin Ooi}, {and} \bibinfo{person}{Kian-Lee Tan}.} \bibinfo{year}{2016}\natexlab{}.
\newblock \showarticletitle{Database Meets Deep Learning: Challenges and Opportunities}.
\newblock  \bibinfo{volume}{45}, \bibinfo{number}{2} (\bibinfo{date}{sep} \bibinfo{year}{2016}), \bibinfo{pages}{17–22}.
\newblock
\showISSN{0163-5808}


\bibitem[\protect\citeauthoryear{Wang, Wen, Xu, Zhou, Wang, and Zhang}{Wang et~al\mbox{.}}{2023}]%
        {wang2023compressionInDDL}
\bibfield{author}{\bibinfo{person}{Zeqin Wang}, \bibinfo{person}{Ming Wen}, \bibinfo{person}{Yuedong Xu}, \bibinfo{person}{Yipeng Zhou}, \bibinfo{person}{Jessie~Hui Wang}, {and} \bibinfo{person}{Liang Zhang}.} \bibinfo{year}{2023}\natexlab{}.
\newblock \showarticletitle{Communication compression techniques in distributed deep learning: A survey}.
\newblock \bibinfo{journal}{\emph{Journal of Systems Architecture}}  \bibinfo{volume}{142} (\bibinfo{year}{2023}), \bibinfo{pages}{102927}.
\newblock
\showISSN{1383-7621}


\bibitem[\protect\citeauthoryear{Waskom}{Waskom}{2021}]%
        {waskom2021seaborn}
\bibfield{author}{\bibinfo{person}{Michael~L. Waskom}.} \bibinfo{year}{2021}\natexlab{}.
\newblock \showarticletitle{seaborn: statistical data visualization}.
\newblock \bibinfo{journal}{\emph{Journal of Open Source Software}} \bibinfo{volume}{6}, \bibinfo{number}{60} (\bibinfo{year}{2021}), \bibinfo{pages}{3021}.
\newblock


\bibitem[\protect\citeauthoryear{Wenig and Papenbrock}{Wenig and Papenbrock}{2022}]%
        {WenigP22}
\bibfield{author}{\bibinfo{person}{Phillip Wenig} {and} \bibinfo{person}{Thorsten Papenbrock}.} \bibinfo{year}{2022}\natexlab{}.
\newblock \showarticletitle{DataGossip: A Data Exchange Extension for Distributed Machine Learning Algorithms}. In \bibinfo{booktitle}{\emph{Proceedings of the 25th International Conference on Extending Database Technology, EDBT 2022, Edinburgh, UK, March 29 - April 1, 2022}}, \bibfield{editor}{\bibinfo{person}{Julia Stoyanovich}, \bibinfo{person}{Jens Teubner}, \bibinfo{person}{Paolo Guagliardo}, \bibinfo{person}{Milos Nikolic}, \bibinfo{person}{Andreas Pieris}, \bibinfo{person}{Jan Mühlig}, \bibinfo{person}{Fatma Özcan}, \bibinfo{person}{Sebastian Schelter}, \bibinfo{person}{H.~V. Jagadish}, {and} \bibinfo{person}{Meihui~Zhang 0001}} (Eds.).
\newblock


\bibitem[\protect\citeauthoryear{Wu, Cai, Xiao, Chen, and Ooi}{Wu et~al\mbox{.}}{2020}]%
        {wu2020privacy_preserv_fl}
\bibfield{author}{\bibinfo{person}{Yuncheng Wu}, \bibinfo{person}{Shaofeng Cai}, \bibinfo{person}{Xiaokui Xiao}, \bibinfo{person}{Gang Chen}, {and} \bibinfo{person}{Beng~Chin Ooi}.} \bibinfo{year}{2020}\natexlab{}.
\newblock \showarticletitle{Privacy preserving vertical federated learning for tree-based models}.
\newblock \bibinfo{journal}{\emph{Proc. VLDB Endow.}} \bibinfo{volume}{13}, \bibinfo{number}{12} (\bibinfo{date}{jul} \bibinfo{year}{2020}), \bibinfo{pages}{2090–2103}.
\newblock
\showISSN{2150-8097}


\bibitem[\protect\citeauthoryear{Yu and Jin}{Yu and Jin}{2019}]%
        {yu2019computation}
\bibfield{author}{\bibinfo{person}{Hao Yu} {and} \bibinfo{person}{Rong Jin}.} \bibinfo{year}{2019}\natexlab{}.
\newblock \showarticletitle{On the Computation and Communication Complexity of Parallel {SGD} with Dynamic Batch Sizes for Stochastic Non-Convex Optimization}. In \bibinfo{booktitle}{\emph{Proceedings of the 36th International Conference on Machine Learning, {ICML} 2019, 9-15 June 2019, Long Beach, California, {USA}}} \emph{(\bibinfo{series}{Proceedings of Machine Learning Research})}, \bibfield{editor}{\bibinfo{person}{Kamalika Chaudhuri} {and} \bibinfo{person}{Ruslan Salakhutdinov}} (Eds.), Vol.~\bibinfo{volume}{97}. \bibinfo{publisher}{{PMLR}}, \bibinfo{pages}{7174--7183}.
\newblock


\bibitem[\protect\citeauthoryear{Yu, Yang, and Zhu}{Yu et~al\mbox{.}}{2019}]%
        {yu2018parallel}
\bibfield{author}{\bibinfo{person}{Hao Yu}, \bibinfo{person}{Sen Yang}, {and} \bibinfo{person}{Shenghuo Zhu}.} \bibinfo{year}{2019}\natexlab{}.
\newblock \showarticletitle{Parallel restarted SGD with faster convergence and less communication: demystifying why model averaging works for deep learning}. In \bibinfo{booktitle}{\emph{Proceedings of the Thirty-Third AAAI Conference on Artificial Intelligence and Thirty-First Innovative Applications of Artificial Intelligence Conference and Ninth AAAI Symposium on Educational Advances in Artificial Intelligence}} \emph{(\bibinfo{series}{AAAI'19/IAAI'19/EAAI'19})}. \bibinfo{publisher}{AAAI Press}, Article \bibinfo{articleno}{698}, \bibinfo{numpages}{8}~pages.
\newblock
\showISBNx{978-1-57735-809-1}


\bibitem[\protect\citeauthoryear{Zhang, McQuillan, Jayaram, Kak, Khanna, Kislal, Valdano, and Kumar}{Zhang et~al\mbox{.}}{2021}]%
        {zhang2021data}
\bibfield{author}{\bibinfo{person}{Yuhao Zhang}, \bibinfo{person}{Frank McQuillan}, \bibinfo{person}{Nandish Jayaram}, \bibinfo{person}{Nikhil Kak}, \bibinfo{person}{Ekta Khanna}, \bibinfo{person}{Orhan Kislal}, \bibinfo{person}{Domino Valdano}, {and} \bibinfo{person}{Arun Kumar}.} \bibinfo{year}{2021}\natexlab{}.
\newblock \showarticletitle{Distributed deep learning on data systems: a comparative analysis of approaches}.
\newblock \bibinfo{journal}{\emph{Proc. VLDB Endow.}} \bibinfo{volume}{14}, \bibinfo{number}{10} (\bibinfo{date}{jun} \bibinfo{year}{2021}), \bibinfo{pages}{1769–1782}.
\newblock
\showISSN{2150-8097}


\bibitem[\protect\citeauthoryear{Zhou, Chen, Das, Min, Yu, Zhao, and Zou}{Zhou et~al\mbox{.}}{2022}]%
        {zhou2022vision}
\bibfield{author}{\bibinfo{person}{Lixi Zhou}, \bibinfo{person}{Jiaqing Chen}, \bibinfo{person}{Amitabh Das}, \bibinfo{person}{Hong Min}, \bibinfo{person}{Lei Yu}, \bibinfo{person}{Ming Zhao}, {and} \bibinfo{person}{Jia Zou}.} \bibinfo{year}{2022}\natexlab{}.
\newblock \showarticletitle{Serving deep learning models with deduplication from relational databases}.
\newblock \bibinfo{journal}{\emph{Proc. VLDB Endow.}} \bibinfo{volume}{15}, \bibinfo{number}{10} (\bibinfo{date}{jun} \bibinfo{year}{2022}), \bibinfo{pages}{2230–2243}.
\newblock
\showISSN{2150-8097}


\bibitem[\protect\citeauthoryear{Zinkevich, Weimer, Li, and Smola}{Zinkevich et~al\mbox{.}}{2010}]%
        {zinkevich2010parallelSGD}
\bibfield{author}{\bibinfo{person}{Martin Zinkevich}, \bibinfo{person}{Markus Weimer}, \bibinfo{person}{Lihong Li}, {and} \bibinfo{person}{Alex Smola}.} \bibinfo{year}{2010}\natexlab{}.
\newblock \showarticletitle{Parallelized Stochastic Gradient Descent}. In \bibinfo{booktitle}{\emph{Advances in Neural Information Processing Systems}}, \bibfield{editor}{\bibinfo{person}{J.~Lafferty}, \bibinfo{person}{C.~Williams}, \bibinfo{person}{J.~Shawe-Taylor}, \bibinfo{person}{R.~Zemel}, {and} \bibinfo{person}{A.~Culotta}} (Eds.), Vol.~\bibinfo{volume}{23}. \bibinfo{publisher}{Curran Associates, Inc.}
\newblock


\end{thebibliography}
